\documentclass[preprint,1p,sort&compress]{elsarticle}
\usepackage{hyperref}
 
\usepackage{graphicx} 
\usepackage{tipa} 
\usepackage{color} 
\usepackage{rotating} 
\usepackage{multirow,multicol} 
\usepackage[section]{placeins} 
\usepackage{arydshln}  
\makeatletter
\def\ps@pprintTitle{%
 \let\@oddhead\@empty
 \let\@evenhead\@empty
 \def\@oddfoot{{\footnotesize\textit{Preprint submitted to Speech Communication Special Issue on AV expressive speech and gesture. July, 2017}}}%
 \let\@evenfoot\@oddfoot}
 \makeatother

\journal{Speech Communication, Special Issue on AV expressive speech and gesture.}
\begin{document} 

\begin{frontmatter}

\title{Phoneme-to-viseme mappings: the good, the bad, and the ugly}

\author[mymainaddress]{Helen L Bear\cortext[mycorrespondingauthor]{Corresponding author}}
\ead{h.l.bear@uel.ac.uk}

\author[mysecondaryaddress]{Richard Harvey}
\ead[url]{https://www.uea.ac.uk/computing/people/profile/r-w-harvey}
 
\address[mymainaddress]{University of East London, London E16 2RD UK}
\address[mysecondaryaddress]{University of East Anglia, Norwich, Norfolk, NR4 7TJ UK}

 
\begin{abstract} 
Visemes are the visual equivalent of phonemes. Although not precisely defined, a working definition of a viseme is ``a set of phonemes which have identical appearance on the lips''. Therefore a phoneme falls into one viseme class but a viseme may represent many phonemes: a many to one mapping. This mapping introduces ambiguity between phonemes when using viseme classifiers. Not only is this ambiguity damaging to the performance of audio-visual classifiers operating on real expressive speech, there is also considerable choice between possible mappings. 

In this paper we explore the issue of this choice of viseme-to-phoneme map. We show that there is definite difference in performance between viseme-to-phoneme mappings and explore why some maps appear to work better than others. We also devise a new algorithm for constructing phoneme-to-viseme mappings from labeled speech data. These new visemes, `Bear' visemes, are shown to perform better than previously known units.
\end{abstract} 
 

\end{frontmatter}


\section{Introduction} 
Recognition and synthesis of expressive audio-visual speech has proven to be a most challenging problem. When comparing audio-visual speech with acoustic recognition, one can identify several sources of difficulty. Firstly, the visual component of speech brings new problems such as pose, lighting, frame rate, resolution, and so on. Secondly, old problems in acoustic recognition, such as person specificity or the optimal recognition units, appear in new ways in the visual domain. While some of these aspects have been partially studied, progress has been hampered by very small datasets. Furthermore, reliable tracking has eluded many researchers which in turn has led to sub-optimal feature extraction, consequent poor performance and hence, incorrect conclusions about the parts of the problem that are tractable or intractable. A further challenge is the lack of consensus on the recognition units and it is commonplace to need to compare, say, word error rates with viseme error rates computed from a different set of visemes. Our contention is that progress in expressive audio-visual speech will remain stunted while this fundamental uncertainty remains. In this paper we review the choice of visual recognition units and provide a comprehensive set of evaluations of the competing phoneme-to-viseme mappings. We give guidance on what works well and provide explanations for the differences in performance. 
We also devise new algorithms for selecting optimal visual units should this be desired.
 
We should note that while this paper tends to focus on visual-only recognition, or lipreading, this aspect is by far the most challenging so progress on lipreading can be used to provide more useful audio-visual systems.

The rest of this paper is structured as follows: we discuss the current restrictions on a conventional lipreading system and identify the limitation of each upon the system. We then study the current sets of published visemes, before presenting a new speaker-dependent clustering algorithm for creating sets of visemes for individual speakers. We show that creating these speaker-dependent visemes follows from simple clustering and merge algorithms. These new visemes are tested on both isolated words and continuous speech datasets before we evaluate the efficacy of the improved performance against the extra investment into a new lipreading system. Since it is computationally simple to develop these speaker-dependent visemes we contend they are also a useful step in the analysis of speaker variability which itself is one of the more challenging problems in general lipreading.
 
 
\section{Limitations in lipreading systems} 
It is often said that lipreading is difficult because not all sounds appear on the lips\footnote{\cite{newman2010} compares the performance of a system that measures, via electromagnetic articulography, the hidden and visual parts of the mouth so the extent of this statement can be quantified.}. This is true but in reality there are a number of problems that can corrupt the lipreading signal even before one reaches the problem of trying to decode the visual signal. Table~\ref{tab:mlraffects} provides a taxonomy of the challenges in lipreading. Some of them relate to the problems of extracting useful information from the visual signal whereas some appear later in the signal processing chain and relate to the coding and classification of the visual signal.
\begin{table}[!ht] 
\centering 
\caption{Challenges to successful machine lipreading. Each challenge has some references.} 
\begin{tabular}{|l|l|l|} 
\hline 
Evaluation & Previously studied? \\ 
\hline \hline 
Motion & Yes, \cite{ong2008robust, Matthews_Baker_2004, 927467} \\ 
Pose & Yes, \cite{6298439, pass2010investigation, Moore2011541, 4218129, kaucic1998accurate, 11011995,lucey2009visual} \\ 
Expression & Yes, \cite{pass2010investigation, Moore2011541} \\ 
Frame rate & Yes, \cite{Blokland199897, saitoh2010study} \\ 
Video quality & Yes \cite{bearicip, heckmann2003effects, ACP:ACP371} \\ 
Color & Yes, \cite{kaucic1998accurate} \\ 
Unit choice & Yes, \cite{cappelletta2012phoneme, howell2013confusion, Hazen1027972, shin2011real, bear2016decoding} \\ 
Feature & Yes, \cite{matthews1998nonlinear, lan2009comparing, improveVis, 927467, Matthews_Baker_2004} \\ 
Classifier technology & Yes, \cite{982900, htk34, zhu2000use, cappelletta2012phoneme, thangthai2015improving} \\ 
Multiple persons & Yes, \cite{871067, simstruct, visualvowelpercept, 871073} \\ 
Speaker identity & Yes, \cite{607030,bear2015speaker,newman2010speaker} \\ 
Rate of speech & Yes, \cite{6854158,bear2016decoding} \\ 
\hline 
\end{tabular} 
\label{tab:mlraffects} 
\end{table} 

{\bf Motion} is an important part of almost all realistic settings. It is therefore essential to have either some form of tracking or to devise features that are invariant to non-informational motions. An early dataset which captured speaker motion (not camera motion) is CUAVE \cite{5745028}. Lipreading experiments on this dataset such as \cite{1415153} examine  two different features, one based on the Discrete Cosine Transform (DCT) and another on the Active Appearance Model (AAM).  The AAM (which can be shape-only, appearance-only or shape and appearance models) \cite{927467} sometimes preceded by Linear Predictors (LP) \cite{ong2008robust}. An AAM \cite{927467} is a model trained on a combination of shape and/or appearance information from a subset of video frames. The model is usually built from video frames manually labeled with landmarks which are chosen to cover the full range of motion throughout the video. In \cite{1415153} they prefer the DCT but  note that there were implementation difficulties with the AAM which meant it was improperly tracked. Further lip-reading experiments on CUAVE \cite{7760575} clarifies how challenging comparing results is, because there is no agreed evaluation protocol which could account for the motion challenge/face alignment. This is attributed to their partial success with particular speakers.  

The majority of automatic lipreading systems use a frontal {\bf pose} in which the speaker's facial place is normal to the principal ray of the camera. However in \cite{Moore2011541} for example, an improvement in expression recognition is seen by both computers and humans when the pose is rotated to 45$^{\circ}$. Other work \cite{4218129, kaucic1998accurate}, looks more specifically at visual speech recognition and suggests that a profile view of a speaker may not lead to catastrophically low accuracies. This observation is consistent with \cite{11011995} which measures human sentence perception from three viewing angles: full-frontal view (0$^{\circ}$), angled view (45$^{\circ}$), and side view (90$^{\circ}$). In this single-subject study a post-lingual deaf woman was tested to measure accuracy at the three angles independently. The three angles were randomly presented in every lipreading session. The results indicated that the side-view angle is most effective. A model for pose-mismatched lipreading is presented in \cite{lucey2009visual} in which it is shown that without training data at the correct pose, the recognition accuracy falls dramatically. However, the authors also show that this can be mitigated by projecting the features back to a canonical pose. This transformation principle is also used in \cite{6298439} which presents a view-independent lipreading system. This investigation uses a continuous speech corpus compared to the small vocabulary dataset in \cite{lucey2009visual}. This later study acknowledges a human lipreaders preference for a non-frontal view and suggests it could be attributed to lip protrusion. They show that the 45$^{\circ}$ angle is preferable. In short, when it comes to pose, there is evidence that it can be accounted for and need not be insurmountable. Therefore, for this work we stick to frontal pose.

{\bf Expression} can be difficult to disentangle with the spoken word when lipreading natural speech. Smiling (a happy expression) has an known effect on lip motions during speech \cite{shor1978production}. Effects on the inner, outer lips and lip protrusions have been measured in \cite{fagel2010effects} who shows that smiling during speech (particularly vowels) places a restriction on lip motion with greater demand placed on the inner lips as variation in outer lips and lip protrusion is reduced. This in turn creates a greater challenge when lipreading non-neutral speech as gestures become less distinct. Furthermore, expression also effects the temporal property of speech \cite{kienast1999articulatory, kienast2000acoustical}. When a particular phoneme is uttered, its duration can be shortened (for example when angry and vowels particularly become shorter) or elongated, for example when a speaker is sad. 

To the best of our knowledge there is no systematic study which specifically investigates lipreading expressive speech. Rather, tasks focus on either, synthesizing expression in faces \cite{shaw2016expressive, hamza2004ibm, khatri2014facial} or expression recognition during speech \cite{happy2015automatic, yan2016sparse, zhang2014multimodal}.
 
Studies such as \cite{Blokland199897} on the effect of low video {\bf frame-rate} on human speech intelligibility during video communications, suggest that lower frame rates, if they are visible to the speaker, encourage humans to over-articulate to compensate for the reduced visual information available, akin to a visual Lombard effect. Accuracy is maximized when the same frame rate is used for both training and testing~\cite{saitoh2010study}. They further recommend that when the training data cannot be recorded at the same frame rate as the test data, then it is best if the training data has a higher frame rate (for feature extraction) than the test data. A further observation is that word classification rates vary in a non-linear fashion as the frame rate is reduced. 
 
When it comes to dependence of lipreading on {\bf video quality}, an investigation into the effects of compression artifacts, visual noise (simulated with white noise) and localization errors in training is presented in \cite{heckmann2003effects}, and in \cite{ACP:ACP371}. The authors undertake two experiments, of which the first includes some attention to spatial resolution (the number of pixels). However, here, resolution varies along with other parameters. Neither of these papers consider the simple removal of information from a smaller image compared to a larger one. A more systematic study of resolution can be found in \cite{bearicip} in which video of varying resolution is parameterized using AAMs \cite{AAMs}. This work shows that machines can lipread continuous speech with as little as two pixels per lip.

With regard to {\bf color}, it has been surprisingly under used. In \cite{kaucic1998accurate} algorithms are derived which contain three key components: shape models, motion models, and focused color feature detectors. In early works it was common to use colored lip-stick or markers to help track the lips (tracking remains challenging) but many authors convert the image to grayscale and use grayscale features.

{\bf Unit choice} refers to the question of whether to use phonemes, visemes, words or something else. Classifiers built on phonemes \cite{howell2013confusion}, visemes \cite{Hazen1027972}, and words \cite{shin2011real} have all been previously presented. Sometimes the unit choice is linked to the problem: word classifiers often use word units, whereas continuous speech has to use phonemes or visemes. It is essentially a trade-off since using phonemes means accepting that there will be units that do not appear on the lips (the words ``bad'', ``pad'', and ``mad'' are usually said to be visually indistinguishable) whereas using visemes leads to better unit accuracy but there is then the problem of homopheny (words that have identical visemic transcriptions but different spellings).
 One study has reviewed how the unit selection affects recognition in relation to the unit selection of the supporting language model \cite{bear2016decoding} and have shown that phoneme networks work best for both phoneme and viseme classifiers. However the practical reality is that many systems use visemes and there is need to resolve which choice of visemes works best. Comparative studies such as \cite{cappelletta2012phoneme} have attempted to compare some previous viseme sets but, these often only consider a few different sets rather than the gulf available.

 Lan \textit{et al.} present in \cite{improveVis} a comparison of different {\bf features} first presented in \cite{927467}. Revisited in \cite{Matthews_Baker_2004}, AAM features are produced as either model-based (using shape information) or pixel-based (using appearance information). In \cite{improveVis} Lan \textit{et al.} observed that state of the art AAM features with appearance parameters outperform other feature types like sieve features, 2D DCT, and eigen-lip features, suggesting appearance is more informative than shape. Also pixel methods benefit from image normalisation to remove shape and affine variation from region of interest (in this example, the mouth and lips). The method in \cite{improveVis} classified words with the an Audio-Visual dataset known as RMAV but recommended in future creating classifiers with viseme labels for lipreading, and advises that most information is from the inner of the mouth.
 
 Some works have attempted to adapt features to address different problems, such as motion described above. For example, in \cite{seymour2008comparison} the authors suggest altering HMM modeling to permit either frozen or occluded frames, and demonstrate that even low level jitter will significantly affect the quality of lip reading features. 
 
 When it comes to the choice of {\bf classifier technology} it is the norm that machine lipreading systems adapt methods from acoustic recognition. This not only follows from the observation that visual and acoustic speech have the same origins but also from the practical observation that language models are expensive to create and it makes sense to re-use the models across the two modalities. The conventional classifier process is 1) data preparation (an acoustic example is creating MFCC's \cite{zhu2000use}, whereas a visual example might be \cite{cappelletta2012phoneme}), 2) build Hidden Markov Model classifiers, and 3) feed the classification outputs through a language network to produce a transcript.   Like feature selection, the choice of classifier is affected by the problem in hand. An optimal audio recognizer will not guarantee optimal performance in an audio-visual, or visual only domain. In \cite{potamianos2004audio}, for example, it is noted that their audio-visual results should not be ``read across'' to lipreading.
 
 More modern deep learning techniques for lipreading are an alternative approach which require much more training data \cite{thangthai2015improving}. A key disadvantage of these methods is a lack of understanding about what exactly a neural network is learning in order for it to classify unseen gestures. So often the results from deep learning are good but the scientific insight can be poor. 
  Thus recent work has begun to demonstrate performance of different deep learning approaches with a variety of neural network architectures. Convolution neural networks (CNN) have been particularly prevalent for image classification (\cite{sharif2014cnn, yan2015hd}) and Long Short Term Memory networks (LSTM) are performing well on temporal problems (e.g. language modeling \cite{sundermeyer2012lstm} or, scene labeling \cite{byeon2015scene}). For lipreading, we have evidence that both of these achieve good recognition rates in end-to-end systems, in \cite{Chung2017} a CNN achieves 61.1\% top $1$ accuracy and in \cite{wand2016lipreading} an LSTM achieves 79.6\% top $1$ accuracy on a small dataset. However, our lipreading is a combination of these challenges, that is a temporal-visual classification problem.
 
For lipreading {\bf multiple persons}, \cite{simstruct, visualvowelpercept} detailed human lipreading of multiple people, \cite{simstruct} recognizes consonants, and \cite{visualvowelpercept} visual vowels. \cite{871073} presents an audio-visual system for HCI which automatically detects a talking person (both spatially and temporally) using video and audio data from a single microphone. In summary there is no reason to think that multi-person lipreading is any less viable than single-person lipreading, although the challenge of variability due to speaker identity is real.
 
{\bf Speaker identity} is a major challenge in machine lipreading because Visual speech is not consistent across individuals. Sometimes this can be advantageous as in \cite{607030} where they use lipreading to identify speakers. With known speakers - lipreading recognition rates can be high, but with unknown speakers (referred to as speaker-independent lipreading) this is as yet not at the same standard as speaker dependent lipreading. In \cite{bear2015speaker} results show that classifiers trained and tested on distinct speakers compared to those trained and tested on the same speakers are statistically significantly different. This is supported in \cite{newman2010speaker} where the authors strive to discriminate languages from visual speech and they conclude that in order to improve performance would be to move away from speaker-dependent features.  

For acoustic speech it is acknowledged that people have different speaking styles, accents and {\bf rates of speech}. For visual speech there is the additional confusion of what we call a ``visual accent'' in which very similar sounds can be made by persons with very different mouth shapes -- examples of visual accent effects include people who talk out of the side of their mouths; ventriloquists and mimics. The rate of speech alters both an utterance duration and articulator positions. Therefore, both the sounds produced, but particularly, visible appearance are altered. In \cite{6854158}, the authors present an experiment which measures the effect of speech rate and shows the effect is significantly higher on visual speech than in acoustic. Anecdotal evidence suggests that speaker visual style can evolve as speakers age due to co-articulation reduction as a person travels/interacts with other adults \cite{bear2016decoding}. 
 
In summary, while audio-visual speech processing has a great number of challenges, one of the pivotal ones is the question of the visual units and how they should be derived. Since all language models are defined in terms of phonemes, the practical question is the choice of the mapping from phonemes to visemes. The literature has presented a great number of these phoneme-to-viseme (P2V) mappings and few consistent comparisons between them so this is the topic for the next section.
 
\section{Comparison of phoneme-to-viseme mappings} 
A summary of published P2V maps is provided in \cite{theobaldPHD} Tables 2.3 and 2.4. This list is not exhaustive and these mappings motivated by: a focus on just consonants \cite{binnie1976visual, fisher1968confusions, franks1972confusion, walden1977effects}; being speaker-dependent \cite{kricos1982differences}, prioritizing particular visemes \cite{owens1985visemes}; or a focus on vowels \cite{disney, montgomery1983physical}. These are useful starting points, but for the purpose of this study we would like the phoneme-to-viseme mappings to include all phonemes in the transcript of the dataset to accurately reflect the range of phonemes used in a full vocabulary. Therefore, some mappings used here are a pairing of two mappings suggested in literature, e.g. one maps for the vowels and one map for the consonants. A full list of the mappings used is in Tables~\ref{tab:vowelmappings} and~\ref{tab:consonantmappings}.  Of these mappings , the most common are `the Disney 12' \cite{disney}, the `lipreading 18' by Nichie \cite{lip_reading18}, and Fisher's \cite{fisher1968confusions}.

\begin{table}[!h] 
\centering 
\caption{Vowel phoneme-to-viseme maps previously presented in literature.} 
\begin{tabular}{|l|l|} 
\hline 
Classification & Viseme phoneme sets \\ 
\hline \hline 
Bozkurt \cite{bozkurt2007comparison} & {\footnotesize \{/ei/ /\textturnv/\} \{/ei/ /e/ /\ae/\} \{/\textrevepsilon/\} \{/i/ /\textsci/ /\textschwa/ /y/\} \{/\textscripta\textupsilon/\} } \\ 
& {\footnotesize \{/\textopeno/ /\textscripta/ /\textopeno\textsci/ /\textschwa\textupsilon/\} \{/u/ /\textupsilon/ /w/\} }\\ 
Disney \cite{disney} & {\footnotesize \{/\textupsilon/ /h/\} \{/\textepsilon\textschwa/ /i/ /ai/ /e/ /\textturnv/\} \{/u/\} \{/\textupsilon\textschwa/ /\textopeno/ /\textopeno\textschwa/\} } \\ 
Hazen \cite{Hazen1027972} & {\footnotesize \{/\textscripta\textupsilon/ /\textupsilon/ /u/ /\textschwa\textupsilon/ /\textopeno/ /w/ /\textopeno\textsci/\} \{/\textturnv/ /\textscripta/\} \{/\ae/ /e/ /ai/ /ei/\} } \\ 
& {\footnotesize \{/\textschwa/ /\textsci/ /i/\} }\\ 
Jeffers \cite{jeffers1971speechreading} & {\footnotesize \{/\textscripta/ /\ae/ /\textturnv/ /ai/ /e/ /ei/ /\textsci/ /i/ /\textopeno/ /\textschwa/ /\textsci/\} \{/\textopeno\textsci/ /\textopeno/\} \{/\textscripta\textupsilon/\} }\\ 
& {\footnotesize \{/\textrevepsilon/ /\textschwa\textupsilon/ /\textupsilon/ /u/\} } \\ 
Lee \cite{lee2002audio} & {\footnotesize \{/i/ /\textsci/\} \{/e/ /ei/ /\ae/\} \{/\textscripta/ /\textscripta\textupsilon/ /ai/ /\textturnv/\} \{/\textopeno/ /\textopeno\textsci/ /\textschwa\textupsilon/\} \{/\textupsilon/ /u/\} }\\ 
Montgomery \cite{montgomery1983physical} & {\footnotesize \{/i/ /\textsci/\} \{/e/ /\ae/ /ei/ /ai/\} \{/\textscripta/ /\textopeno/ /\textturnv/\} \{/\textupsilon/ /\textrevepsilon/ /\textschwa/\}\{/\textopeno\textsci/\} } \\ 
& {\footnotesize \{/i/ /hh/\} \{/\textscripta\textupsilon/ /\textschwa\textupsilon/\} \{/u/ /u/\} }\\ 
Neti \cite{neti2000audio} & {\footnotesize \{/\textopeno/ /\textturnv/ /\textscripta/ /\textrevepsilon/ /\textopeno\textsci/ /\textscripta\textupsilon/ /\textipa{H}/\} \{/u/ /\textupsilon/ /\textschwa\textupsilon/\} \{/\ae/ /e/ /ei/ /ai/\} } \\ 
& {\footnotesize \{/\textsci/ /i/ /\textschwa/\} }\\ 
Nichie \cite{lip_reading18} & {\footnotesize \{/uw/\} \{/\textupsilon/ /\textschwa\textupsilon/\} \{/\textscripta\textupsilon/\} \{/i/ /\textturnv/ /ay/\} \{/\textturnv/\} \{/iy/ /\ae/\} \{/e/ /\textsci\textschwa/\} } \\ 
& {\footnotesize \{/u/\} \{/\textschwa/ /ei/\} }\\ 
\hline 
\end{tabular} 
\label{tab:vowelmappings} 
\end{table}

\begin{table}[!h] 
\centering 
\caption{Consonant phoneme-to-viseme maps previously presented in literature.} 
\resizebox{\columnwidth}{!}{%
\begin{tabular}{|l|l|} 
\hline 
Classification & Viseme phoneme sets \\ 
\hline \hline 
Binnie \cite{binnie1976visual} & {\footnotesize \{/p/ /b/ /m/\} \{/f/ /v/\} \{/\textipa{T}/ /\textipa{D}/\} \{/\textipa{S}/ /\textipa{Z}/\} \{/k/ /g/\} \{/w/\} \{/r/\} }\\ 
& {\footnotesize \{/l/ /n/\} \{/t/ /d/ /s/ /z/\} } \\ 
Bozkurt \cite{bozkurt2007comparison} & {\footnotesize \{/g/ /\textipa{H}/ /k/ /\textipa{N}/\} \{/l/ /d/ /n/ /t/\} \{/s/ /z/\} \{/t\textipa{S}/ /\textipa{S}/ /d\textipa{Z}/ /\textipa{Z}/\} \{/\textipa{T}/ /\textipa{D}/\} }\\ 
& {\footnotesize \{/r/\} \{/f/ /v/\} \{/p/ /b/ /m/\} }\\ 
Disney \cite{disney} & {\footnotesize \{/p/ /b/ /m/\} \{/w/\} \{/f/ /v/\} \{/\textipa{T}/\} \{/l/\} \{/d/ /t/ /z/ /s/ /r/ /n/\} }\\ 
& {\footnotesize \{/\textipa{S}/ /t\textipa{S}/ /j/\} \{/y/ /g/ /k/ /\textipa{N}/\} }\\ 
Finn \cite{finn1988automatic} & {\footnotesize \{/p/ /b/ /m/\} \{/\textipa{T}/ /\textipa{D}/\} \{/w/ /s/\} \{/k/ /h/ /g/\} \{/\textipa{S}/ /\textipa{Z}/ /t\textipa{S}/ /j/\} } \\ 
& {\footnotesize \{/y/\} \{/z/\} \{/f/\} \{/v/\} \{/t/ /d/ /n/ /l/ /r/\} } \\ 
Fisher \cite{fisher1968confusions} & {\footnotesize \{/k/ /g/ /\textipa{N}/ /m/\} \{/p/ /b/\} \{/f/ /v/\} \{/\textipa{S}/ /\textipa{Z}/ /d\textipa{Z}/ /t\textipa{S}/\} }\\ 
& {\footnotesize \{/t/ /d/ /n/ /\textipa{T}/ /\textipa{D}/ /z/ /s/ /r/ /l/\} } \\ 
Franks \cite{franks1972confusion} & {\footnotesize \{/p/ /b/ /m/\} \{/f/\} \{/r/ /w/\} \{/\textipa{S}/ /d\textipa{Z}/ /t\textipa{S}/\} }\\ 
Hazen \cite{Hazen1027972} & {\footnotesize \{/l/\} \{/r/\} \{/y/\} \{/b/ /p/\} \{m\} \{/s/ /z/ /h/\} \{/t\textipa{S}/ /d\textipa{Z}/ /\textipa{S}/ /\textipa{Z}/\} }\\ 
& {\footnotesize \{/t/ /d/ /\textipa{T}/ /\textipa{D}/ /g/ /k/\} \{/\textipa{N}/\} \{/f/ /v/\} }\\ 
Heider \cite{heider1940experimental} & {\footnotesize \{/p/ /b/ /m/\} \{/f/ /v/\} \{/k/ /g/\} \{/\textipa{S}/ /t\textipa{S}/ /d\textipa{Z}/\} \{/\textipa{T}/\} \{/n/ /t/ /d/\} } \\ 
& {\footnotesize \{/l/\} \{/r/\} } \\ 
Jeffers \cite{jeffers1971speechreading} & {\footnotesize \{/f/ /v/\} \{/r/ /q/ /w/\} \{/p/ /b/ /m/\} \{/\textipa{T}/ /\textipa{D}/\} \{/t\textipa{S}/ /d\textipa{Z}/ /\textipa{S}/ /\textipa{Z}/\} } \\ 
& {\footnotesize \{/s/ /z/\} \{/d/ /l/ /n/ /t/\} \{/g/ /k/ /\textipa{N}/\} }\\ 
Kricos \cite{kricos1982differences} & {\footnotesize \{/p/ /b/ /m/\} \{/f/ /v/\} \{/w/ /r/\} \{/t/ /d/ /s/ /z/\} } \\ 
& {\footnotesize \{/k/ /\textipa{n}/ /j/ /h/ /\textipa{N}/ /g/\} \{/l/\} \{/\textipa{T}/ /\textipa{D}/\} \{/\textipa{S}/ /\textipa{Z}/ /t\textipa{S}/ /d\textipa{Z}/\} }\\ 
Lee \cite{lee2002audio} & {\footnotesize \{/d/ /t/ /s/ /z/ /\textipa{T}/ /\textipa{D}/\} \{/g/ /k/ /n/ /\textipa{N}/ /l/ /y/ /\textipa{H}/\} \{/d\textipa{Z}/ /t\textipa{S}/ /\textipa{S}/ /\textipa{Z}/\} }\\ 
& {\footnotesize \{/r/ /w/\} \{/f/ /v/\} \{/p/ /b/ /m/\} }\\ 
Neti \cite{neti2000audio} & {\footnotesize\{/l/ /r/ /y/\} \{/s/ /z/\} \{/t/ /d/ /n/\} \{/\textipa{S}/ /\textipa{Z}/ /d\textipa{Z}/ /t\textipa{S}/\} \{/p/ /b/ /m/\} }\\ 
& {\footnotesize \{/\textipa{N}/ /k/ /g/ /w/\} \{/f/ /v/\} \{/\textipa{T}/ /\textipa{D}/\} }\\ 
Nichie \cite{lip_reading18} & {\footnotesize \{/p/ /b/ /m/\} \{/f/ /v/\} \{/\textipa{W}/ /w/\} \{/r/\} \{/s/ /z/\} \{/\textipa{S}/ /\textipa{Z}/ /t\textipa{S}/ /j/\} } \\ 
& {\footnotesize \{/\textipa{T}/\} \{/l/\} \{/k/ /g/ /\textipa{N}/\} \{/\textipa{H}/\} \{/t/ /d/ /n/\} \{/y/\} }\\ 
Walden \cite{walden1977effects} & {\footnotesize \{/p/ /b/ /m/\} \{/f/ /v/\} \{/\textipa{T} /\textipa{D}/\} \{/\textipa{S}/ /\textipa{Z}/\} \{/w/\} \{/s/ /z/\} \{/r/\} } \\ 
& {\footnotesize \{/l/\} \{/t/ /d/ /n/ /k/ /g/ /j/\} }\\ 
Woodward \cite{woodward1960phoneme} & {\footnotesize \{/p/ /b/ /m/\} \{/f/ /v/\} \{/w /r/ /\textipa{W}/\} } \\ 
& {\footnotesize \{/t/ /d/ /n/ /l/ /\textipa{T}/ /\textipa{D}/ /s/ /z/ /t\textipa{S}/ /d\textipa{Z}/ /\textipa{S}/ /\textipa{Z}/ /j/ /k/ /g/ /h/\}}\\ 
\hline 
\end{tabular} %
} 
\label{tab:consonantmappings} 
\end{table} 

In total, eight vowel- and fifteen consonant-maps are identified here and all of these are paired with each other to provide 120 P2V maps to test. 

Recent comparisons between maps include \cite{cappelletta2012phoneme} and as part of \cite{theobaldPHD}. In \cite{theobaldPHD} the following list of reasons are given for discrepancies between classifier sets. 
\begin{itemize} 
\item Variation between speakers - i.e. speaker identity. 
\item Variation between viewers - indicating lipreading ability varies by individuals, those with more practice are better able to identify visemes. 
\item The context of the speech presented - context has an influence on how consonants appear on the lips. In real tasks the context will enable easier distinction between indistinguishable phonemes in syllable only tests. 
\item Clustering criteria - the grouping methods vary between authors. For example, `phonemes are said to belong to a viseme if, when clustered, the percent correct identification for the viseme is above some threshold, which is typically between 70 - 75\% correct. A stricter grouping criterion has a higher threshold, so more visemes are identified.'\cite{theobaldPHD}. 
\end{itemize} 

These last two points are reinforced by \cite{cappelletta2012phoneme} who achieved highest accuracy with the phoneme-to-viseme map of Jeffers in an HMM-based lipreading system. They attribute this to the use of continuous speech which encapsulates the same viseme in more contexts within the training data, and suggest that the Jeffers map has better clustering of consonant visemes for those contexts. 


In Table~\ref{tab:lit_visemes_compare} we have described the sources and derivation methods for all of the phoneme-to-viseme maps used in our comparison study. We see the majority are constructed using human testing with few test subjects, for example Finn \cite{finn1988automatic} used only one lipreader, and Kricos \cite{kricos1982differences} twelve. Data-driven methods are most recent, e.g. Lee's \cite{lee2002audio} visemes were presented in 2002 and Hazen's \cite{Hazen1027972} in 2004. The remaining visemes are based around linguistic/phonemic rules. 
 
\begin{table}[!h] 
\centering 
\caption{A comparison of literature phoneme-to-viseme maps.} 
\resizebox{\columnwidth}{!}{%
\begin{tabular}{|l|l|l|l|l|} 
\hline 
Author & Year & Inspiration & Description & Test subjects \\ 
\hline \hline 
Binnie & 1976 & Human testing & Confusion patterns & unknown \\ 
Bozkurt & 2007 & Subjective linguistics & Common tri-phones & 462\\ 
Disney & --- & Speech synthesis & Observations & unknown \\ 
Finn & 1988 & Human perception & Montgomerys visemes & 1 \\ 
&		&			& and /\textipa{H}/ & \\ 
Fisher & 1986 & Human testing & Multiple-choice & 18 \\ 
&		&		&	intelligibility test & \\ 
Franks & 1972 & Human perception & Confusions among sounds & unknown\\ 
&		&		& produced in similar & \\ 
&		&		& articulatory positions & 275\\ 
Hazen & 2004 & Data-driven & Bottom-up clustering & 223 \\ 
Heider & 1940 & Human perception & Confusions post-training & unknown \\ 
Jeffers & 1971 & Linguistics & Sensory and cognitive & unknown \\ 
&		&		&	correlates & \\ 
Kricos & 1982 & Human testing & Hierarchical clustering & 12 \\ 
Lee & 2002 & Data-driven & Merging of Fisher visemes & unknown \\ 
Montgomery & 1983 & Human perception & Confusion patterns & 10 \\ 
Neti & 2000 & Linguistics & Decision tree clusters & 26 \\ 
Nichie & 1912 & Human observations & Human observation of & unknown \\ 
&		&				& lip movements & \\ 
Walden & 1977 & Human testing & Hierarchical clustering & 31 \\ 
Woodward & 1960 & Linguistics & Language rules & unknown \\ 
&		&		&	and context & \\ 
\hline 
\end{tabular} %
} 
\label{tab:lit_visemes_compare} 
\end{table} 
 
 As an example, the clustering method of Hazen \cite{Hazen1027972} involved bottom-up clustering using maximum Bhattacharyya distances \cite{bhattachayya1943measure} to measure similarity between the phoneme-labeled Gaussian models. Before clustering, some phonemes were manually merged, $/em/$ with $/m/$, $/en/$ with $/n/$, and $/\textipa{Z}/$ with $/\textipa{S}/$. 

\newpage
A P2V map may be summarized as a ratio we call ``compression factor,'' $CF_s$ 
\begin{equation} 
\centering 
CF_s = \frac{NV}{NP} 
\label{eq:cfequation} 
\end{equation} 
which is the ratio of number output visemes, $NV$ to input phonemes $NP$. The compression factors for the P2V maps are listed in Table~\ref{tab:Confusion_Factors}. Silence and garbage visemes are not included in Compression Factors. 
\begin{table}[!h] 
\centering 
\caption{Compression factors for viseme maps previously presented in literature.} 
\begin{tabular}{| l | r | r | l | r | r |} 
\hline 
Consonant Map & V:P & CF & Vowel Map & V:P & CF \\ 
\hline \hline 
Woodward & 4:24 & 0.16 & Jeffers & 3:19 & 0.16\\ 
Disney & 6:22 & 0.18 & Neti & 4:20 & 0.20\\ 
Fisher & 5:21 & 0.23 & Hazen & 4:18 & 0.22 \\ 
Lee & 6:24 & 0.25 & Disney & 4:11 & 0.36 \\ 
Franks & 5:17 & 0.29 & Lee & 5:14 & 0.36\\ 
Kricos & 8:24 & 0.33 & Bozkurt & 7:19 & 0.37\\ 
Jeffers & 8:23 & 0.35 & Montgomery & 8:19 & 0.42\\ 
Neti & 8:23 & 0.35 & Nichie & 9:15 & 0.60\\ 
Bozkurt & 8:22 & 0.36 & - & - & -\\ 
Finn & 10:23 & 0.43 & - & - & -\\ 
Walden & 9:20 & 0.45 & - & - & -\\ 
Binnie & 9:19 & 0.47 & - & - & -\\ 
Hazen & 10:21 & 0.48 & - & - & -\\ 
Heider & 8:16 & 0.50 & - & - & -\\ 
Nichie & 18:33 & 0.54 & - & - & -\\ 
 
\hline 
\end{tabular} 
\label{tab:Confusion_Factors} 
\end{table}
 
Because we have a British English dataset and some works were formulated using American English diacritics \cite{diacritics} we omit the following phonemes from some mappings: $/si/$ (Disney \cite{disney}), $/axr/$ $/en/$ $/el/$ $/em/$ (Bozkirt \cite{bozkurt2007comparison}), $/axr/$ $/em/$ $/epi/$ $/tcl/$ $/dcl/$ $/en/$ $/gcl/$ $kcl/$(Hazen \cite{Hazen1027972}), and $/axr/$ $/em/$ $/el/$ $/nx/$ $/en/$ $/dx/$ $/eng/$ $/ux/$ (Jeffers \cite{jeffers1971speechreading}). Moreover, Kricos provides speaker-dependent visemes \cite{kricos1982differences}. These have been generalized for our tests using the most common mixtures of phonemes. Where a viseme map does not include phonemes present in the ground truth transcript these are grouped into one viseme denoted ($/gar/$). Note that all phonemes in each P2V map are in the dataset but no mapping includes all 29 phonemes in the AVL2 vocabulary. 

\subsection{Data preparation} 
\label{sec:currentp2vmaps} 
The AVLetters2 (AVL2) dataset \cite{cox2008challenge} is used to train and test HMM classifiers based upon our 120 P2V mappings with HTK \cite{htk34}. AAM features (concatenated as in (\ref{equation:app})) are used as they are known to outperform other feature methods in machine lipreading~\cite{cappelletta2012phoneme}. 
AVL2 \cite{cox2008challenge} is an HD version of the AVLetters dataset \cite{matthews1998nonlinear}. It is a single word dataset of five male British English speakers reciting the alphabet seven times. We use four of these speakers at the fifth tracked too poorly to have confidence in lipreading accuracy. The speakers in this dataset are illustrated in~\cite{yogiThesis}. AVL2 has 28 videos of between $1,169$ and $1,499$ frames between $47s$ and $58s$ in duration. As the dataset provides isolated words of single letters, it lends itself to controlled experiments without needing to address matters such as varying co-articulation. 

\begin{table}[!h] 
\centering 
\caption{The number of parameters in shape, appearance and combined shape \& appearance AAM features for each speaker in the AVLetters2 dataset for each speaker. Features retain 95\% variance of facial information.} 
\begin{tabular}{| l | r | r | r |} 
\hline 
Speaker	& Shape & Appearance & Combined \\ 
\hline \hline 
S1	& 11 & 27	& 38 \\ 
S2 	& 9 & 19 & 28 \\ 
S3 	& 9 & 17 & 25 \\ 
S4	& 9 & 17 & 25 \\ 
\hline 
\end{tabular} 
\label{tab:feature_parameters} 
\end{table} 

Table~\ref{tab:feature_parameters} describes the features extracted from the AVL2 videos. These features have been derived after tracking a full-face Active Appearance Model throughout the video before extracting features containing only the lip area. Therefore, they contain information representing only the speaker's lips and none of the rest of the face. Speakers 2, 3 and 4 are similar in number of parameters contained in the features. The combined features are the concatenation of the shape and appearance features \cite{Matthews_Baker_2004}. All features retain 95\% variance of facial shape and appearance information. 

\begin{figure}[!h] 
\centering 
\includegraphics[width=0.85\textwidth]{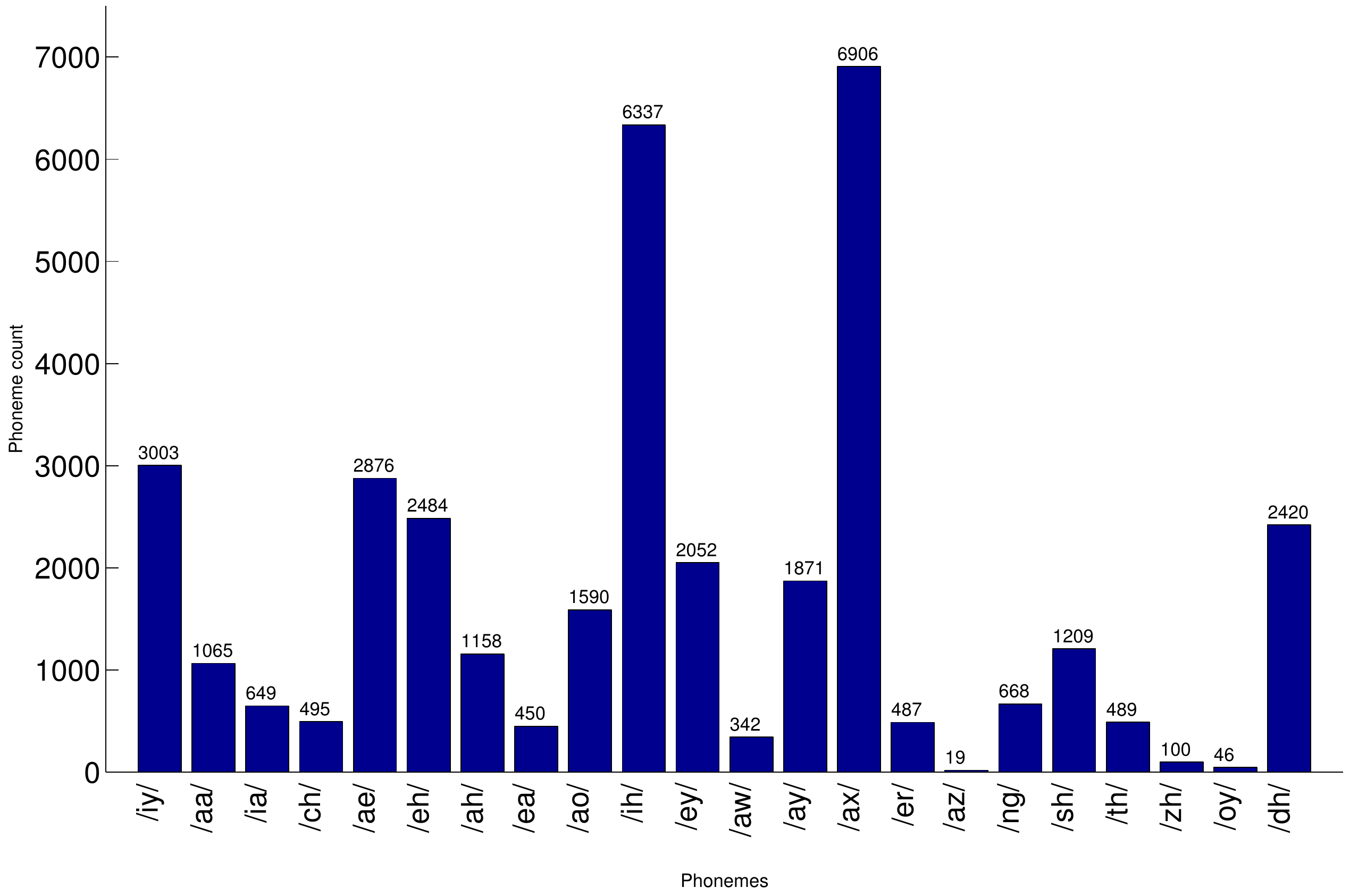} \\ 
\includegraphics[width=0.85\textwidth]{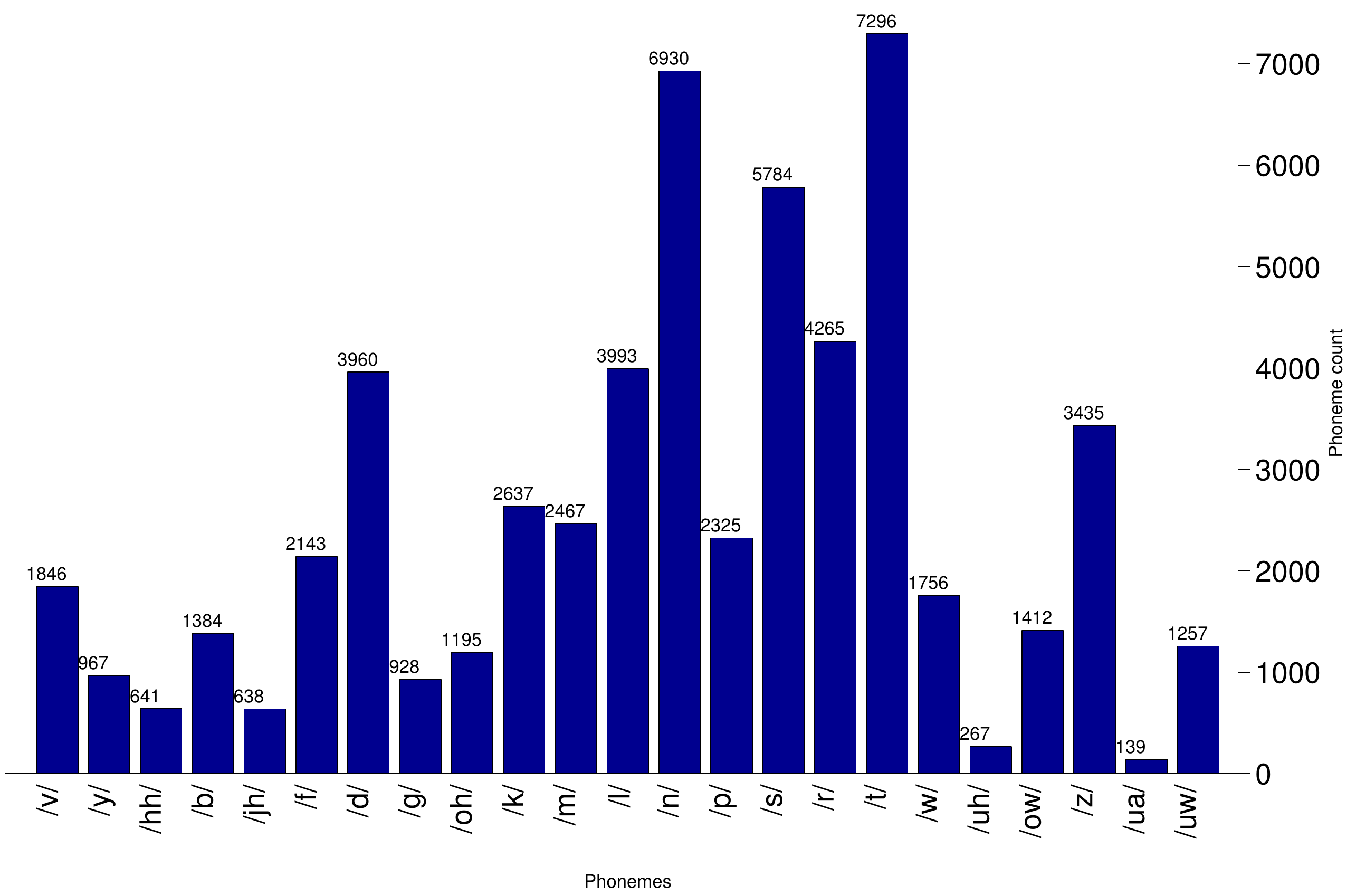} 
\caption{Occurrence frequency of phonemes in the RMAV dataset.} 
\label{fig:histogram} 
\end{figure} 

The RMAV dataset consists of 20 British English speakers (we use 12 speakers,seven male and five female, who have been tracked to maintain comparability with earlier work), 200 utterances per speaker of a subset of the Resource Management (RM) context independent sentences from \cite{fisher1986darpa} which totals around 1000 words each. The sentences are selected to maintain a good coverage all phonemes \cite{lan2010improving} and to represent the coverage of phonemes in spoken speech.  The original videos were recorded in high definition and in a full-frontal position. Individual speakers are tracked using Active Appearance Models \cite{Matthews_Baker_2004} and AAM features of concatenated shape and appearance information have been extracted. 

Figure~\ref{fig:histogram} plots the frequency of all phonemes within the RMAV dataset over 200 sentences and Table~\ref{tab:parameterslilirfeatures} lists the number of parameters of shape, appearance, and combined shape and appearance AAM features where the features retain 95\% variance of facial information. 
 
\begin{table}[!h] 
\centering 
\caption{The number of parameters of shape, appearance, and combined shape and appearance AAM features for the RMAV dataset speakers. Features retain 95\% variance of facial information.} 
\begin{tabular}{| l | r | r | r |} 
\hline 
Speaker	& Shape & Appearance & Combined \\ 
\hline \hline 
S1	& 13 & 46	& 59 \\ 
S2 	& 13 & 47 & 60 \\ 
S3	& 13 & 43	& 56 \\ 
S4	& 13 & 47	& 60 \\ 
S5 	& 13 & 45 & 58 \\ 
S6	& 13 & 47	& 60 \\ 
S7 	& 13 & 37 & 50 \\ 
S8 	& 13 & 46 & 59 \\ 
S9	& 13 & 45	& 58 \\ 
S10 	& 13 & 45 & 58 \\ 
S11 	& 14 & 72 & 86 \\ 
S12	& 13 & 45	& 58 \\ 
\hline 
\end{tabular} 
\label{tab:parameterslilirfeatures} 
\end{table} 
 
\subsection{Classification method} 
\label{sec:classification} 
The method for these speaker-dependent classification tests on our combined shape and appearance features uses HMM classifiers built with HTK \cite{htk34}. The features selected are from the AVL2 and RMAV datasets. The videos are tracked with a full-face AAM (Figure~\ref{fig:egshape} (left)) and the features extracted consist of only the lip information (Figure~\ref{fig:egshape} (right)). This means that we obtain a robust tracking from the full-face model, then using this fit information, we apply a sub-active appearance model of only the lips. The HMM classifiers are based upon viseme labels within each P2V map. A ground truth for measuring correct classification is a viseme transcription produced using the BEEP British English pronunciation dictionary \cite{beep} and a word transcription. The phonetic transcript is converted to a viseme transcript assuming the visemes in the mapping being tested (Tables~\ref{tab:consonantmappings} and~\ref{tab:vowelmappings}). We test using a leave-one-out seven-fold cross validation. Seven folds are selected as we have seven utterances of the alphabet per speaker in AVL2, this is increased to 10-fold cross-validation for RMAV speakers. The HMMs are initialized using `flat start' training and re-estimated eight times and then force-aligned using HTK's \texttt{HVite}. Training is completed by re-estimating the HMMs three more times with the force-aligned transcript. 
 
\subsection{Active appearance models} 
\begin{figure}[!h] 
\centering 
\begin{tabular}{l r}
\includegraphics[width=0.25\textwidth]{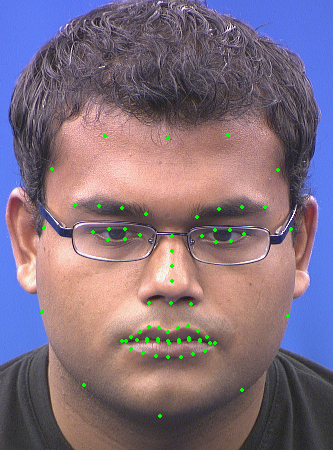} &
\includegraphics[width=0.25\textwidth]{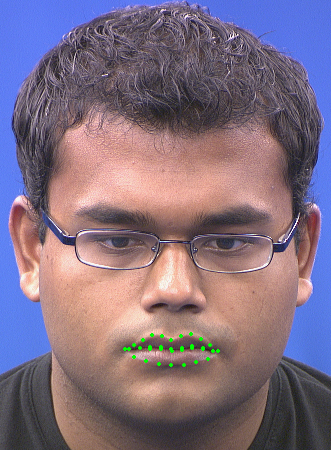}
\end{tabular}
\caption{Example Active Appearance Model shape mesh (left), a lips only model is on the right.} 
\label{fig:egshape} 
\end{figure}

An example full-face shape model example is in Figure~\ref{fig:egshape} where there are $76$ landmarks, $34$ of which are modeling the inner and outer lip contours.

The shape $s$ of an AAM is the collection of coordinates of the $v$ vertices (landmarks) which make up a mesh, 
 \begin{equation} 
s = (x_1,y_1, x_2,y_2, ..., x_v,y_v)^T 
\label{eq:meshvertices} 
\end{equation} 
These landmarks are aligned and normalized via Procrustes analysis \cite{procrustes} and then analyzed via a Principal Component Analysis (PCA) to 
\begin{equation} 
s = s_0 + \sum_{i=1}^np_is_i 
\label{equation:Shape} 
\end{equation} 
 
where $s_0$ is the mean shape, $p_i$ are coefficient shape parameters, and $s_i$ are the eigenvectors of the co-variance matrix of the $n$ largest eigenvalues \cite{Matthews_Baker_2004}. 

Having built an Active Shape Model, the next step is to augment it with appearance data and hence compute an Active Appearance Model (AAM). Each shape model is used to warp the image data back to the mean shape. The appearance of those warped images is now modeled again using PCA~\cite{927467},
\begin{equation} 
A(x) = A_0(x) + \sum_{i=1}^m\lambda_iA_i(x) \quad \forall x \in s_0 
\label{equation:app} 
\end{equation} 
where $\lambda_i$ are the appearance parameters, $A_0$ is the shape-free-mean appearance, and $A_i(x)$ are the appearance image eigenvectors of the co-variance matrix.

Usually the best results are obtained using both shape and appearance information combined within a single AAM~\cite{982900,927467}. Therefore, unless explicitly stated otherwise, we use these. Once an AAM is built and trained, we fit the model using the Inverse Compositional algorithm~\cite{inversecompAlg} to all frames in the video sequence~\cite{Matthews_Baker_2004}. 
 
\subsection{Comparison of current phoneme-to-viseme maps} 
\label{sec:comparison} 
Recognition performance of the HMMs can be measured by both correctness, $C$, and accuracy, $A$, 
\begin{multicols}{2}
 	\begin{equation} 
		C = \displaystyle \frac{N-D-S}{N} 
		\label{eq:correctness} 
	\end{equation}\break
 	\begin{equation} 
		A = \displaystyle \frac{N-D-S-I}{N} 
		\label{eq:accuracy} 
	\end{equation} 
\end{multicols}

where $S$ is the number of substitution errors, $D$ is the number of deletion errors, $I$ is the number of insertion errors and $N$ the total number of labels in the reference transcriptions~\cite{htk34}. An insertion error (which are notoriously common in lip reading \cite{hazen2006automatic}) occurs when the recognizer output has extra words/visemes missing from the original transcript \cite{htk34}. As an example one could say ``Once upon a midnight dreary'', but the recognizer outputs ``Once upon upon midnight dreary dreary". Here the recognizer has inserted two words which were never present and has deleted one\footnote{Once this utterance has been translated to one of viseme labels rather than words, as an example using Montgomery's visemes, this sentence becomes ``v09 v12 v04 v05 - v12 v01 v12 v04 - v12 - v01 v10 v04 v11 v04 - v04 v07 v16 v07 v16'' (hyphens are included to show breaks between words). In this case, the same insertion errors would create predicted outputs of ``v09 v12 v04 v05 - v12 v01 v12 v04 - v12 v01 v12 v04 - v01 v10 v04 v11 v04 - v04 v07 v16 v07 v16 - v04 v07 v16 v07 v16.''}.

In this experiment, classification performance of the HMMs is measured by correctness, $C$ (\ref{eq:correctness}), as there are no insertion errors to consider \cite{htk34}. It is acknowledged that word classification is not as high performing as viseme classification. However, as each viseme set being tested has a different number of phonemes and visemes, words, are used so we can compare different viseme sets. It is the difference between each set, rather than the individual performance, which is of interest in this investigation. 

\begin{figure}[!h]
\centering 
\begin{tabular}{c}
\includegraphics[width=1 \textwidth]{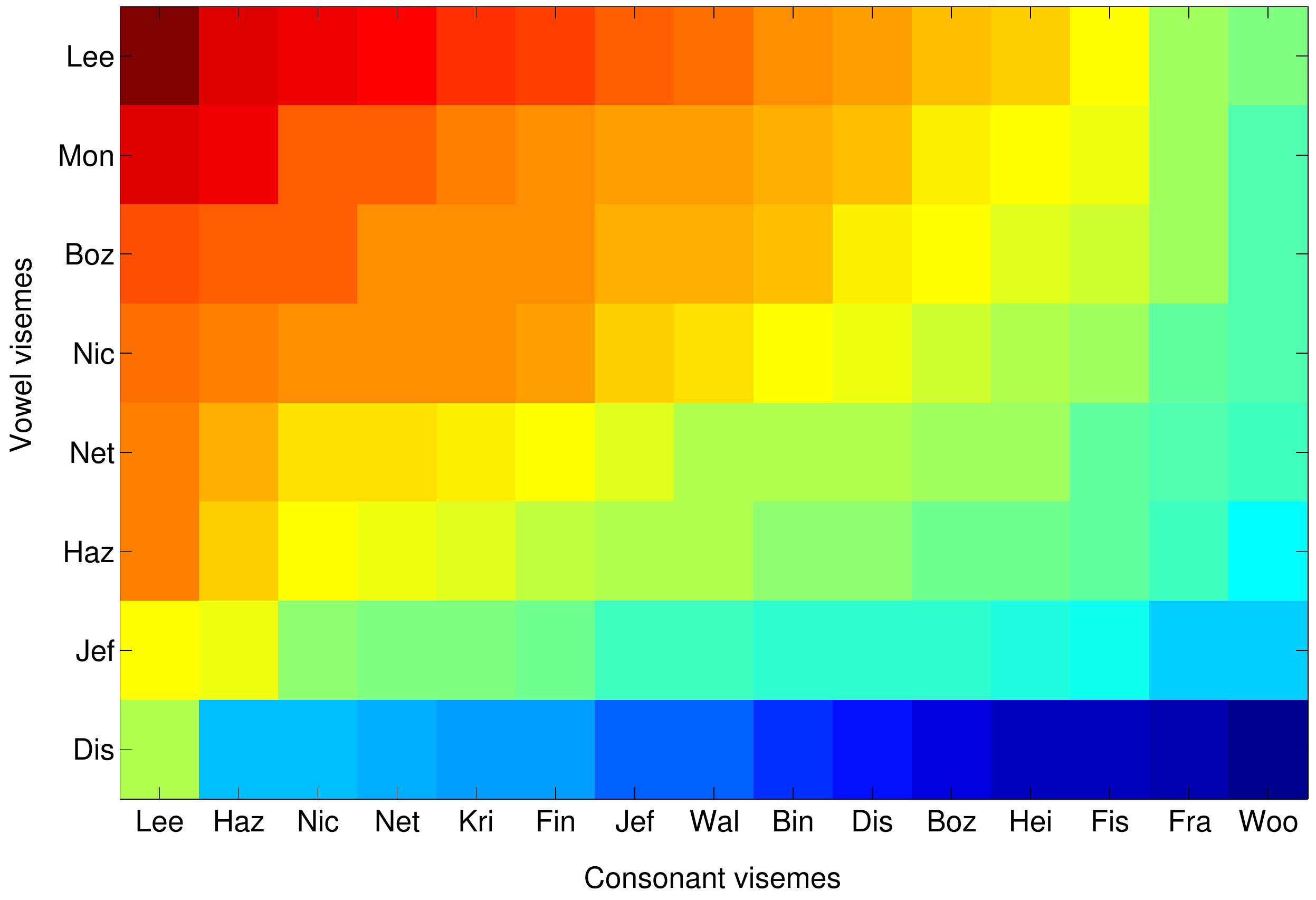} \\
\includegraphics[width=1 \textwidth]{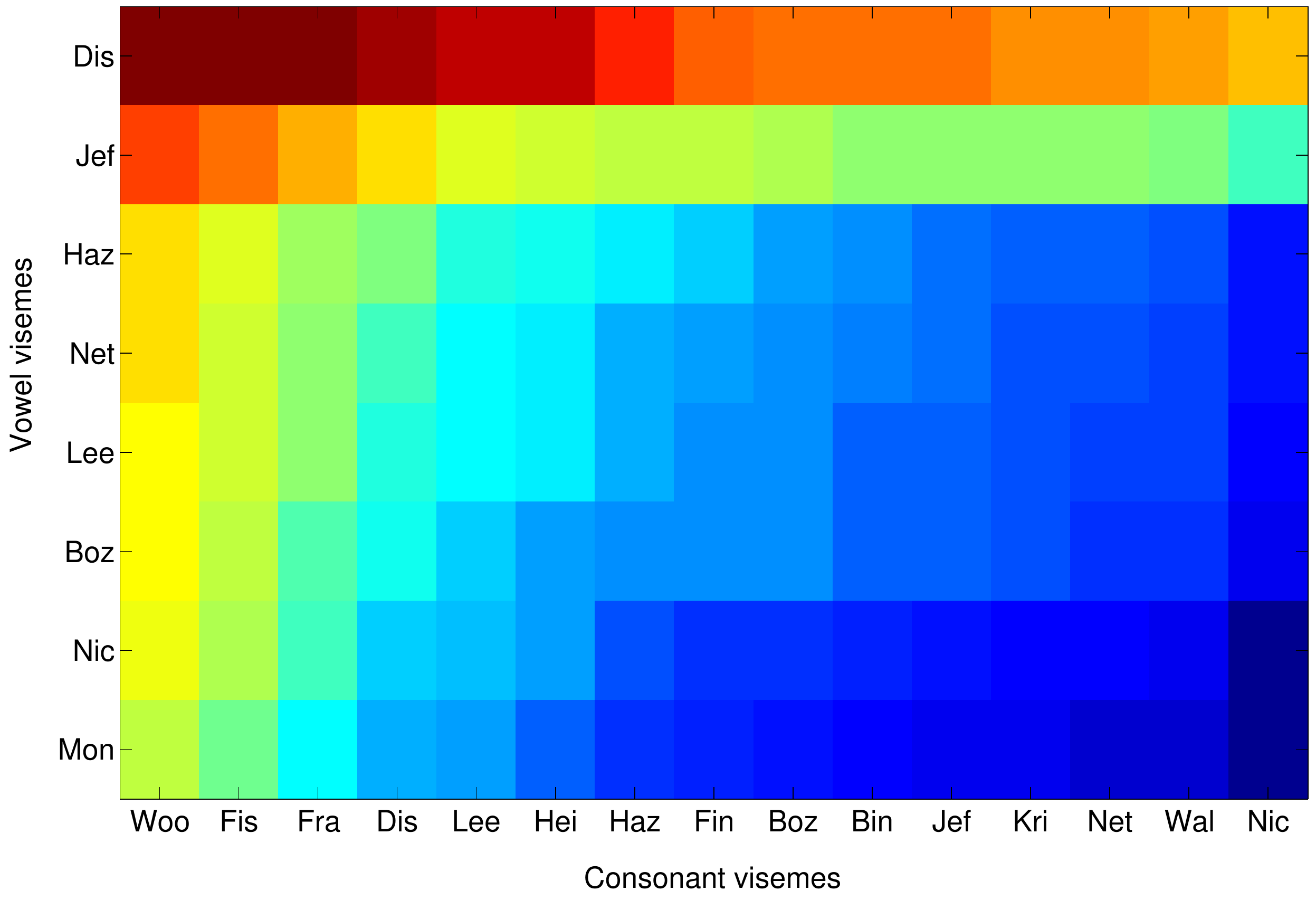} \\
\end{tabular}	
\caption{Speaker-dependent all-speaker mean word classification, $C$, comparing viseme classes on isolated word speech (top) and continuous speech (bottom)}
\label{fig:cs_heatmap}
\end{figure}
Figure~\ref{fig:cs_heatmap} shows the correctness of each pair of viseme sets. On the top is the isolated word case (the AVL2 data) and on the bottom the continuous data (RMAV). Each diagram is ordered by the mean correctness over all speakers. For the isolated words the Lee vowel and consonant sets \cite{lee2002audio} are the best with the Montgomery vowels \cite{montgomery1983physical} and Hazen consonants \cite{Hazen1027972} close behind. The worst performers are Disney vowels \cite{disney} and the Franks \cite{franks1972confusion} and Woodward consonants \cite{woodward1960phoneme}. For continuous speech the Disney vowels are the best performer~\cite{disney} as are the Woodward consonants~\cite{woodward1960phoneme}. It is notable that for continuous speech the high compression factor visemes sets work better than those with larger numbers of visemes. The most likely explanation is that continuous speech has additional variability due to co-articulation so a few coarsely defined visemes are better than a greater number of finely defined ones.

 \begin{figure}[!h]
\centering 
\begin{tabular}{c}
\includegraphics[width=.9 \textwidth]{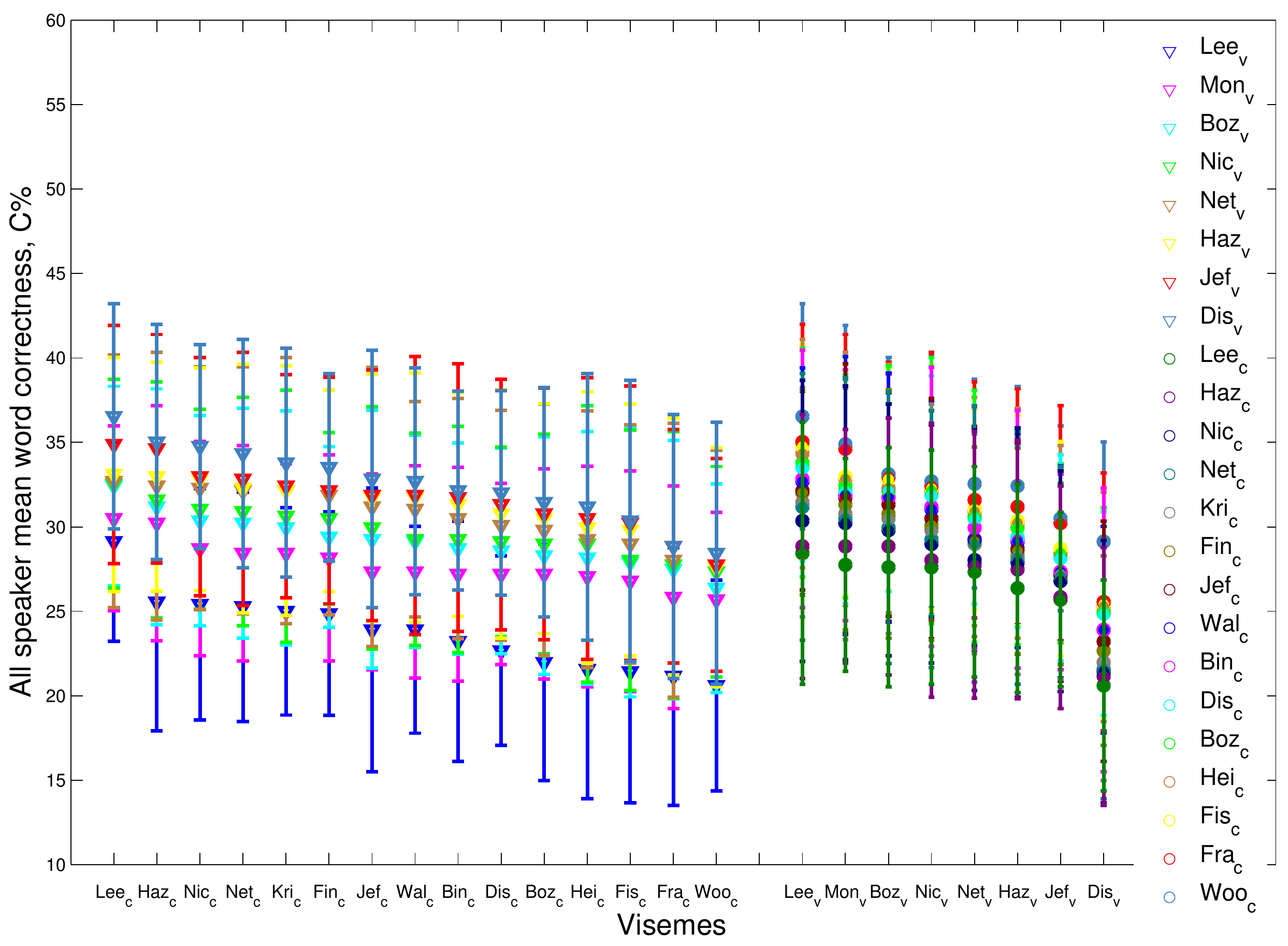} \\
\includegraphics[width=.9 \textwidth]{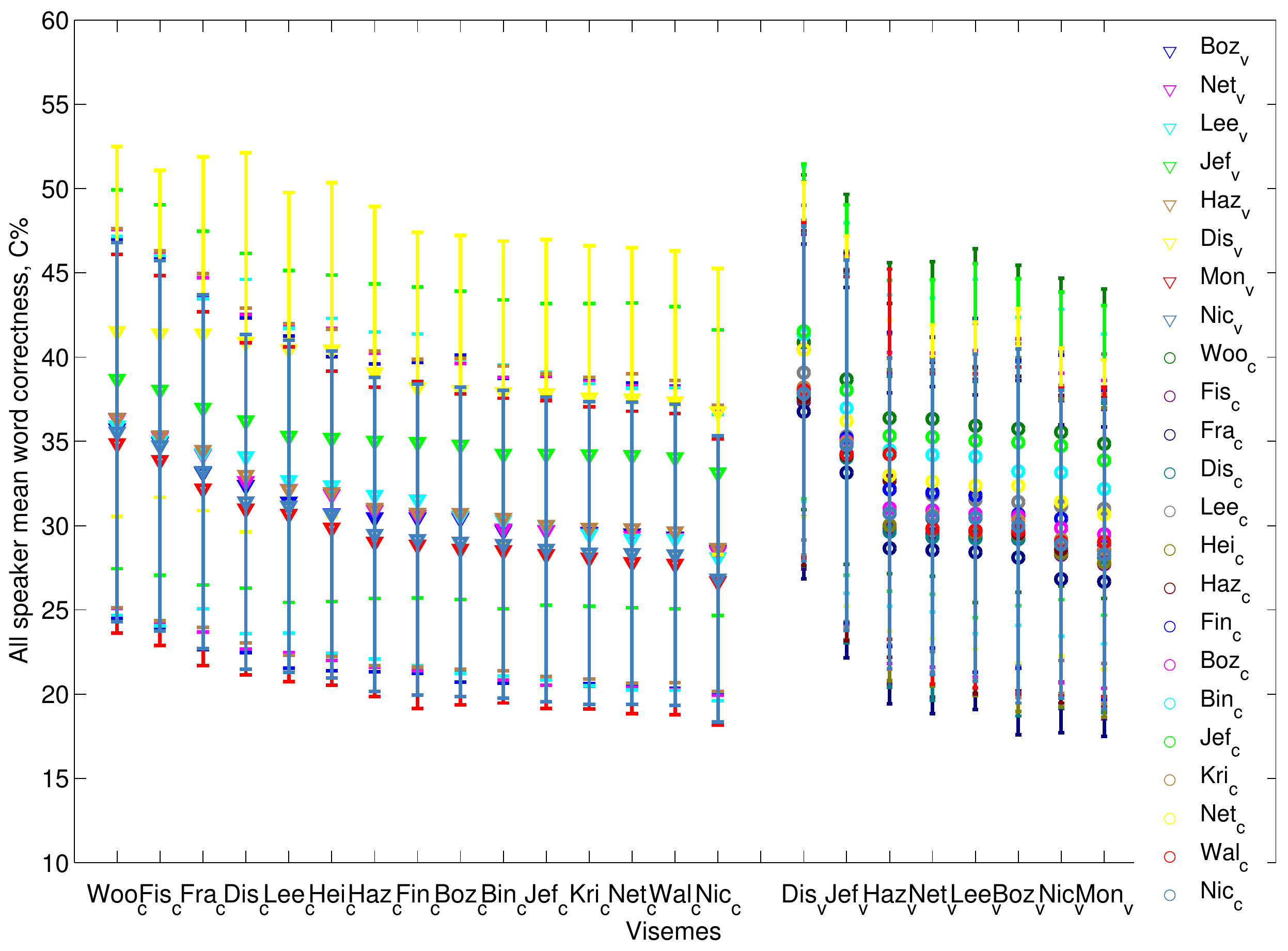} \\
\end{tabular}	
\caption{Speaker-independent all-speaker mean word classification, $C\pm1 s.e$. For a given mapping ($x-$axis) the performance is measured after pairing with all vowel mappings (left) and vice versa on the right on AVL2 isolated words (top) and RMAV continuous (bottom)}
\label{fig:all_avl2}
\end{figure}
Figure~\ref{fig:all_avl2} shows the mean word correctness, $C$, over all speakers, $\pm1 s.e$ for pairings of vowel and consonant maps ordered by correctness from left to right. Again, isolated word results (the AVL2 data) at the top and continuous (RMAV) on the bottom.
As previously, for isolated words, the Disney vowels are significantly worse than all others when paired with all consonant difference over the whole group. The Lee \cite{lee2002audio}, Montgomery~\cite{montgomery1983physical} and Bozkurt \cite{bozkurt2007comparison} vowels are consistently above the mean and above the upper error bar for Disney \cite{disney}, Jeffers \cite{jeffers1971speechreading} and Hazen \cite{Hazen1027972} vowels. In comparing the consonants, Lee \cite{lee2002audio} and Hazen \cite{Hazen1027972} are the best whereas Woodward \cite{woodward1960phoneme} and Franks \cite{franks1972confusion} are the bottom performers. There is a significant difference between the `best' visemes for individual speakers which arises from the unique way in which everyone articulates their speech. 

The continuous speech experiment results in Figure~\ref{fig:all_avl2} (bottom) show that, for vowel visemes, the Disney set surpasses all others, whereas Woodward's consonants are now a better fit. This is interesting as neither viseme set are data-derived. We recall that Disney's \cite{disney} are designed from human perception for synthesis of characters, and Woodward's \cite{woodward1960phoneme} are from a pilot investigation into phoneme perception in lipreading using linguistic rules.   As we move to more realistic data , continuous speech, many of the data-driven approaches  degrade which implies that they data used to derive these visemes was unrealistic.  For example the Lee visemes~\cite{lee2002audio} were derived without any use of video data at all so it is hardly surprising that they are fragile when presented with more realistic data.
 
The idea that vowel and consonant visemes should be treated differently is no surprise. The suggestion that vowel visemes are essentially mouth shapes and the consonants govern how we move in and out of them was first presented by Nichie in 1912 from human observations by a profoundly deaf educator \cite{lip_reading18} and is supported by results in \cite{bear2014phoneme} which show we should not mix vowel and consonant visemes for best results. Therefore, it is reassuring to see that the better speaker-independent phoneme-to-viseme mapping for continuous speech is a combination of two previous maps, where the two maps have differing derivation methods; perception and language rules. 

Generally speaking the continuous case (bottom of Figure~\ref{fig:all_avl2}) gives improved accuracies compared to the isolated word case (top of Figure~\ref{fig:all_avl2}. The first response to explain this is to suggest the increase is caused by better training of classifiers with the greater volume of training samples in RMAV than in AVL2. However, we should note that this effect is marginally countered by the co-articulation effects in continuous speech, so a set of classifiers trained on a larger isolated word dataset and compared to AVL2 would provide a greater increase in recognition. 

\begin{figure}[ht] 
\centering 
\begin{tabular}{l c c}
& Consonants & Vowels \\
\begin{sideways} Isolated words \end{sideways} &	\includegraphics[width=.5\textwidth]{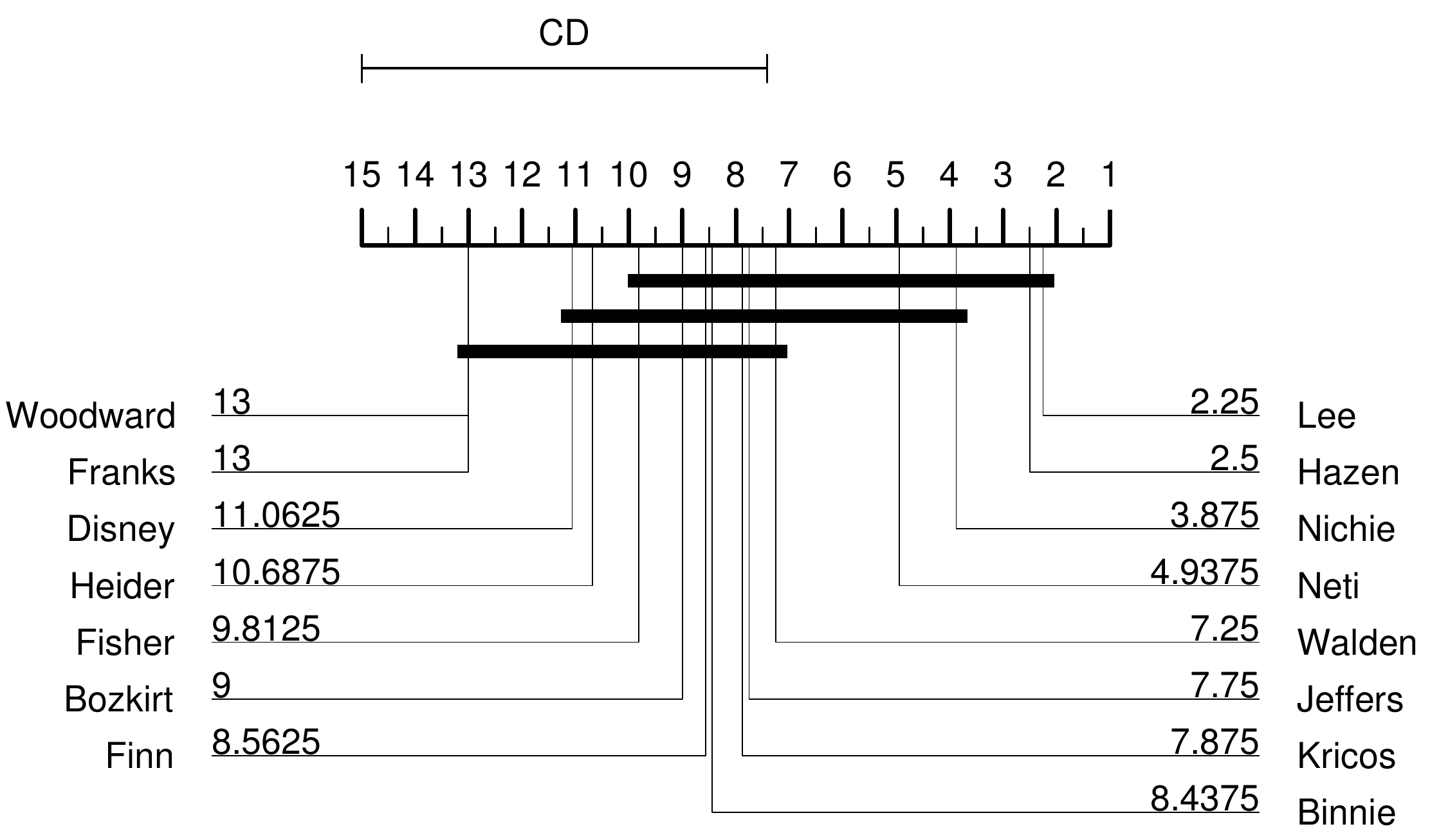} &
	\includegraphics[width=.5\textwidth]{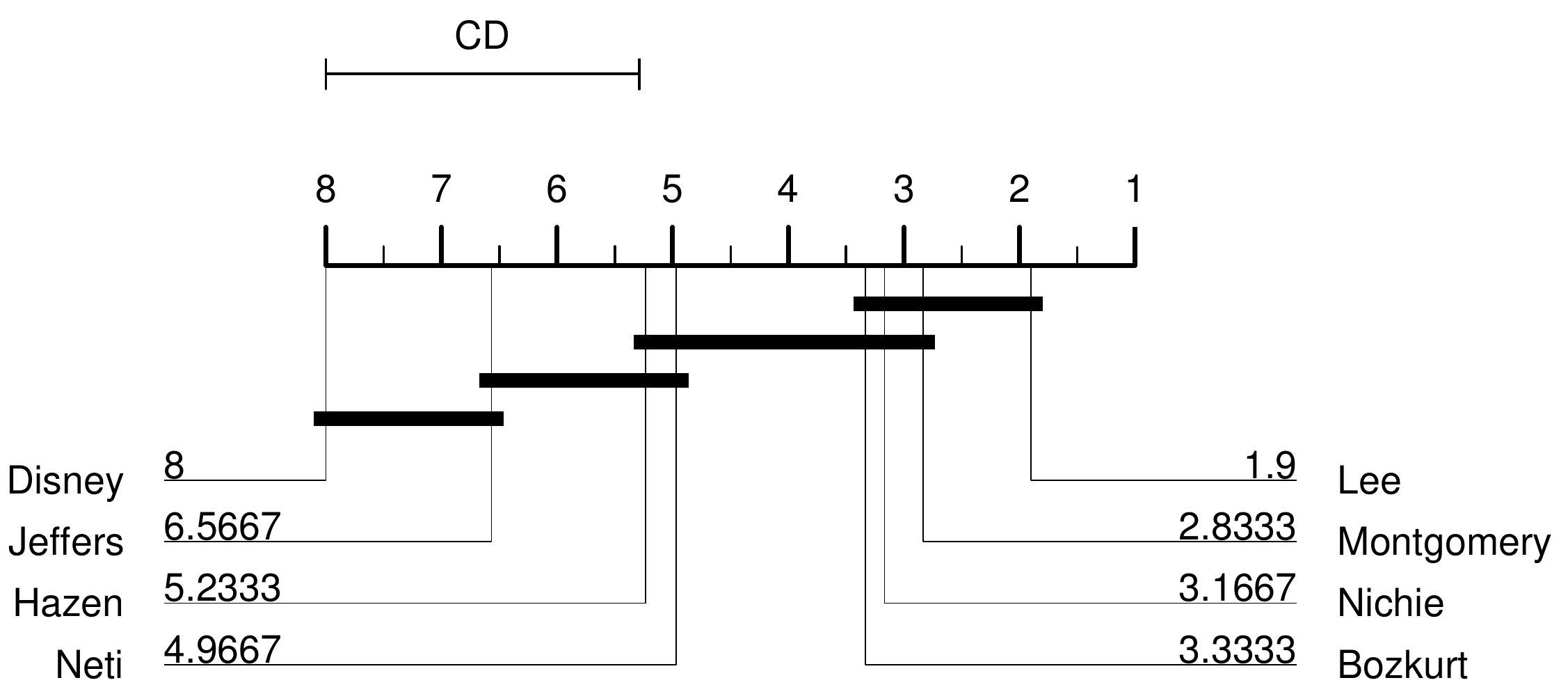} \\
\begin{sideways} Continuous speech \end{sideways} & \includegraphics[width=.5\textwidth]{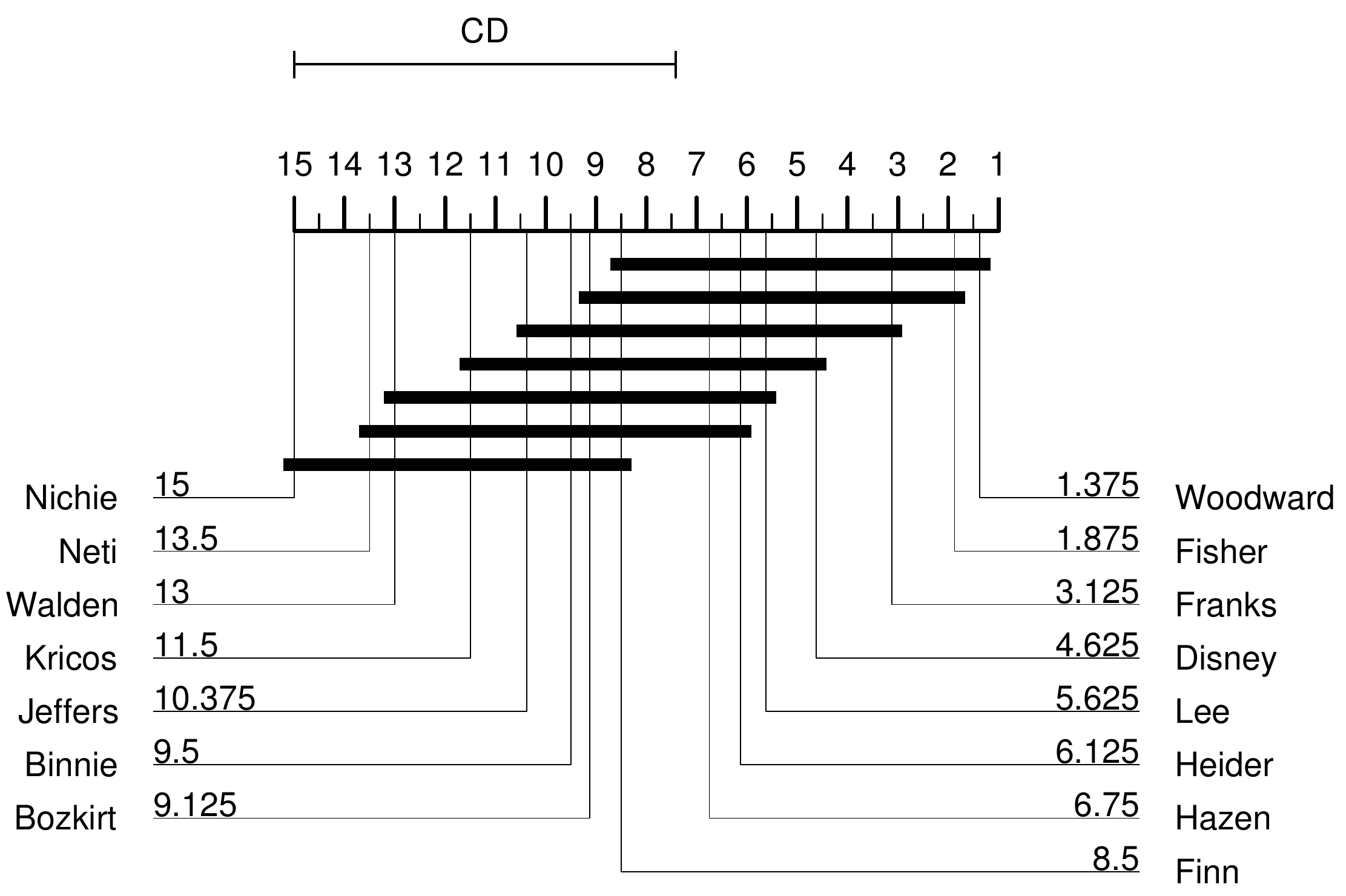} &
	\includegraphics[width=.5\textwidth]{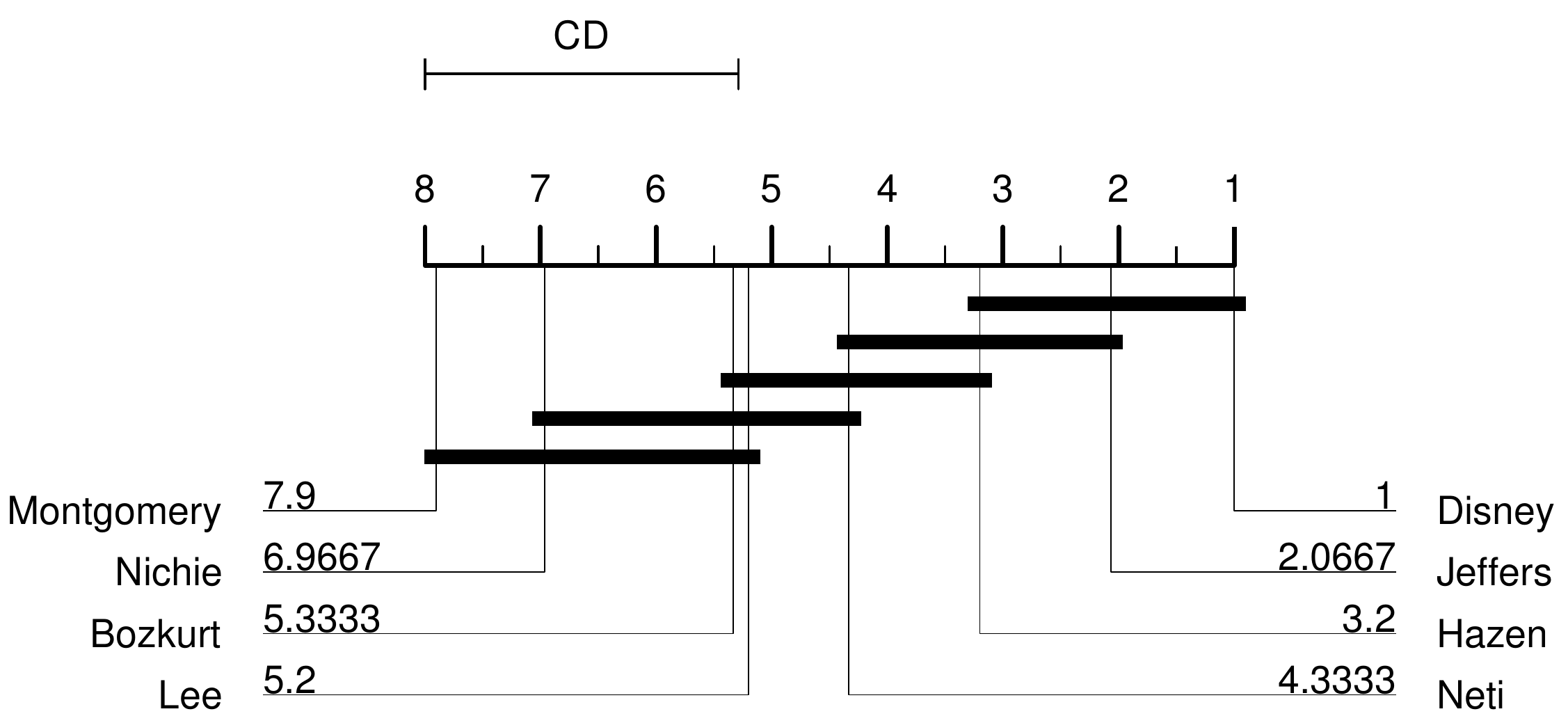} 
\end{tabular}
\caption{Critical difference of all phoneme-to-viseme maps independent of phoneme-to-viseme pair partner. Vowel maps are on the left side, consonants on the right. Isolated words are in the top row, and continuous speech along the bottom row.}
\label{fig:crit_diff}
\end{figure} 
Figure~\ref{fig:crit_diff} are critical difference plots between the viseme class sets based upon their classification performance \cite{criticaldiff} with isolated word training. Critical difference is a measure of the confidence intervals between different machine learning algorithms derived from Friedman tests on the ranked scores (here $p=0.05$). Two assumptions within critical difference are: all measured results are `reliable', and all algorithms are evaluated using the same random samples \cite{criticaldiff}. As we use the HTK standard metrics \cite{young2006htk}, and use results with consistent random sampling across folds, these assumptions are not a concern. We have selected critical differences here as these evaluate the performance of multiple classifiers on different datasets, whereas such as \cite{bouckaert2004evaluating, bengio2004no}, often require paired data or identical datasets.

Figure~\ref{fig:crit_diff} shows a significant difference between some sub-sets of visemes. This is shown by the horizontal bars which do not overlap all viseme sets. Where the horizontal bars do overlap, this shows the viseme sets are indistinguishable at a 95\% confidence.   When comparing isolated words with continuous speech we see  fewer significant differences with continuous speech despite there being more test data.

Table~\ref{tab:viseme_set_changes} summarises the best-performing visemes (consonant and vowels) for the isolated and continuous word data.  The first column  shows that the Lee consonants are the best performing for isolated words.  But also that  Hazan, Nichie, Neti etc are indistinguishable from Lee (they within  Lee's critical difference).  For  continuous speech, the Woodward consonant visemes are the best but Fisher,  Franks Disney etc  are indistinguisable.  In bold  are the  viseme sets that are common to both  isolated words and continuous speech:  Lee, Hazen, Finn and Fisher.  For  the vowels (second column) there are no common sets.  However  if  we look at best and second-best (the third column of Table~\ref{tab:viseme_set_changes}) then  Hazen and Neti emerge as common. 
\begin{table}[!h]
\centering 
\caption{Critically different viseme sets changes with isolated word and continuous speech data. Sets are listed in the order they appear in Figure~\ref{fig:crit_diff}.} 
\resizebox{\columnwidth}{!}{%
\begin{tabular} {| l | l | l |} 
\hline 
First Position Consonants & First Position Vowels & Second Position Vowels \\
\hline \hline
\textbf{Lee} & Lee & Montgomery \\
\textbf{Hazen} & Montgomery & Nichie \\
Nichie & Nichie & Bozkurt \\
Neti & Bozkurt & \textbf{Hazen} \\
Walden & & \textbf{Neti} \\
Jeffers & & \\
Kricos & & \\
Binnie  & & \\
\textbf{Finn} & & \\
Bozkurt  & & \\
\textbf{Fisher}  & & \\
\hline 
Woodward & Disney & Jeffers \\
\textbf{Fisher} & Jeffers & \textbf{Hazen} \\
Franks & Hazen & \textbf{Neti} \\
Disney  & & \\
\textbf{Lee}  & & \\
Heider  & & \\
\textbf{Hazen}  & & \\
\textbf{Finn}  & & \\
\hline
\end{tabular} %
}
\label{tab:viseme_set_changes} 
\end{table}
Looking across all sets the common method that performs near the top  is that due to Hazen~\cite{Hazen1027972}.  Interestingly these visemes were derived using the most realistic data  (an audio-visual corpus based on TIMIT)  and formed by a tree-based clustering of  phoneme-trained HMMs.   Note that the Hazan visemes were derived from American English data whereas here we use British English speakers.

\begin{figure}[!h] 
\centering 
\begin{tabular}{c}
	\includegraphics[width=0.9\textwidth]{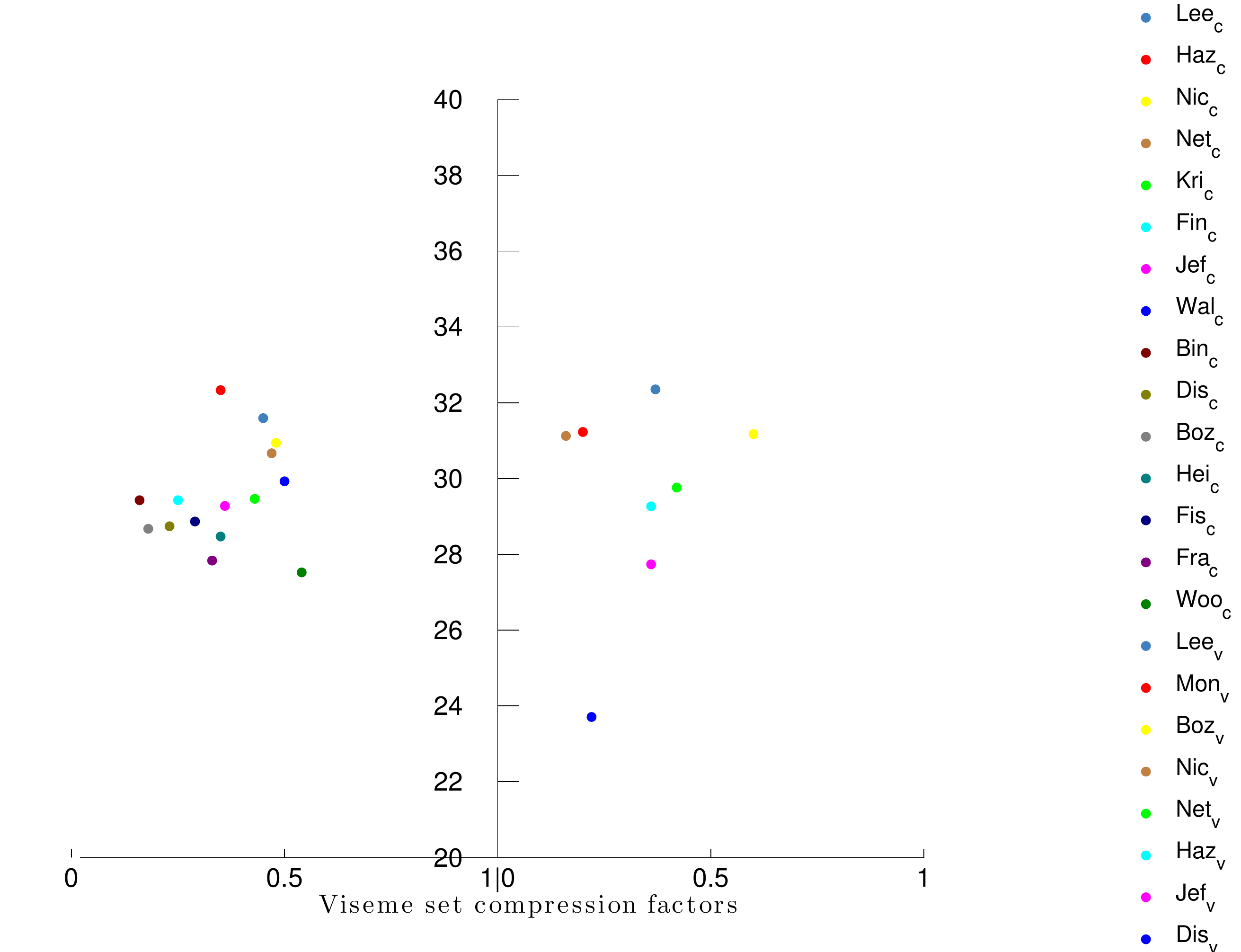} \\ \\
	\includegraphics[width=0.9\textwidth]{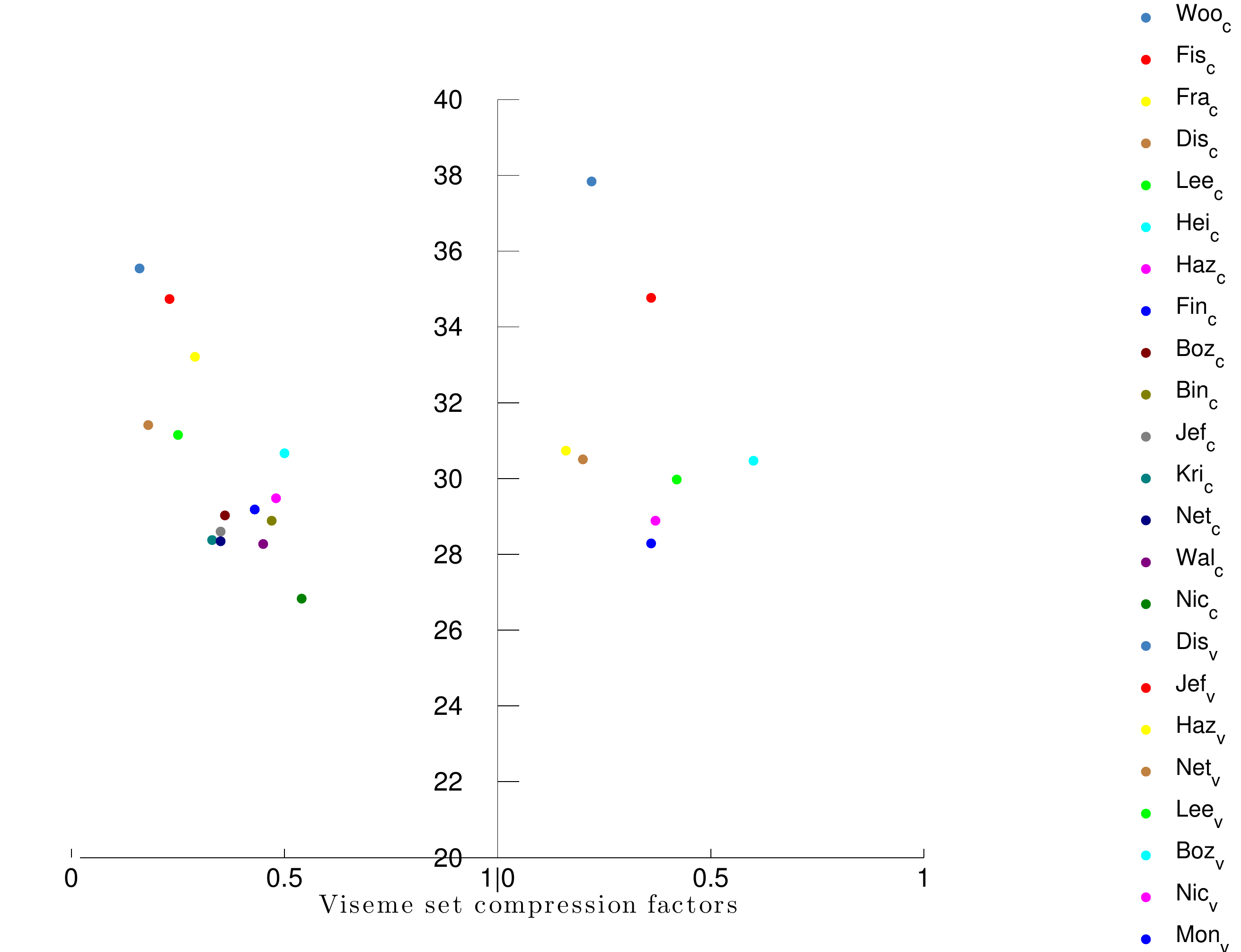} 
\end{tabular}
\caption{Scatter plot showing the relationship between compression factors, $CF_s$ ($x$-axes), and word correctness, $C$, classification ($y$-axes)  with consonant phoneme-to-viseme maps (left) and vowel phoneme-to-viseme maps (right), isolated word results are at the top, and continuous speech along the bottom.} 
\label{fig:scatter_con} 
\end{figure}

The effectiveness of each mapping as a function of compression factor is presented in Figure~\ref{fig:scatter_con}.  The two plots representing continuous speech (bottom of Figure~\ref{fig:scatter_con}) show  improving performance with decreasing compression factor -- we speculated earlier that the coarser visemes were better able to handle co-articulation.  For the isolated word case  (top) there is little difference.  Very roughly, the best performing methods appear to have around 2 to 4 phonemes per viseme.

So far we have seen that there  are noticeable differences between classification performances associated with a variety of  viseme sets in the literature.  Given that quite a few of the viseme sets are incremental improvements on previous sets, it is  good to see confirmation that these sets are have rather similar performance.  We have identified the best sets for the various conditions and have used critical difference plots to explain the similarity between methods. We have identified that the most robust methods  seem to be based on  clustering large amounts of data but  a questions arises when it comes to individual speakers -- is it viable to create viseme sets per speaker and, if so, how similar are they?  This is the topic of the next section.

 
\section{Encoding speaker-dependent visemes} 
In the second part of our phoneme-to-viseme mapping study, two approaches are used to find a better method of mapping phonemes to visemes. These approaches are both speaker-dependent and data-driven from phoneme classification. Two cases are considered: 
\begin{enumerate} 
\item a strictly coupled map, where a phoneme can be grouped into a viseme only if it has been confused with \textit{all} the phonemes within the viseme, and 
\item a relaxed coupled case, where phonemes can be grouped into a viseme if it has been confused with \textit{any} phoneme within the viseme. 
\end{enumerate} 
 
With all new P2V mappings each phoneme can be allocated to only one viseme class. These new P2V maps are tested on the AVL2 dataset using the same classification method as described in Section~\ref{sec:classification}. The results from the best performing P2V map from our comparison study (Lee \cite{lee2002audio} or Woodward \cite{woodward1960phoneme} and Disney \cite{disney}) is the benchmark to measure improvements with respect to the training data. 

\subsection{Viseme classes with strictly confusable phonemes} 
\label{sec:strict_confuse} 
Our approaches for identifying visemes are speaker-dependent, data-driven and based on phoneme confusions within the classifier. The idea of speaker-dependent visemes is not new \cite{visualvowelpercept, bear2015speaker} but our algorithm is, and in conjunction with the fixed outputs available from HTK enables easy reuse. The first undertaking in this work is to complete classification using phoneme labeled HHM classifiers. The classifiers are built in HTK with flat-start HMMs and force-aligned training-data for each speaker. The HMMs are re-estimated 11 times in total over seven folds of leave-one-out cross validation. This overall classification task does not perform well (see Table~\ref{tab:phonemeCorrAVL}) particularly for an isolated word dataset. However, the HTK tool \texttt{HResults} is used to output a confusion matrix for each fold detailing which phoneme labels confuse with others and how often. 
\begin{table}[!h] 
\centering 
\caption{Mean per speaker Correctness, $C$, of phoneme-labeled HMM classifiers.} 
\begin{tabular}{|l|r|r|r|r|} 
\hline 
& Speaker 1 & Speaker 2 & Speaker 3 & Speaker 4 \\ 
\hline \hline 
Phoneme $C$ 	& $24.72$ & $23.63$ & $57.69$ & $43.41$ \\ 
\hline 
\end{tabular} 
\label{tab:phonemeCorrAVL} 
\end{table} 
For both data-driven speaker-dependent approaches, this is the first step of completing phoneme classification is essential to create the data to derive the P2V maps from. This is completed for each speaker in both AVL2 and RMAV datasets. Now, let us use a smaller seven-unit confusion matrix example, as in Table~\ref{fig:demoCMforclustering}, to explain our clustering method.
 
\begin{table}[!h] 
\caption{Demonstration confusion matrix showing confusions between phoneme-labeled classifiers to be used for clustering to create new speaker-dependent visemes. True positive classifications are shown in red, confusions of either false positives and false negatives are shown in blue. The estimated classes are listed horizontally and the real classes are vertical.} 
\centering 
\begin{tabular}{|l||r|r|r|r|r|r|r|} 
\hline 
& $/p1/$ & $/p2/$ & $/p3/$ & $/p4/$ & $/p5/$ & $/p6/$ & $/p7/$ \\ 
\hline \hline 
$/p1/$ & {\textcolor{red}1} & 0 & 0 & 0 & 0 & 0 & {\textcolor{blue}4} \\ 
$/p2/$ & 0 & {\textcolor{red}0} & 0 & {\textcolor{blue}2} & 0 & 0 & 0 \\ 
$/p3/$ & {\textcolor{blue}1} & 0 & {\textcolor{red}0} & 0 & 0 & 0 & {\textcolor{blue}1} \\ 
$/p4/$ & 0 & {\textcolor{blue}2} & {\textcolor{blue}1} & {\textcolor{red}0} & {\textcolor{blue}2} & 0 & 0 \\ 
$/p5/$ & {\textcolor{blue}3} & 0 & {\textcolor{blue}1} & {\textcolor{blue}1} & {\textcolor{red}1} & 0 & 0 \\ 
$/p6/$ & 0 & 0 & 0 & 0 & 0 & {\textcolor{red}4} & 0 \\ 
$/p7/$ & {\textcolor{blue}1} & 0 & {\textcolor{blue}3} & 0 & 0 & 0 & {\textcolor{red}1} \\ 
\hline 
\end{tabular}
\label{fig:demoCMforclustering} 
\end{table} 
 
For the `strictly-confused' viseme set (remembering there is one per speaker), the second step of deriving the P2V map is to check for single-phoneme visemes. Any phonemes which have only been correctly recognized and have no false positive/negative classifications are permitted to be single phoneme visemes. In Table~\ref{fig:demoCMforclustering} we have highlighted the true positive classifications in red and both false positives and false negative classifications in blue which shows $/p6/$ is the only phoneme to fit our `single-phoneme viseme' definition. $/p6/$ has a true positive value of $+4$ and zero false classifications. Therefore this is our first viseme. $/v1/ = \{/p6/\}$. This action is followed by defining all combinations of remaining phonemes which can be grouped into visemes and identifying the grouping that contains the largest number of confusions by ordering all the viseme possibilities by descending size (Table~\ref{fig:listofpossiblecombs}). 
 
\begin{table}[!h] 
\caption{List of all possible subgroups of phonemes with an example set of seven phonemes} 
\label{fig:listofpossiblecombs} 
\centering 
\begin{tabular} {l l l} 
$\{/p1/, /p2/, /p3/, /p4/, /p5/, /p7/\}$ & $\{/p1/, /p2/, /p4/\}$ & $\{/p1/, /p4/, /p7/\}$ \\
$\{/p1/, /p2/, /p3/, /p4/, /p5/ \}$ & $\{/p1/, /p2/, /p5/\}$ & $\{/p2/, /p4/, /p7/\}$ \\ 
$\{/p1/, /p2/, /p3/, /p4/, /p7/\}$ & $\{/p1/, /p2/, /p7/\}$ & $\{/p1/, /p3/\}$ \\
$\{/p1/, /p2/, /p3/, /p5/, /p7/\}$ & $\{/p2/, /p3/, /p4/\}$ & $\{/p1/, /p4/\}$ \\ 
$\{/p1/, /p2/, /p4/, /p5/, /p7/\}$ & $\{/p2/, /p3/, /p5/\}$ & $\{/p1/, /p5/\}$ \\
$\{/p1/, /p3/, /p4/, /p5/, /p7/\}$ & $\{/p2/, /p3/, /p7/\}$ & $\{/p1/, /p7/\}$ \\ 
$\{/p2/, /p3/, /p4/, /p5/, /p7/\}$ & $\{/p3/, /p4/, /p5/\}$ & $\{/p2/, /p3/\}$ \\
$\{/p1/, /p2/, /p3/, /p4/ \}$ & $\{/p3/, /p4/, /p7/\}$ & $\{/p2/, /p4/\}$ \\
$\{/p1/, /p2/, /p3/, /p5/ \}$ & $\{/p1/, /p3/, /p4/\}$ & $\{/p2/, /p5/\}$ \\
$\{/p1/, /p2/, /p3/, /p7/ \}$ & $\{/p4/, /p5/, /p7/\}$ & $\{/p2/, /p7/\}$ \\
$\{/p2/, /p3/, /p4/, /p5/ \}$ & $\{/p1/, /p4/, /p5/\}$ & $\{/p3/, /p4/\}$ \\
$\{/p2/, /p3/, /p4/, /p7/ \}$ & $\{/p2/, /p4/, /p5/\}$ & $\{/p3/, /p5/\}$ \\ 
$\{/p3/, /p4/, /p5/, /p7/ \}$ & $\{/p1/, /p5/, /p7/\}$ & $\{/p4/, /p5/\}$ \\
$\{/p1/, /p3/, /p4/, /p5/ \}$ & $\{/p2/, /p5/, /p7/\}$ & $\{/p4/, /p7/\}$ \\
$\{/p1/, /p4/, /p5/, /p7/ \}$ & $\{/p3/, /p5/, /p7/\}$ & $\{/p5/, /p7/\}$ \\ 
$\{/p2/, /p4/, /p5/, /p7/ \}$ & $\{/p1/, /p3/, /p5/\}$ & \\
$\{/p1/, /p2/, /p3/\}$ & $\{/p1/, /p3/, /p7/\}$ & \\
 
\end{tabular} 
\end{table} 
 
Our grouping rule states that phonemes can be grouped into a viseme class only if all of the phonemes within the candidate group are mutually confusable. This means each pair of phonemes within a viseme must have a total false positive and false negative classification greater than zero. Once a phoneme has been assigned to a viseme class it can no longer be considered for grouping, and so any possible phoneme combinations that include this viseme are discarded. This ensures phonemes can belong to only a single viseme. 
 
By iterating though our list of all possibilities in order, we check if all the phonemes are mutually confused. This means all phonemes have a positive confusion value (a blue value in Table~\ref{fig:demoCMforclustering}) with all others. 
 
The first phoneme possibility in our list where this is true is $\{/p1/, /p3/, /p7/\}$. 
 
This is confirmed by the Table~\ref{fig:demoCMforclustering} values: \\ 
\[
N\{/p1/|/p3/\} + N\{/p3/|/p1/\} = 0 + 1 = 1 > 0 
\]
also,
\[
N\{/p1/|/p7/\} + N\{/p7/|/p1/\} = 4 + 1 = 5 > 0
\]
and,
\[
N\{/p3/|/p7/\} + N\{/p7/|/p3/\} = 1 + 3 = 4 > 0. 
\]
 
This becomes our second viseme and thus our current viseme list looks like Table~\ref{tab:thirdpass}. 
 
\begin{table}[!h] 
\centering 
\caption{Demonstration example 1: first-iteration of clustering, a phoneme-to-viseme map for strictly-confused phonemes.} 
\begin{tabular} {|l|l|} 
\hline 
Viseme & Phonemes \\ 
\hline \hline 
$/v1/$ & $\{/p6/\} $ \\ 
$/v2/$ & $\{/p1/, /p3/, /p7/\}$ \\ 
\hline 
\end{tabular} 
\label{tab:thirdpass} 
\end{table} 
We now only have three remaining phonemes to cluster, $/p2/, /p4/$ and $/p5/$. This reduces our list of possible combinations substantially, see Table~\ref{fig:listofpossiblecombs1}.\\ 
\begin{table} [!h]
\caption{List of all possible subgroups of phonemes with an example set of seven phonemes after the first viseme is formed.} 
\label{fig:listofpossiblecombs1} 
\centering 
\begin{tabular} {l} 
$\{/p2/, /p4/, /p5/\}$ \\ 
$\{/p2/, /p4/\} $ \\ 
$\{/p2/, /p5/\} $ \\ 
$\{/p4/, /p5/\} $ \\ \\ 
\end{tabular} 
\end{table} 
 
The next iteration of our clustering algorithm identifies the combination of remaining phonemes which correspond to the next largest number of confusions, and so on, until no phonemes can be merged. This leaves us with the final visemes in Table~\ref{tab:firstpass}. 
\begin{table}[!h] 
\centering 
\caption{Demonstration example 2: final phoneme-to-viseme map for strictly-confused phonemes.} 
\begin{tabular} {|l|l|} 
\hline 
Viseme & Phonemes \\ 
\hline \hline 
$/v1/$ & $\{/p6/\} $ \\ 
$/v2/$ & $\{/p1/, /p3/, /p7/\}$ \\ 
$/v3/$ & $\{/p2/, /p4/\}$ \\ 
$/v4/$ & $\{/p5/\} $ \\ 
\hline 
\end{tabular} 
\label{tab:firstpass} 
\end{table} 
 
Our original phoneme classification has produced confusion matrices which permit confusions between vowel and consonant phonemes. We can see in Section~\ref{sec:currentp2vmaps} (Tables~\ref{tab:vowelmappings} and~\ref{tab:consonantmappings}), previously presented P2V maps that vowel and consonant phonemes are not commonly mixed within visemes. Therefore, we make two types of P2V maps: one which permits vowels and consonant phonemes to be mixed within the same viseme, and a second which restricts visemes to be vowel or consonant only by putting an extra condition in when checking for confusions greater than zero. 
 
It should be remembered that not all phonemes present in the ground truth transcripts will have been recognized and included in the phoneme confusion matrix. Any of the remaining phonemes which have not been assigned to a viseme are grouped into a single garbage $/gar/$ viseme. This approach ensures any phonemes which have been confused are grouped into a viseme and we do not lose any of the `rarer', and less common visual phonemes. For example, $/ea/$, $/oh/$, $/ao/$, and $/r/$ are not in the original transcript and so can be placed into $/gar/$. But for Speaker 2, $/gar/$ also contains $/ay/$ and $/p/$, and for Speaker 4 $/gar/$ also contains $/p/$ and $/z/$, as these do not show up in the speaker's phoneme classification outputs. This task has been undertaken for all four speakers in our dataset. The final P2V maps are shown in Table~\ref{tab:tcvisemes_split}. 
 
\begin{table}[h] 
\centering 
\caption{Strictly-confused phoneme speaker-dependent visemes. The score in brackets is the compression factor.$B1$ is listed on top, $B2$ visemes are listed at the bottom.} 
\begin{tabular} {|l|l|} 
\hline 
Classification & P2V mapping - permitting mixing of vowels and consonants\\ 
\hline \hline 
Speaker1	& {\footnotesize \{/\textturnv/ /ai/ /i/ /n/ /\textschwa\textupsilon/\} \{/b/ /e/ /ei/ /y/ \} \{/d/ /s/\} \{/t\textipa{S}/ /l/\} \{/\textschwa/ /v/\}} \\ 
(CF:0.48)	& {\footnotesize \{/w/\} \{/f/\} \{/k/\} \{/\textschwa/ /v/\} \{/d\textipa{Z}/ /z/\} \{/\textscripta/ /u/\} 	 \{/t/\} }\\ 
Speaker2 	& {\footnotesize\{/\textschwa/ /ai/ /ei/ /i/ /s/\} \{/e/ /v/ /w/ /y/\} \{/l/ /m/ /n/\} \{/b/ /d/ /p/\} }\\ 
(CF: 0.44)	& {\footnotesize \{/z/\} \{t\textipa{S}/\} \{/t/\} \{/\textscripta/\} \{/d\textipa{Z}/ /k/\} \{/\textturnv/ /f/\} \{/\textschwa\textupsilon/ /u/\}}	\\ 
Speaker3 	& {\footnotesize \{/ei/ /f/ /n/\} \{/d/ /t/ /p/\} \{/b/ /s/\} \{/l/ /m/\} \{/\textschwa/ /e/\} \{/i/\} \{/u/\} }\\ 
(CF: 0.68)	& {\footnotesize \{/\textscripta/\} \{/d\textipa{Z}/\} \{/\textschwa\textupsilon/\} \{/z/\} \{/y/\} \{/t\textipa{S}\}/ \{/ai/\} \{/\textturnv/\} \{/\textscripta/\} \{/d\textipa{Z}/\} \{/\textschwa\textupsilon/\}	} \\ 
& {\footnotesize \{/k/ /w/\} \{/v/\} \{/z/\} }\\ 
Speaker4	& {\footnotesize \{/\textturnv/ /ai/ /i/ /ei/ \} \{/m/ /n/\} \{/\textschwa/ /e/ /p/\} \{/k/ /w/\} \{/d/ /s/\} \{/d\textipa{Z}/ /t/\} } \\ 
(CF: 0.64)	& {\footnotesize \{/f/\} \{/v/\} \{/\textscripta/\} \{/z/\} \{/t\textipa{S}/\} \{/b/\} \{/\textschwa\textupsilon/\}	 \{/\textschwa\textupsilon/\} \{/l/\} \{/u/\} \{/b/\} } \\ 
\hline 
\hline 
Classification & P2V mapping - restricting mixing of vowels and consonants \\ 
\hline \hline 
Speaker1	& {\footnotesize \{/\textturnv/ /i/ /\textschwa\textupsilon/ /u/\} \{/\textscripta/ /ei/\} \{/\textschwa/ /e/ /ei/\} \{/d/ /s/ /t/ \} \{/t\textipa{S}/ /l/ \} \{/k/\} }\\ 
(CF:0.50)	& {\footnotesize \{/z/\} \{/w/\} \{/f/\} \{/m/ /n/\} \{/d\textipa{Z}/ /v/\} \{/b/ /y/\} 	}\\ 
Speaker2 	& {\footnotesize \{/ai/ /ei/ /i/ /u/\} \{/\textschwa\textupsilon/\} \{/\textschwa/\} \{/e/\} \{/\textturnv/\} \{/\textscripta/\} \{/v/ /w/\} \{/d\textipa{Z}/ /p/ /y/\}}\\ 
(CF: 0.58)	& {\footnotesize \{/d/ /b/\} \{/t/\} \{/k/\} \{/t\textipa{S}/\} \{/l/ /m/ /n/\} \{/f/ /s/\} }\\ 
Speaker3 	& {\footnotesize \{/ei/ /i/\} \{/ai/\} \{/\textschwa/ /e/\} \{/\textturnv/\} \{/d/ /p/ /t/\} \{/l/ /m/\} \{/k/ /w/\} \{/v/\} }\\ 
(CF: 0.68)	& {\footnotesize \{/t\textipa{S}/\} \{/\textschwa\textupsilon/\} \{/y/\} \{/u/\} \{/\textscripta/\} \{/z/\} \{/f/ /n/\} \{/b/ /s/\} \{/d\textipa{Z}/\} }\\ 
Speaker4 	& {\footnotesize \{/\textturnv/ /ai/ /i/ /ei/\} \{/\textschwa/ /e/\} \{/m/ /n/\} \{/k/ /l/\} \{/d\textipa{Z}/ /t/\} \{/d/ /s/\} \{/t\textipa{S}/\} }\\ 
(CF: 0.65)	& {\footnotesize \{/\textschwa\textupsilon/\} \{/y/\} \{/u/\} \{/\textscripta/\} \{/w/\} \{/f/\} \{/v/\} \{/b/\} }\\ 
\hline 
\end{tabular} 
\label{tab:tcvisemes_split} 
\end{table} 
 
\subsection{Viseme classes with relaxed confusions between phonemes} 
\label{sec:loose_confuse} 
A disadvantage of the strictly confusable viseme set is that it contains some spurious single-phoneme visemes where the phoneme cannot be grouped because it is not confused with \emph{all} other phonemes in the viseme. These types of phonemes are likely to be either: borderline cases at the extremes of a viseme cluster, i.e. they have subtle visual similarities to more than one phoneme cluster, or they do not occur frequently enough in the training data to be differentiated from other phonemes. 

\begin{table}[!t] 
\centering 
\caption{Demonstration example 3: final phoneme-to-viseme map for relaxed-confused phonemes.} 
\begin{tabular} {|l|l|} 
\hline 
Viseme & Phonemes \\ 
\hline \hline 
$/v1/$ & $\{/p6/\} $ \\ 
$/v2/$ & $\{/p1/, /p3/, /p5/, /p7/\}$ \\ 
$/v3/$ & $\{/p2/, /p4/\}$ \\ 
\hline 
\end{tabular} 
\label{tab:secondpass} 
\end{table} 
To address this we complete a second pass-through of the strictly-confused visemes listed in Table~\ref{tab:firstpass}. We begin with the visemes as they currently stand (in our demonstration example containing four classes) and relax the condition requiring confusion with all of the phonemes. Now any single phoneme viseme (in our demonstration, $/v4/$) can be allocated to a previously existing viseme if it has been confused with any phoneme in the viseme. In Table~\ref{fig:demoCMforclustering} we see $/p5/$ was confused with $/p1/$, $/p3/$, and $/p4/$. Because $/p4/$ is not in the same viseme as $/p1/$ and $/p3/$ we use the value of confusion to decide which to allocate it to as follows. \\ 
\[
N\{ /p1/|/p5/\} + N\{/p5/|/p1/\} = 0 + 3 = 3
\]
\[
N\{/p3/|/p5/\} + N\{/p5/|/p3/\} = 0 + 1 = 1
\]
\[
N\{/p4/|/p5/\} + N\{/p5/|/p4/\} = 2 + 1 = 3
\]
Therefore; for $p5$ the total confusion with $/v2/$ is $3+1=4$, whereas the total confusion with $/v3/$ is $3$. We select the viseme with most confusion to incorporate the unallocated phoneme $/p5/$. This reduces the number of viseme classes by merging single-phoneme visemes from Table~\ref{tab:firstpass} to form a second set shown in Table~\ref{tab:secondpass}. This has the added benefit that we have also increased the number of training samples for each classifier.

 \begin{table}[h] 
\centering 
\caption{The four variations on speaker-dependent phoneme-to-viseme maps derived from phoneme confusion in phoneme classification.} 
\begin{tabular}{|l|l|} 
\hline 
Bear1, $B1$: & Bear2, $B2$: \\ 
Mixed vowels and consonants & Split vowels and consonants \\ 
\multicolumn{1}{|c|}{$+$} & \multicolumn{1}{c|}{$+$} \\ 
Strict-confusion of phonemes & Strict-confusion of phonemes \\ 
\hline 
Bear3, $B3$: & Bear4, $B4$ \\ 
Mixed vowels and consonants & Split vowels and consonants \\ 
\multicolumn{1}{|c|}{$+$} & \multicolumn{1}{c|}{$+$} \\ 
Relaxed-confusion of phonemes & Relaxed-confusion of phonemes \\ 
\hline 
\end{tabular} 
\label{fig:sdtypes} 
\end{table} 
Remember, as we have two versions of Table~\ref{tab:firstpass} - one with mixed vowel and consonant phonemes and a second with divided vowels and consonant phonemes - the same still applies to our relaxed-confused visemes sets. This means we end up with four types of speaker-dependent phoneme-to-viseme maps, described in Table~\ref{fig:sdtypes}. For our strictly-confused P2V maps in Table~\ref{tab:tcvisemes_split}, these become the relaxed P2V maps in Table~\ref{tab:lcvisemes_split}. In Table~\ref{fig:sdtypes} we have labeled each of the four variations $B1$, $B2$, $B3$ and $B4$ for ease of reference. 
 
\begin{table}[!h] 
\centering 
\caption{Relaxed-confused phoneme speaker-dependent visemes. The score in brackets is the ratio of visemes to phonemes. $B3$ visemes are on top, and $B4$ listed below.} 
\begin{tabular} {|l|l|} 
\hline 
Classification & P2V mapping - permitting mixing of vowels and consonants \\ 
\hline \hline 
Speaker1 	& {\footnotesize \{/b/ /e/ /ei/ /p/ /w/ /y/ /k/\} \{/\textturnv/ /ai/ /f/ /i/ /m/ /n/ /\textschwa\textupsilon/\}}\\ 
(CF:0.28)	& {\footnotesize \{/d\textipa{Z}/ /z/\} \{/\textscripta/ /u/\} \{/d/ /s/ /t/\} \{/t\textipa{S}/ /l/\} \{/\textschwa/ /v/\}\{/\textschwa/ /v/\} 	}\\ 
Speaker2 	& {\footnotesize \{/\textscripta/ /\textschwa/ /ai/ /ei/ /i/ /s/ /t\textipa{S}/\} \{/e/ /t/ /v/ /w/ /y/\} \{/l/ /m/ /n/\} }\\ 
(CF: 0.32)	& {\footnotesize \{/\textturnv/ /f/\} \{/z/\} \{/b/ /d/ /p/\} \{/\textschwa\textupsilon/ /u/\} \{/d\textipa{Z}/ /k/\} }\\ 
Speaker3	& {\footnotesize \{/\textturnv/ /ai/ /ei/ /f/ /i/ /n/\} \{/\textschwa/ /e/ /y/ /t\textipa{S}/\} \{/b/ /s/ /v/\} \{/l/ /m/ /u/\} 	}\\ 
(CF: 0.40)	& {\footnotesize \{/d\textipa{Z}/\} \{/\textschwa\textupsilon/\} \{/z/\} \{/d/ /p/ /t/\} \{/k/ /w/\} \{/\textscripta/\} }\\ 
Speaker4 	& {\footnotesize \{/\textturnv/ /ai/ /t\textipa{S}/ /i/ /ei/ \} \{/\textscripta/ /m/ /u/ /n/\} \{/\textschwa/ /e/ /p/ /v/ /y/\} }\\ 
(CF: 0.32)	& {\footnotesize \{/d\textipa{Z}/ /t/\} \{/k/ /l/ /w/\} \{/\textschwa\textupsilon/\} \{/d/ /f/ /s/\} \{/b/\} }\\ 
\hline 
%
\hline 
Classification & P2V mapping - restricting mixing of vowels and consonants\\ 
\hline \hline 
Speaker1 	& {\footnotesize \{/\textturnv/ /i/ /\textschwa\textupsilon/ /u/\} \{/\textscripta/ /ai/\} \{/\textschwa/ /e/ /ei/\} \{/b/ /w/ /y/\} \{/d/ /f/ /s/ /t/\} }\\ 
(CF:0.47)	& {\footnotesize \{/k/\} \{/z/\} \{/m/\} \{/l/\} \{/t\textipa{S}/\} \{/d\textipa{Z}/ /k/ /v/ /z/\} }\\ 
Speaker2 	& {\footnotesize \{/\textscripta/ /\textturnv/ /\textschwa/ /ai/ /ei/ /i/ /\textschwa\textupsilon/ /u/\} \{/k/ /t/ /v/ /w/\} \{/t\textipa{S}/ /l/ /m/ /n/\} }\\ 
(CF: 0.29)	& {\footnotesize \{/f/ /s/\} \{/d\textipa{Z}/ /p/ /y/\} \{/b/ /d/\} \{/z/\}	}\\ 
Speaker3 	& {\footnotesize \{/\textturnv/ /ai/ /i/ /ei/\} \{/\textschwa/ /e/\} \{/b/ /s/ /v/\} \{/d/ /p/ /t/\} \{/l/ /m/\} }\\ 
(CF: 0.56)	& {\footnotesize \{/y/\} \{/d\textipa{Z}/\} \{/\textschwa\textupsilon/\} \{/z/\} \{/u/\} \{/\textschwa/ /e/\} \{/k/ /w/\} \{/f/ /n/\} \{/\textscripta/\} \{/t\textipa{S}/\} }	\\ 
Speaker4 	& {\footnotesize \{/\textturnv/ /ai/ /i/ /ei/\} \{/t\textipa{S}/ /k/ /l/ /w/\} \{/d/ /f/ /s/ /v/\} \{/m/ /n/\} }\\ 
(CF: 0.50)	& {\footnotesize \{/f/\} \{/\textscripta/\} 	\{/d\textipa{Z}/ /t/\} \{/\textschwa\textupsilon/\} \{/u/\} \{/y/\} \{/b/\} }\\ 
\hline 
\end{tabular} 
\label{tab:lcvisemes_split} 
\end{table} 
 
Now, and this is why these visemes are defined as relaxed, any remaining phonemes which have confusions, but are so far not assigned to a viseme, the phoneme-pair confusions are used to map the remaining phonemes to an appropriate viseme, even though it does not confuse with all phonemes already in it. Any remaining phonemes which are not assigned to a viseme are grouped into a new garbage $/gar/$ viseme. This approach ensures any phonemes which have been confused with any other are grouped into a viseme. 

\clearpage
\subsection{Results analysis} 
\begin{figure}[!h]
\centering
\begin{tabular}{c}
	\includegraphics[width=1\textwidth]{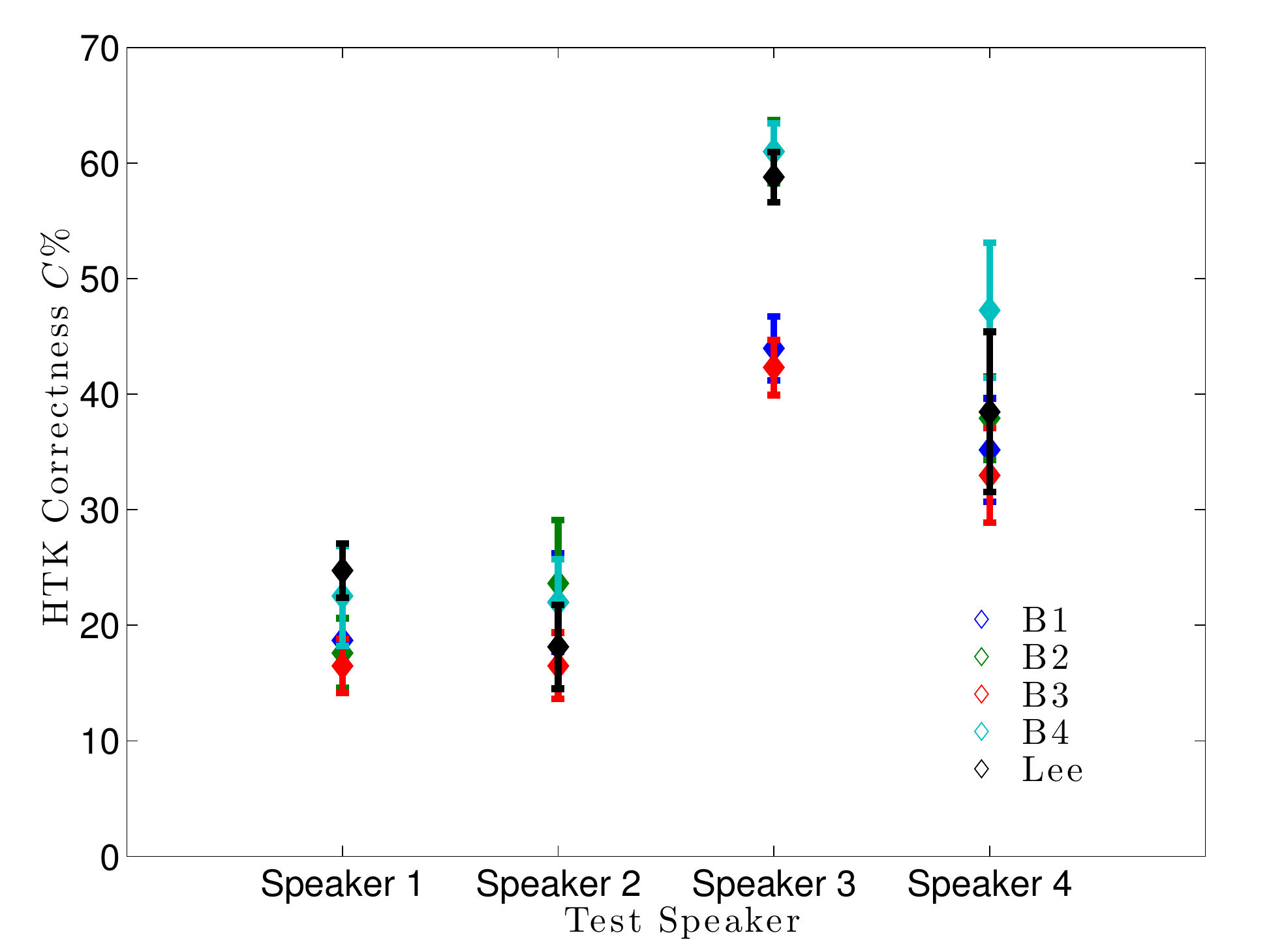} \\
	\includegraphics[width=1\textwidth]{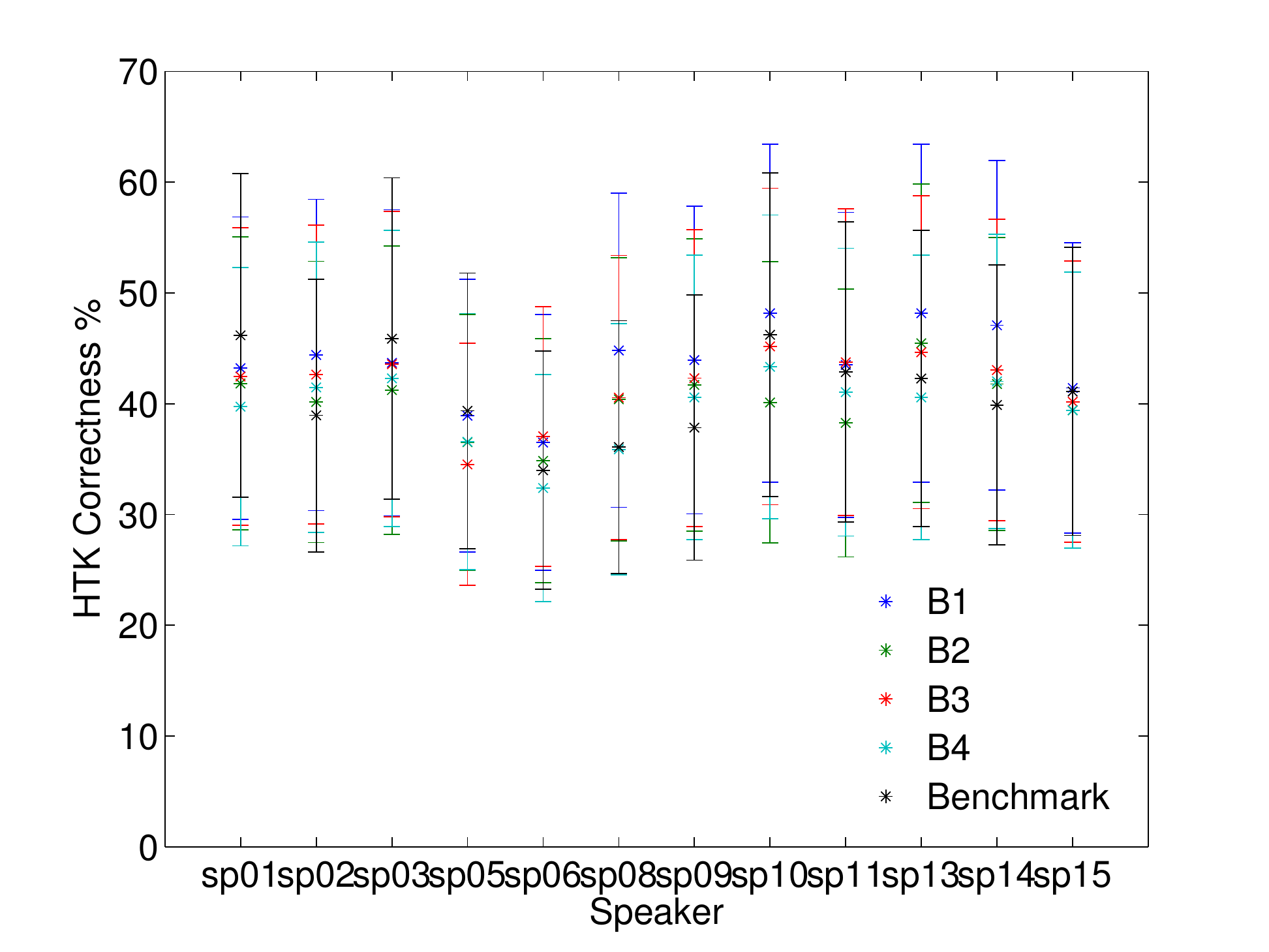} 
\end{tabular} 
\caption{Word classification correctness $C \pm1 \mbox{se}$, using all four new methods of deriving speaker dependent visemes. AVL2 (top) and RMAV (bottom) speakers against Lee (top) and Woodward and Disney (bottom) benchmarks in black.}
\label{fig:bear_visemes} 
\end{figure}

Figure~\ref{fig:bear_visemes} (top) compares the new speaker-dependent viseme method with the Lee visemes which are the benchmark from the isolated word study.
For Speaker 1 and Speaker 3, no new viseme map significantly improves upon Lee's performance although we do see improvements for both Speaker 2 and Speaker 4. The strictly-confused and split viseme map improves upon Lee's previous best word classification. 
 
The second set of our experiments with continuous speech training data (RMAV) is to repeat our investigation with speaker-dependent visemes. These have been derived with the same methods described in Section~\ref{sec:strict_confuse} \&~\ref{sec:loose_confuse} and are listed in full for each speaker in~\ref{app:rmavP2Vmaps}. Our classification method is identical to that used previously with HMMs. In the previous work of \cite{bear2014phoneme}, we see limited improvement in word classification with viseme classes due to the size of the dataset. 

In Figure~\ref{fig:bear_visemes} (bottom) we have plotted the word correctness achieved for each RMAV speaker using all four variants of the speaker-dependent visemes. Our first observation is that on this figure, the correctness scores achieved range from $26.67\%$ to $41.53\%$, whereas in Figure~\ref{fig:bear_visemes} (top) the values range from $20.60\%$ to $36.53\%$. As before, this overall increase is attributed to the larger volume of training samples in RMAV compared to AVLetters2. 
 
Compared to the benchmark of the Disney vowels and Montgomery consonant visemes which has been plotted in black on Figure~\ref{fig:bear_visemes} (bottom) we see that the comparison between speaker-dependent visemes and the best speaker-independent visemes is subject to the speaker. For three out of 12 speakers (sp01, sp03, sp05), the speaker-dependent visemes are all worse than our benchmark. For another three of our 12 speakers (sp02, sp09, sp14) all of the speaker-dependent visemes out-perform the benchmark. For all six remaining speakers, the results are mixed. This suggests that it is possible that speaker-dependent visemes could improve on speaker-independent ones, but that it is essential that they are exactly right for the individual otherwise they become at worse, detrimental, or a lot of effort for no significant improvement. 

Careful observation of Figure~\ref{fig:bear_visemes} (top) shows that when considering the performance of mixed or split visemes, split visemes signfificantly ($>1$se) outperform mixed.  When considering relaxed versus split  the split has a marginal advantage but it is not significant ($<$1se).

The comparison of strict and split visemes for continuous speech (Figure~\ref{fig:bear_visemes} (bottom) is consistent with the isolated word observations. The strictly-confused visemes perform better than those with a relaxed confusion, but not statistically significantly ($<$1se). Again, we see that mixing vowel and consonants phonemes within individual viseme classes reduces the classification performance but not significantly.

\begin{figure}[h] 
\centering 
\includegraphics[width=0.85\textwidth]{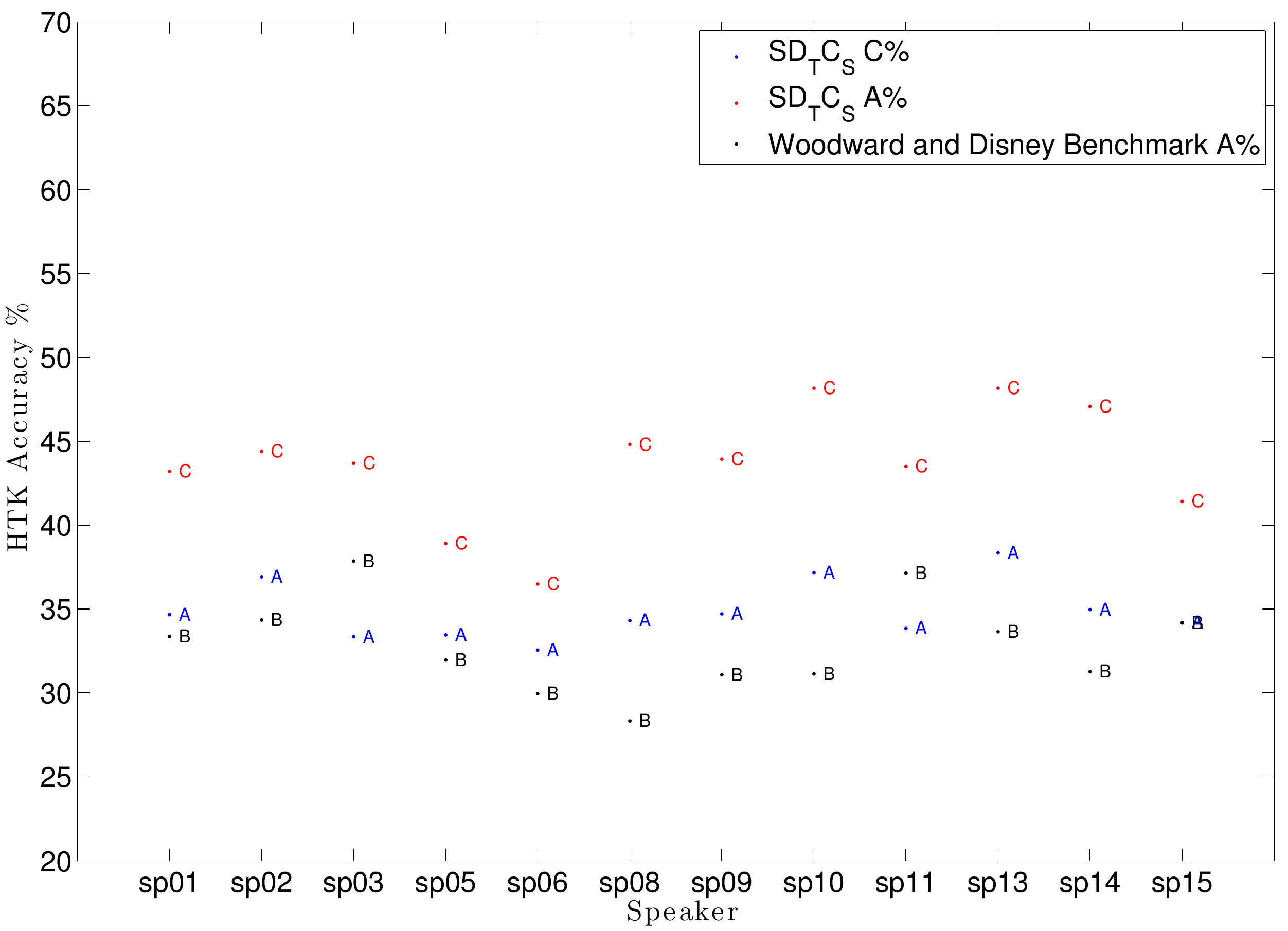} 
\caption{Comparing the accuracy change between strict and relaxed visemes to show the improvement in accuracy/reduction in insertion errors for all 12 speakers in continuous speech. The baseline is the correctness classification which ignores insertion error penalties.} 
\label{fig:accuracy_improvement}
\end{figure} 
 
In Figure~\ref{fig:accuracy_improvement} we have plotted accuracy, $A$, and correctness, $C$, for our best performing speaker-dependent visemes ($B1$) on continuous speech. We also plot, the accuracy scores of our benchmark from Woodward and Disney's visemes. These are compared with the correctness scores as a baseline to show the improvement. Whilst the improvement of speaker-dependent visemes is not significant when measured by Correctness, by plotting the accuracy of the viseme classifiers we can see that they do have a positive influence in reducing insertion errors which are a bugbear of lipreading.

\section{Performance of individual visemes} 

\begin{figure}[!h] 
\centering 
\begin{tabular}{c}
	\includegraphics[width=\textwidth]{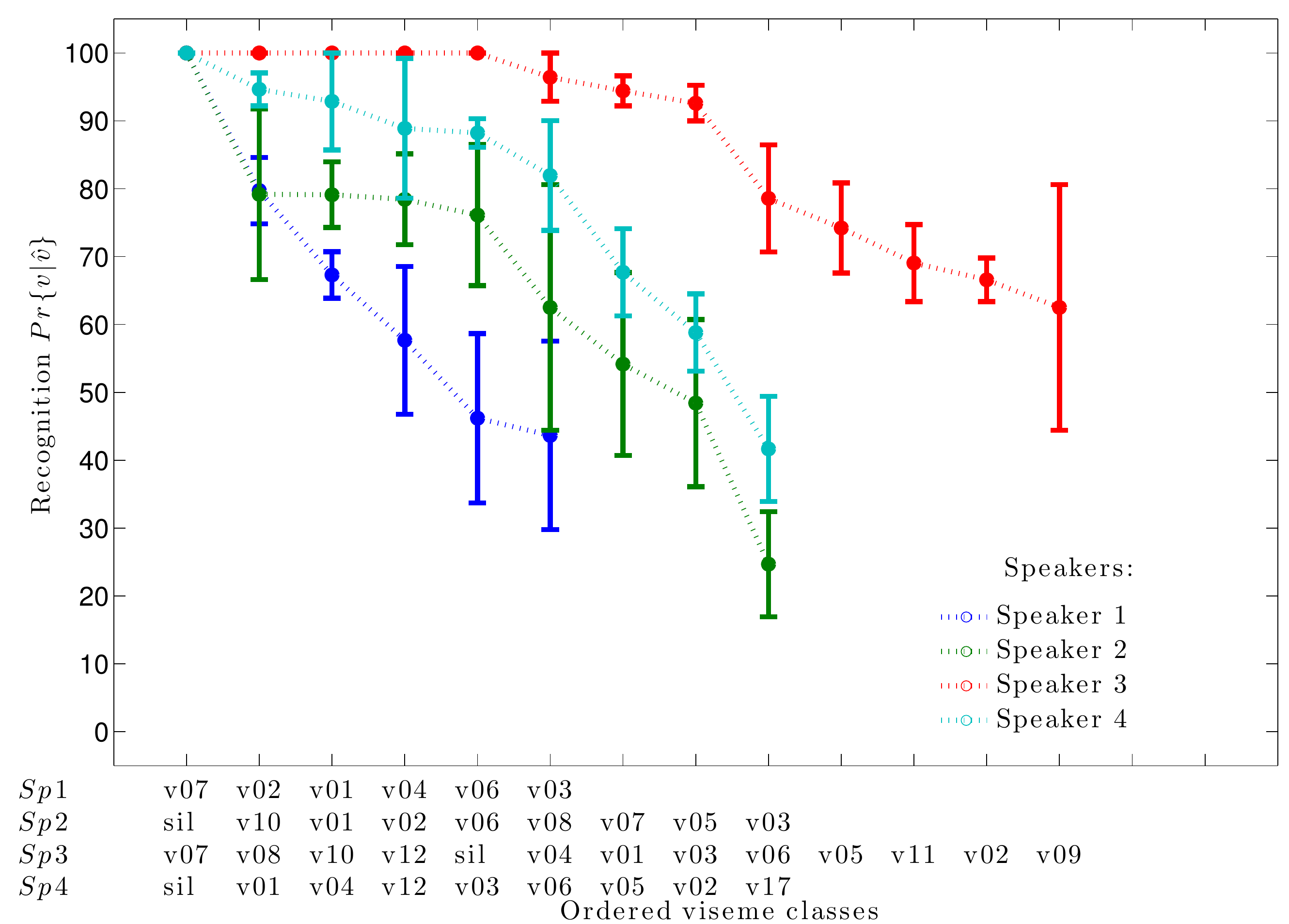} \\
	\includegraphics[width=\textwidth]{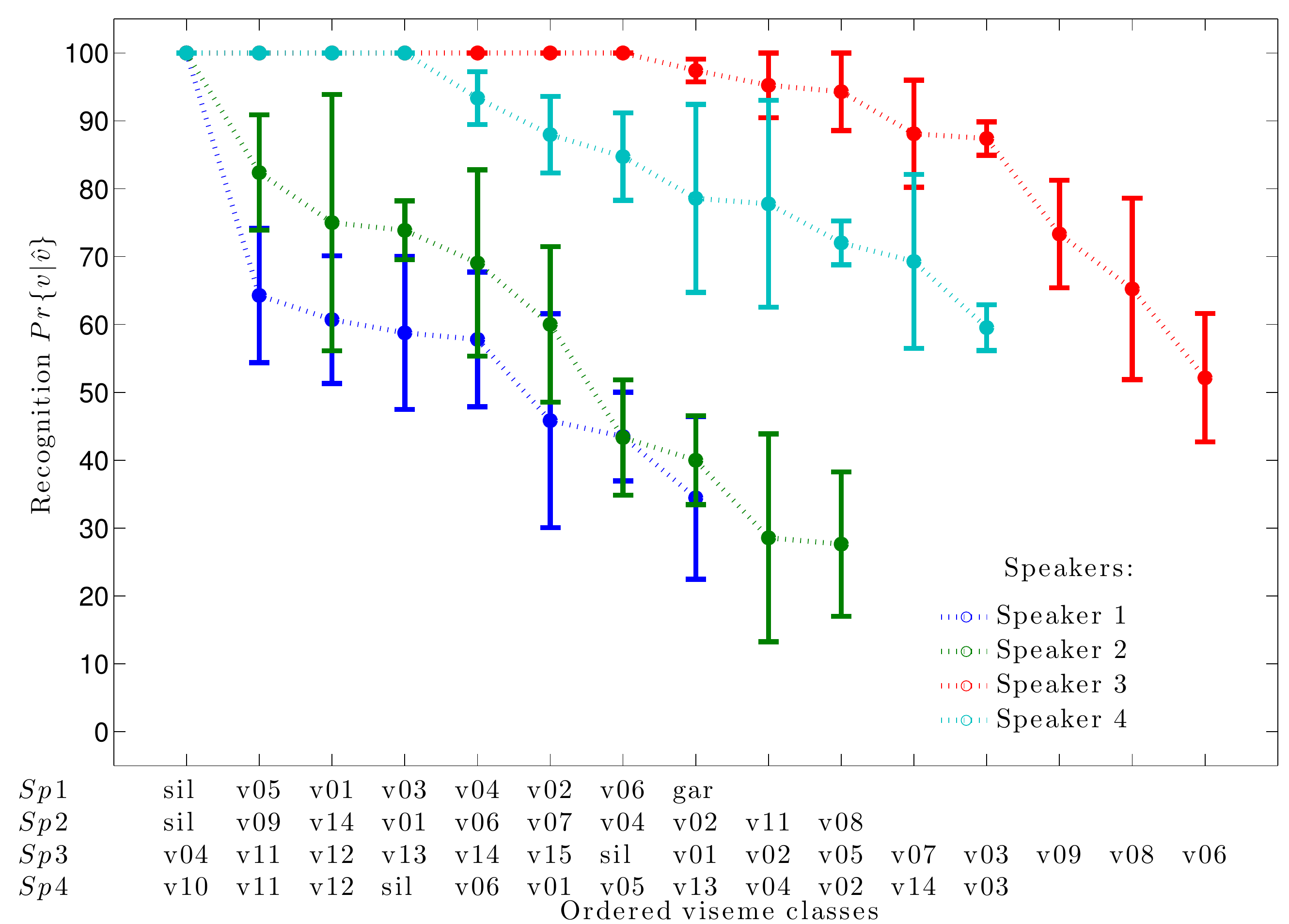} 
\end{tabular}
\caption{Individual viseme classification, Pr$\{v|\hat{v}\}$ with speaker-dependent visemes for four speakers with isolated word training of classifiers B1 visemes (top) and B2 visemes (bottom).}
\label{fig:contrib_rm} 
\end{figure}

\begin{figure}
\centering 
\begin{tabular}{c}
	\includegraphics[width=\textwidth]{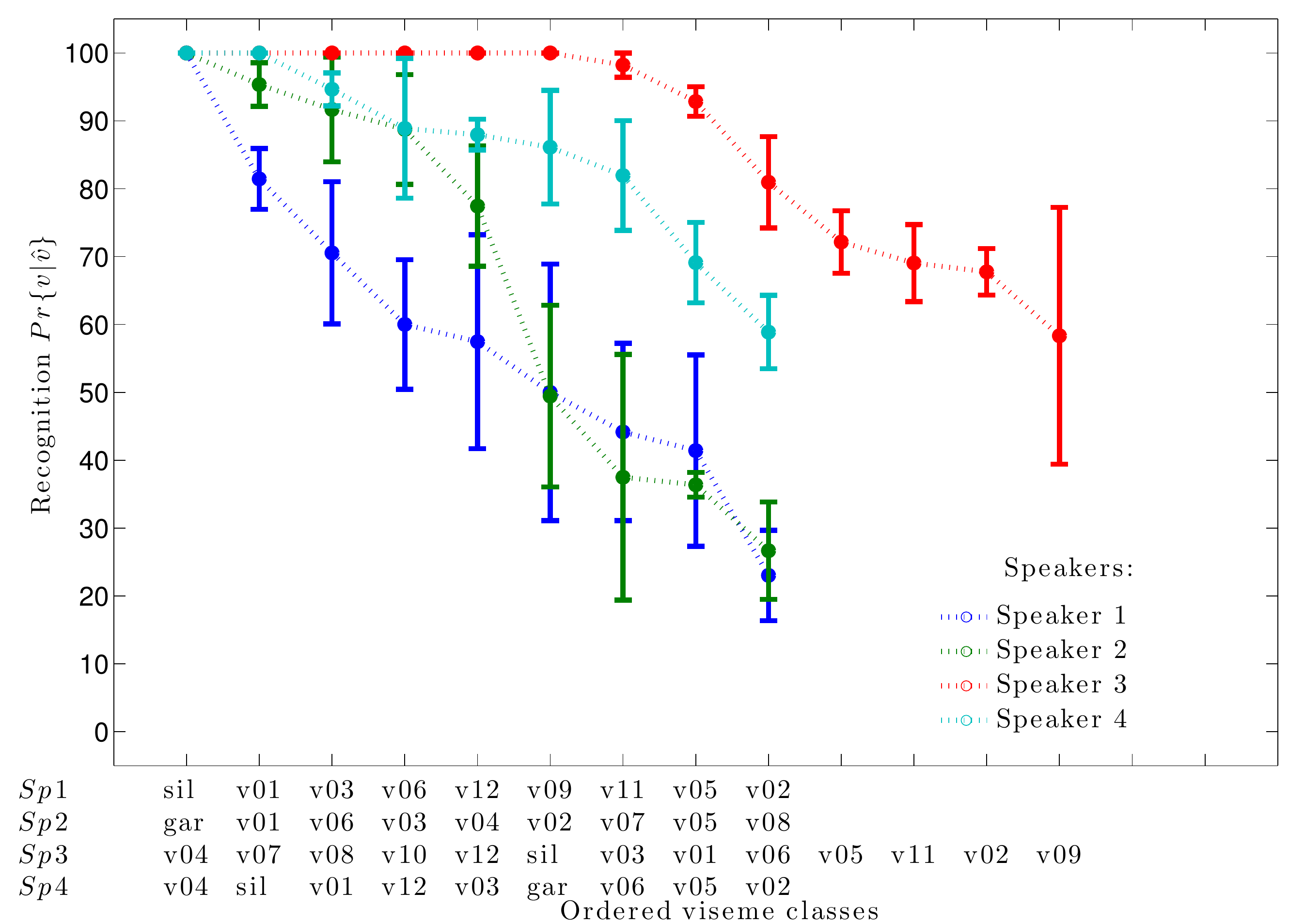} \\
		\includegraphics[width=\textwidth]{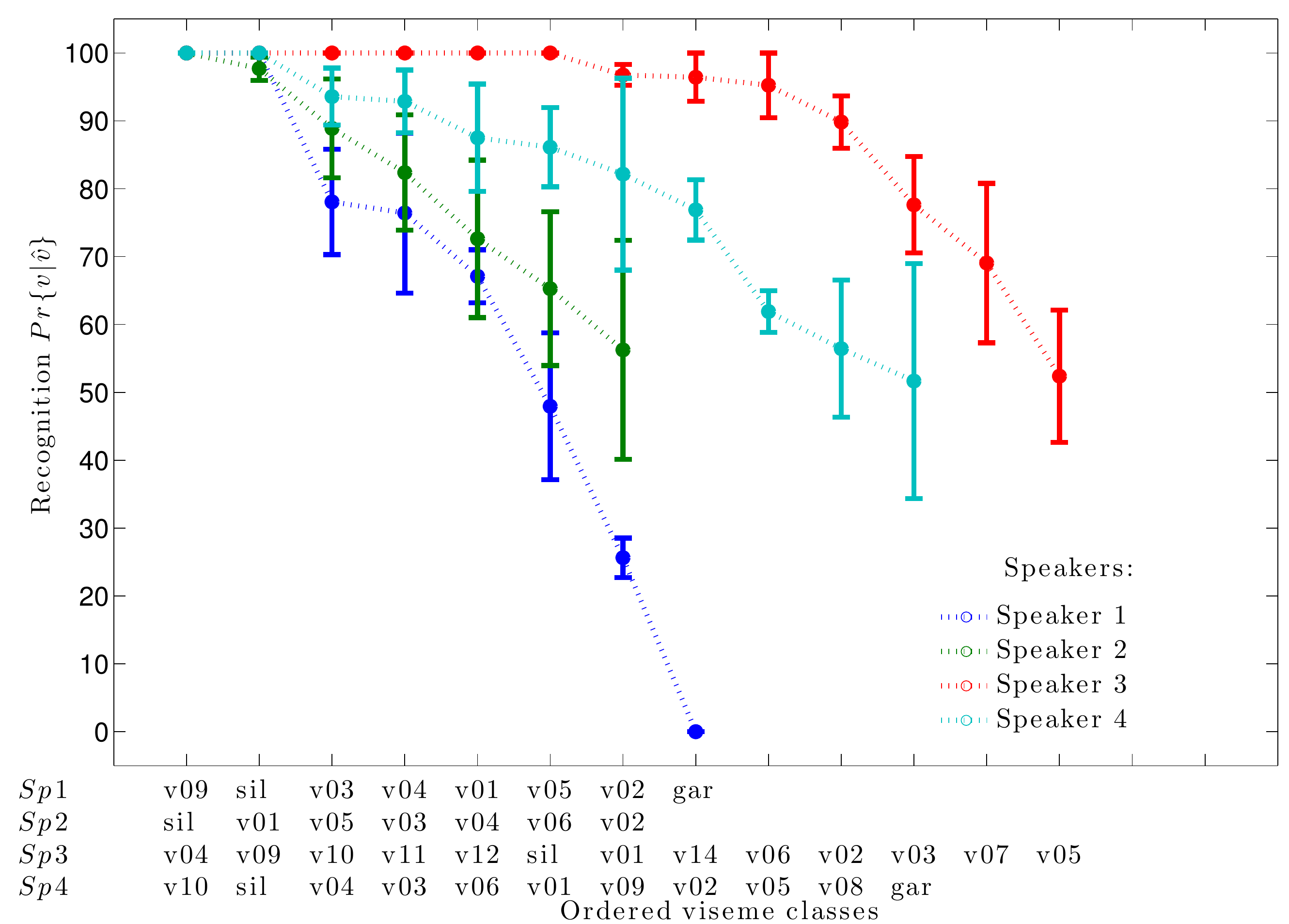} 
\end{tabular}
\caption{Individual viseme classification, Pr$\{v|\hat{v}\}$ with speaker-dependent visemes for four speakers with isolated word training of classifiers. B3 visemes (top) and B4 visemes (bottom).}
\label{fig:contrib_sm}
\end{figure} 

In Figures~\ref{fig:contrib_rm} and \ref{fig:contrib_sm}, the contribution of each viseme has been listed in descending order along the $x-$axis for each speaker in AVL2. The contribution of each viseme is measured as the probability of each class, Pr$\{v|\hat{v}\}$. These values have been calculated from the \texttt{HResults} confusion matrices. 


This analysis of visemes within a set is also used in \cite{bear2014some}, which proposes a threshold subject to the information in the features.

\begin{figure}[!h] 
\centering 
\begin{tabular}{c}
	\includegraphics[width=\textwidth]{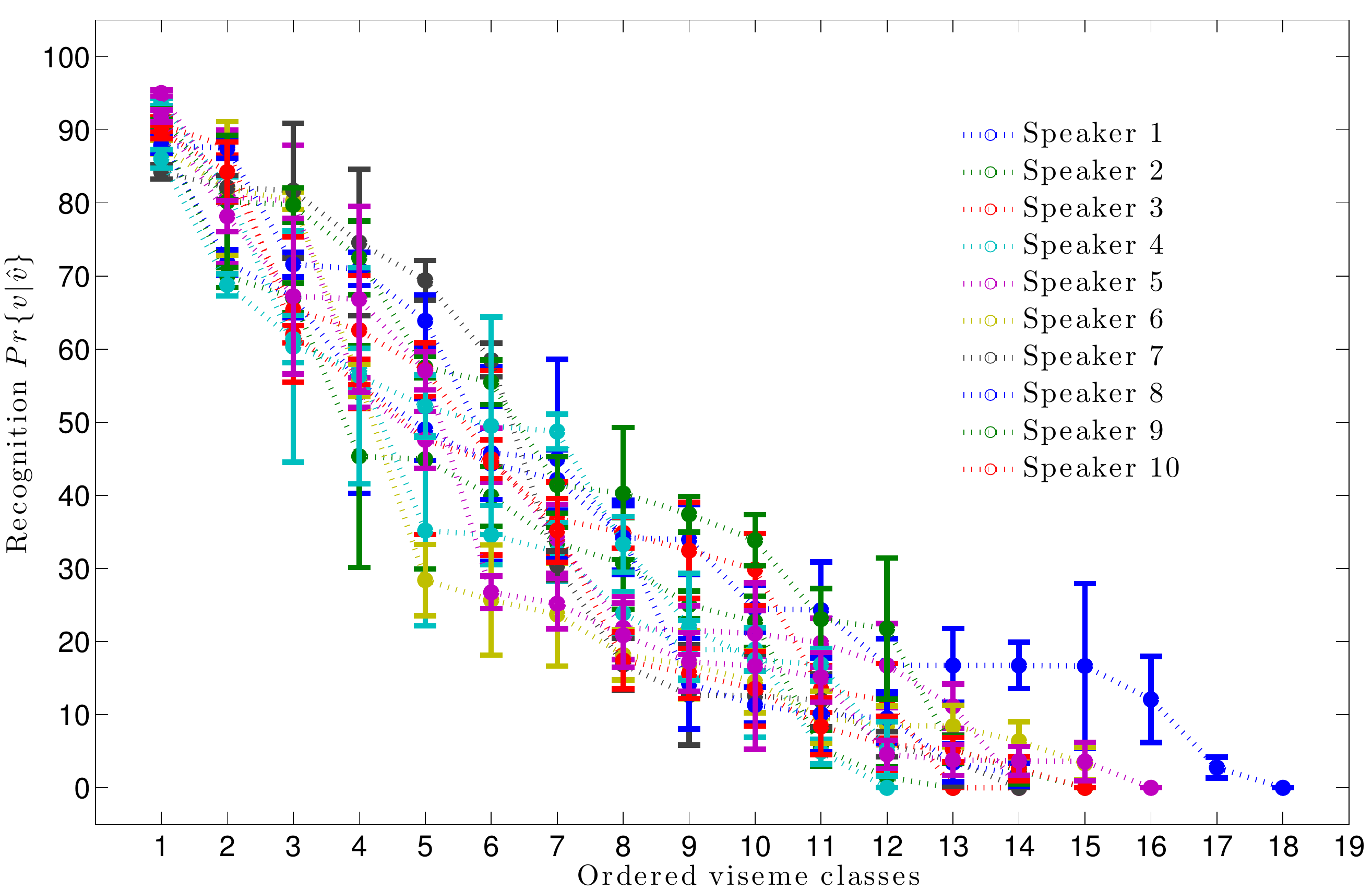} \\
	\includegraphics[width=\textwidth]{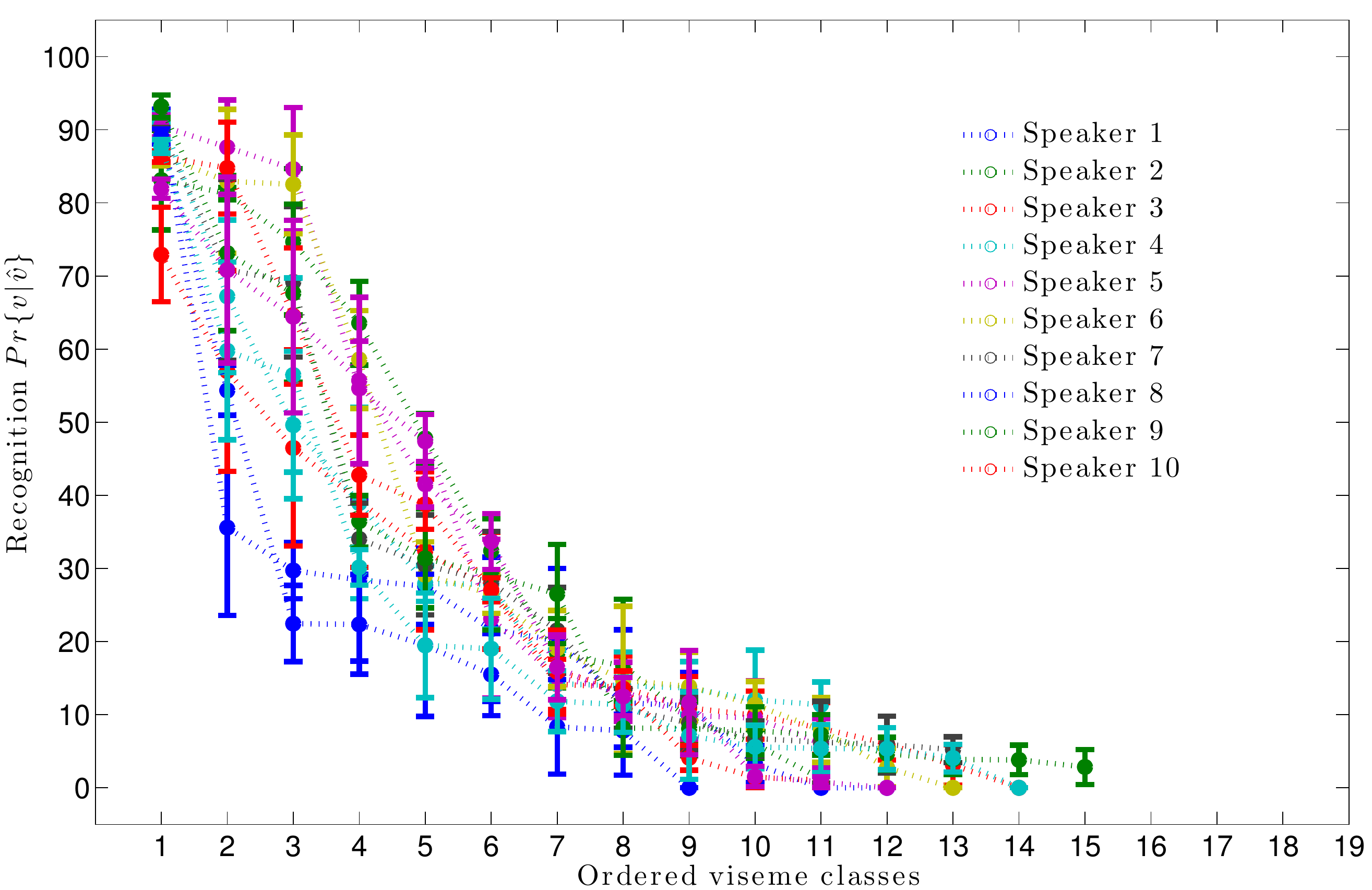} 
\end{tabular}
\caption{Individual viseme classification, Pr$\{v|\hat{v}\}$ with speaker-dependent visemes for twelve speakers with continuous speech training of classifiers. B1 visemes (top) and B2 visemes (bottom).}
\label{fig:contrib_rm_lilir} 
\end{figure}

\begin{figure}[!h]
\centering
\begin{tabular}{c}
	\includegraphics[width=\textwidth]{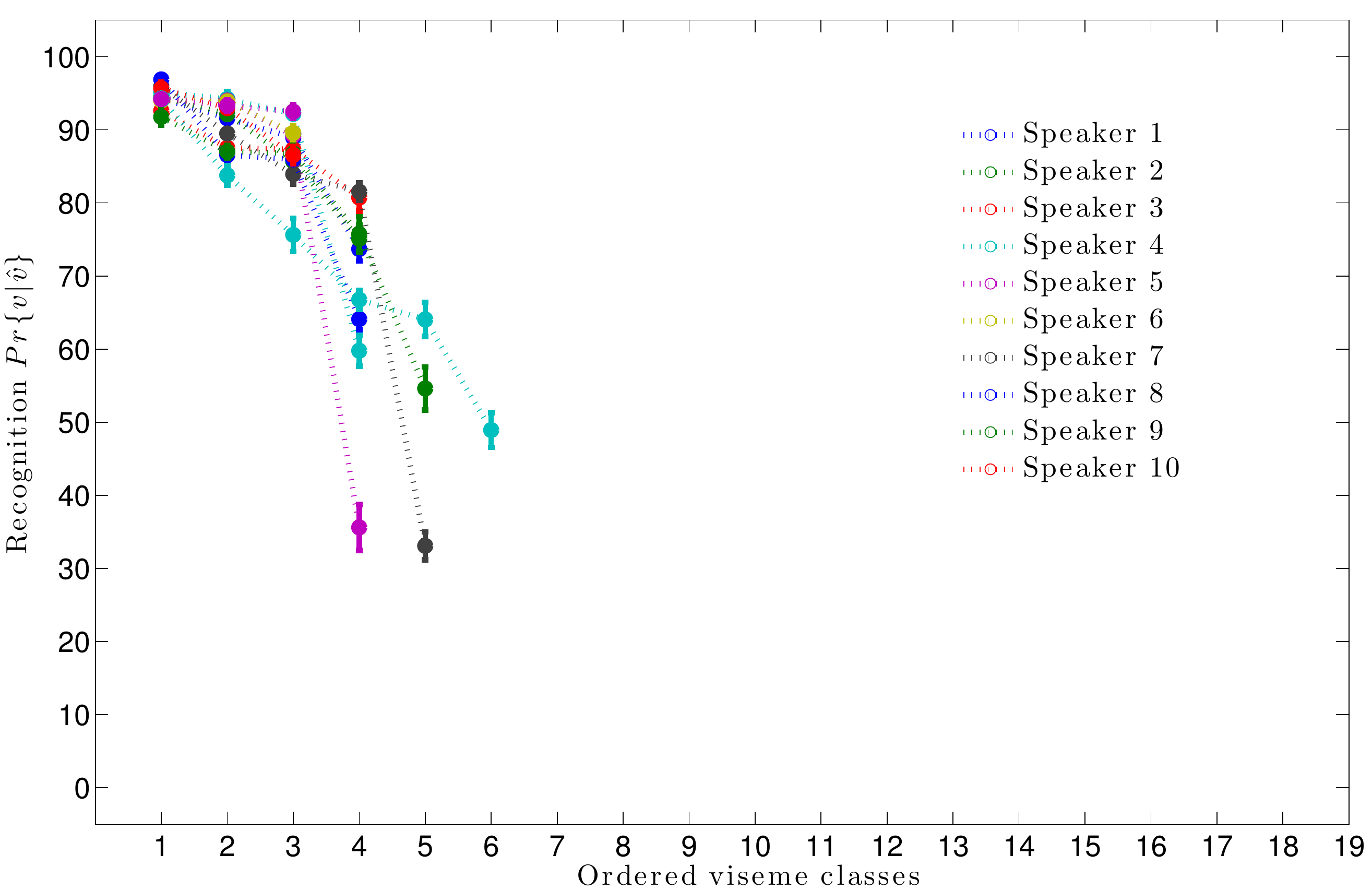} \\
	\includegraphics[width=\textwidth]{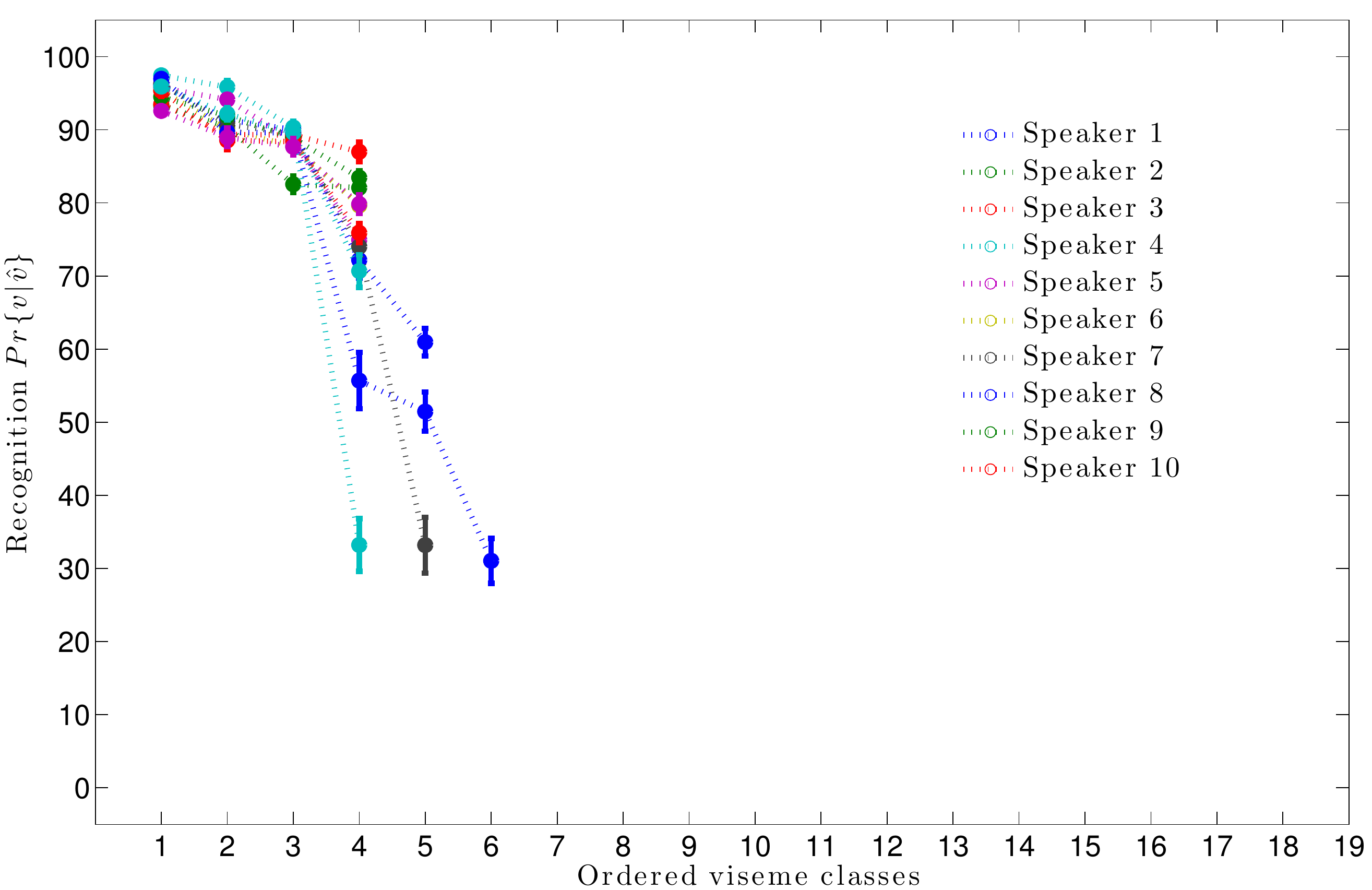} 
\end{tabular}
\caption{Individual viseme classification, Pr$\{v|\hat{v}\}$ with speaker-dependent visemes for twelve speakers with continuous speech training of classifiers. B3 visemes (top) and B4 visemes (bottom).}
\label{fig:indVisContributions_lilir}
\end{figure} 
 
The same viseme comparison analysis has been repeated for our continuous speech recognition experiments and the results are shown in Figures~\ref{fig:contrib_rm_lilir} and \ref{fig:indVisContributions_lilir}.

In the isolated word data (Figures~\ref{fig:contrib_rm} and \ref{fig:contrib_sm}) the difference  between a high-performing speaker map and a poor one is striking.  Speaker 3 for example has at least five visemes in which $\mbox{Pr}\{v | \hat{v}\}=1$ (more in some configurations) whereas  Speaker 1 has only one good viseme.  Referring to  Tables~\ref{tab:tcvisemes_split} and \ref{tab:lcvisemes_split} there is no consistency on the best viseme although generally visual silence appears to be easy to spot.  This variation is to be expected -- speaker variablity is a very serious problem in lipreading.

Figures~\ref{fig:contrib_rm_lilir} and \ref{fig:indVisContributions_lilir} show the same thing for the continuous speech data.  Now  there is a shallower drop-off  to the curve and there are certainly no visemes for which  $\mbox{Pr}\{v | \hat{v}\}=1$.  Although there appears to be less variablity among speakers this is an illusion caused by the poorly-performing visemes to be similar among speakers -- within the top five visemes there are significant differences among speakers.

\section{Conclusions}
While lipreading and hence expressive audio-visual speech recognition face a number of challenges, one the persistent difficulties has been the multiplicity of mappings between phonemes and visemes. This paper has described a study of previously suggested Phoneme-to-Viseme (P2V) maps. For isolated word classification, Lee's \cite{lee2002audio} is the best of the previously published maps. For continuous speech a combination of Woodward's and Disney's visemes are better.  The best performing viseme sets have on average, between two and four phonemes per viseme. 
 
When looking at speaker-independent visemes, whilst most viseme sets do not experience any difference in correctness between isolated and continuous speech, it is interesting to note that Woodward consonant visemes are better for continuous speech and are linguistically derived, whereas Lee visemes are better for isolated words and are data-derived. This suggests that an optimal set of visemes for all speakers would need to consider both the visual speech gestures of the individual \emph{and} the rules of language. Which in essence is the dilemma for visemes: does one choose units that make sense in terms of likely visual gestures or in terms of the linguistic problem that is trying to be solved.
\begin{figure}[h]
\centering
\includegraphics[width=0.95\textwidth]{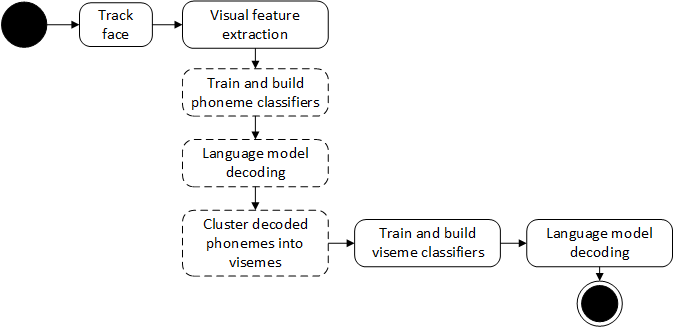}
\caption{A simple augment to the conventional lip-reading system to include speaker-dependent visemes.}
\label{fig:augment1}
\end{figure}

We have also derived some new visemes, the `Bear' visemes. These new data-driven visemes respect speaker individuality in speech and uses this property to demonstrate that our second data-driven method tested, a strictly-confused viseme derivation with split vowel and consonant phonemes, can improve word classification. The best of Bear visemes is the strict confused phonemes with split vowels and consonants ($B2$) for both isolated and continuous speech.

Furthermore, a review of these speaker-dependent visemes (listed in Tables~\ref{tab:tcvisemes_split},~\ref{tab:lcvisemes_split}, and~\ref{app:rmavP2Vmaps}) shows that formally `accepted' visemes such as \{ $/p/$ $/b/$ $/m/$ \} and \{/\textipa{S}/ /\textipa{Z}/ /d\textipa{Z}/ /t\textipa{S}/\} are no longer present. Similarly with our previous vowel based visemes, six of our eight prior viseme sets pair /\textturnv/ with /\textscripta/ (albeit not as a complete viseme, others are also present) but with our best speaker-dependent visemes these two phonemes are not paired. This is an interesting insight because it suggests that formerly `accepted' strong visemes might not be so useful for all speakers, and some adaptability, or further investigation into understanding viseme variation is still needed. Our suggestion at this time, is that linguistics or co-articulation in continuous speech, are a strong influence causing this variation.

In practical terms, our new viseme derivation method is simple and can be included within a conventional lipreading system easily. This is demonstrated in Figure~\ref{fig:augment1} where our clustering method is shown in dashed boxes.  We recommend this approach for viseme classification since speaker-independent visemes are unlikely to perform  well.

In general, for cases, Speaker-dependent visemes reduce insertion errors when classifying continuous speech. This is thought to be because the phoneme confusions in speaker-dependent visemes are affected by speaker specific visual co-articulation. For all viseme sets, not mixing vowel and consonant phonemes significantly improves classification.

 \section{Acknowledgments}
 We gratefully acknowledge the assistance of Dr Yuxuan Lan and Dr Barry-John Theobald, formerly of the University of East Anglia for their help with HTK and general advice and guidance. 
 
 This work was conducted while Helen L. Bear was in receipt of a studentship from the UK Engineering and Physical Sciences Research Council (EPSRC).
 
 
%

\bibliography{refs} 
 
 \newpage
\appendix
\section{RMAV Speaker-dependent P2V maps} 
\label{app:rmavP2Vmaps} 
\begin{table}[!ht] 
\centering 
\caption{A speaker-dependent phoneme-to-viseme mapping derived from phoneme recognition confusions for RMAV speaker sp01} 
\resizebox{\columnwidth}{!}{%
\begin{tabular}{|l||l|l|l|l|l|l|l|l|} 
\hline 
Speaker & \multicolumn{2}{| c |}{Bear1} & \multicolumn{2}{| c |}{Bear2} & \multicolumn{2}{| c |}{Bear3} & \multicolumn{2}{| c |}{Bear4} \\ 
& Viseme & Phonemes & Viseme & Phonemes & Viseme & Phonemes & Viseme & Phonemes \\ 
\hline\hline 
\multirow{25}{*}{sp01} & /v01/ & /d\textipa{Z}/ /m/ & /v01/ & /\ae/ /\textturnv/ /\textschwa/ /ay/ & /v01/ & /\textsci\textschwa/ /t/ /\textipa{T}/ /uw/ & /v01/ & /\ae/ /\textturnv/ /\textschwa/ /ay/ \\ 
& /v02/ & /\textrevepsilon/ /\textsci/ /iy/ /k/ & & /eh/ /\textsci\textschwa/ /\textsci/ /iy/ & & /z/ & & /eh/ /\textsci\textschwa/ /\textsci/ /iy/ \\ 
& & /n/ /\textipa{N}/ /r/ /s/ & /v02/ & /\textturnscripta/ /\textschwa\textupsilon/ & /v02/ & /\textrevepsilon/ /\textsci/ /iy/ /k/ & /v02/ & /\textipa{S}/ /\textipa{T}/ /v/ /w/ \\ 
& /v03/ & /ey/ & /v03/ & /\textopeno/ /\textrevepsilon/ /ey/ & & /n/ /\textipa{N}/ /r/ /s/ & & /z/ \\ 
& /v04/ & /\textschwa/ /\textipa{D}/ /\textipa{E}/ /eh/ & /v04/ & /\textscripta/ /sil/ & /sil/ & /sil/ /sil/ /sp/ & /v03/ & /b/ /d/ /f/ /k/ \\ 
& & /\textupsilon/ & /v05/ & /uw/ & /gar/ & /gar/ /\textscripta/ /\ae/ /\textturnv/ /\textopeno/ & & /m/ /n/ /\textipa{N}/ /p/ /r/ \\ 
& /v05/ & /\textscripta/ & /v06/ & /\textupsilon/ & & /\textschwa/ /ay/ /\textschwa/ /b/ /t\textipa{S}/ & & /r/ /s/ /t/ \\ 
& /v06/ & /\textsci\textschwa/ /t/ /\textipa{T}/ /uw/ & /v07/ & /\textopeno\textschwa/ & & /t\textipa{S}/ /d/ /\textipa{D}/ /\textipa{E}/ /eh/ & /sil/ & /sil/ /sil/ /sp/ \\ 
& & /z/ & /v08/ & /\textopeno\textsci/ & & /eh/ /ey/ /f/ /g/ /\textipa{H}/ & /gar/ & /gar/ /\textscripta/ /\textopeno/ /\textscripta\textupsilon/ /\textschwa/ \\ 
& /v07/ & /\textturnscripta/ /\textschwa\textupsilon/ /p/ /w/ & /v09/ & /\textschwa/ & & /\textipa{H}/ /d\textipa{Z}/ /m/ /\textturnscripta/ /\textschwa\textupsilon/ & & /\textipa{D}/ /\textrevepsilon/ /ey/ /g/ /\textipa{H}/ \\ 
& /v08/ & /\textipa{S}/ & /v10/ & /\textscripta\textupsilon/ & & /\textschwa\textupsilon/ /\textopeno\textsci/ /p/ /\textipa{S}/ /\textopeno\textschwa/ & & /\textipa{H}/ /d\textipa{Z}/ /\textturnscripta/ /\textschwa\textupsilon/ /\textopeno\textsci/ \\ 
& /v09/ & /\textopeno/ & /v11/ & /b/ /d/ /f/ /k/ & & /\textopeno\textschwa/ /\textupsilon/ /w/ /y/ /\textipa{Z}/ & & /\textopeno\textsci/ /\textopeno\textschwa/ /\textupsilon/ /uw/ /\textipa{Z}/ \\ 
& /v10/ & /\ae/ & & /m/ /n/ /\textipa{N}/ /p/ /r/ & & /\textipa{Z}/ & & /\textipa{Z}/ \\ 
& /v11/ & /d/ /g/ /\textipa{H}/ & & /r/ /s/ /t/ & & & & \\ 
& /v12/ & /b/ & /v12/ & /\textipa{D}/ /d\textipa{Z}/ & & & & \\ 
& /v13/ & /y/ & /v13/ & /\textipa{S}/ /\textipa{T}/ /v/ /w/ & & & & \\ 
& /v14/ & /\textturnv/ /ay/ & & /z/ & & & & \\ 
& /v15/ & /\textipa{Z}/ & /v14/ & /g/ & & & & \\ 
& /v16/ & /\textopeno\textschwa/ & /v15/ & /t\textipa{S}/ /\textipa{H}/ & & & & \\ 
& /v17/ & /sil/ & /v16/ & /\textipa{Z}/ & & & & \\ 
& /v18/ & /\textopeno\textsci/ & /sil/ & /sil/ /sil/ /sp/ & & & & \\ 
& /v19/ & /t\textipa{S}/ & & & & & & \\ 
& /v20/ & /\textschwa/ & & & & & & \\ 
& /v21/ & /\textscripta\textupsilon/ & & & & & & \\ 
& /gar/ & /gar/ /sp/ & & & & & & \\ 
 
\hline 
\end{tabular} %
}
\label{tab:lilirvmapssp01} 
\end{table} 
 
\begin{table}[!ht] 
\centering 
\caption{A speaker-dependent phoneme-to-viseme mapping derived from phoneme recognition confusions for RMAV speaker sp02} 
\resizebox{\columnwidth}{!}{%
\begin{tabular}{|l||l|l|l|l|l|l|l|l|} 
\hline 
Speaker & \multicolumn{2}{| c |}{Bear1} & \multicolumn{2}{| c |}{Bear2} & \multicolumn{2}{| c |}{Bear3} & \multicolumn{2}{| c |}{Bear4} \\ 
& Viseme & Phonemes & Viseme & Phonemes & Viseme & Phonemes & Viseme & Phonemes \\ 
\hline\hline 
\multirow{18}{*}{sp02} & /v01/ & /l/ /m/ /n/ /p/ & /v01/ & /\textschwa/ /ay/ /\textipa{E}/ /eh/ & /v01/ & /\textschwa/ /ay/ /b/ /d/ & /v01/ & /\textschwa/ /ay/ /\textipa{E}/ /eh/ \\ 
& & /s/ /\textipa{S}/ /t/ /v/ /w/ & & /ey/ /\textsci/ /iy/ & & /eh/ /ey/ /d\textipa{Z}/ & & /ey/ /\textsci/ /iy/ \\ 
& & /w/ & /v02/ & /\textopeno/ /\textsci\textschwa/ /\textturnscripta/ /\textschwa\textupsilon/ & /v02/ & /l/ /m/ /n/ /p/ & /v02/ & /b/ /m/ /n/ /\textipa{N}/ \\ 
& /v02/ & /g/ /\textipa{H}/ /\textsci\textschwa/ /\textsci/ & /v03/ & /\ae/ /\textturnv/ /\textscripta\textupsilon/ /\textopeno\textsci/ & & /s/ /\textipa{S}/ /t/ /v/ /w/ & & /r/ /s/ /\textipa{S}/ /t/ /v/ \\ 
& & /k/ & /v04/ & /\textupsilon/ /uw/ & & /w/ & & /v/ /w/ /y/ /z/ \\ 
& /v03/ & /\textschwa/ /ay/ /b/ /d/ & /v05/ & /\textopeno\textschwa/ & /sil/ & /sil/ /sil/ /sp/ & /sil/ & /sil/ /sil/ /sp/ \\ 
& & /eh/ /ey/ /d\textipa{Z}/ & /v06/ & /sil/ & /gar/ & /gar/ /\textscripta/ /\ae/ /\textturnv/ /\textopeno/ & /gar/ & /gar/ /\textscripta/ /\ae/ /\textturnv/ /\textopeno/ \\ 
& /v04/ & /\textscripta/ /\textopeno/ & /v07/ & /\textscripta/ & & /\textschwa/ /t\textipa{S}/ /\textipa{E}/ /\textrevepsilon/ /f/ & & /\textschwa/ /t\textipa{S}/ /d/ /\textipa{D}/ /f/ \\ 
& /v05/ & /\textrevepsilon/ /uw/ /y/ /z/ & /v08/ & /b/ /m/ /n/ /\textipa{N}/ & & /f/ /g/ /\textipa{H}/ /\textsci\textschwa/ /\textsci/ & & /f/ /g/ /\textipa{H}/ /\textsci\textschwa/ /d\textipa{Z}/ \\ 
& /v06/ & /\textturnscripta/ /\textschwa\textupsilon/ & & /r/ /s/ /\textipa{S}/ /t/ /v/ & & /\textsci/ /iy/ /k/ /\textipa{N}/ /\textturnscripta/ & & /d\textipa{Z}/ /k/ /l/ /\textturnscripta/ /\textschwa\textupsilon/ \\ 
& /v07/ & /\ae/ /\textturnv/ /\textscripta\textupsilon/ /\textopeno\textsci/ & & /v/ /w/ /y/ /z/ & & /\textturnscripta/ /\textschwa\textupsilon/ /\textopeno\textsci/ /\textipa{T}/ /\textopeno\textschwa/ & & /\textschwa\textupsilon/ /\textopeno\textsci/ /\textipa{T}/ /\textopeno\textschwa/ /\textupsilon/ \\ 
& /v08/ & /f/ /\textipa{N}/ /\textopeno\textschwa/ & /v09/ & /d\textipa{Z}/ & & /\textopeno\textschwa/ /\textupsilon/ /uw/ /y/ /z/ & & /\textupsilon/ /uw/ /\textipa{Z}/ \\ 
& /v09/ & /\textipa{E}/ & /v10/ & /d/ /\textipa{D}/ /f/ /g/ & & /z/ /\textipa{Z}/ & & \\ 
& /v10/ & /t\textipa{S}/ /\textipa{T}/ & & /k/ /l/ & & & & \\ 
& /v11/ & /\textipa{Z}/ & /v11/ & /t\textipa{S}/ /\textipa{T}/ & & & & \\ 
& /v12/ & /\textupsilon/ & /v12/ & /\textipa{Z}/ & & & & \\ 
& /v13/ & /sil/ & /sil/ & /sil/ /sil/ /sp/ & & & & \\ 
& /gar/ & /gar/ /\textschwa/ /sp/ & /gar/ & /gar/ /\textschwa/ & & & & \\ 
 
\hline 
\end{tabular} %
}
\label{tab:lilirvmapssp02} 
\end{table} 
 
\begin{table}[!ht] 
\centering 
\caption{A speaker-dependent phoneme-to-viseme mapping derived from phoneme recognition confusions for RMAV speaker sp03} 
\resizebox{\columnwidth}{!}{%
\begin{tabular}{|l||l|l|l|l|l|l|l|l|} 
\hline 
Speaker & \multicolumn{2}{| c |}{Bear1} & \multicolumn{2}{| c |}{Bear2} & \multicolumn{2}{| c |}{Bear3} & \multicolumn{2}{| c |}{Bear4} \\ 
& Viseme & Phonemes & Viseme & Phonemes & Viseme & Phonemes & Viseme & Phonemes \\ 
\hline\hline 
\multirow{20}{*}{sp03} & /v01/ & /ey/ /f/ /\textsci/ /iy/ & /v01/ & /\textipa{E}/ /\textrevepsilon/ /sil/ /uw/ & /v01/ & /ey/ /f/ /\textsci/ /iy/ & /v01/ & /ay/ /eh/ /ey/ /\textsci\textschwa/ \\ 
& & /k/ /l/ /m/ /n/ /\textipa{S}/ & /v02/ & /\textupsilon/ & & /k/ /l/ /m/ /n/ /\textipa{S}/ & & /iy/ /\textturnscripta/ /\textschwa\textupsilon/ \\ 
& & /\textipa{S}/ & /v03/ & /ay/ /eh/ /ey/ /\textsci\textschwa/ & & /\textipa{S}/ & /v02/ & /g/ /k/ /l/ /m/ \\ 
& /v02/ & /\textipa{D}/ /g/ & & /iy/ /\textturnscripta/ /\textschwa\textupsilon/ & /v02/ & /\textipa{E}/ /r/ /s/ /t/ & & /p/ /r/ /s/ /t/ /\textipa{T}/ \\ 
& /v03/ & /\textipa{E}/ /r/ /s/ /sil/ & /v04/ & /\textopeno/ & & /z/ & & /\textipa{T}/ \\ 
& & /uw/ /z/ & /v05/ & /\textopeno\textschwa/ & /sil/ & /sil/ /sil/ /sp/ & /sil/ & /sil/ /sil/ /sp/ \\ 
& /v04/ & /d/ /\textipa{T}/ /v/ /w/ & /v06/ & /\ae/ /\textturnv/ /\textschwa/ & /gar/ & /gar/ /\textscripta/ /\ae/ /\textturnv/ /\textopeno/ & /gar/ & /gar/ /\textscripta/ /\ae/ /\textturnv/ /\textopeno/ \\ 
& /v05/ & /\textopeno/ /\textschwa\textupsilon/ /p/ & /v07/ & /\textschwa/ & & /\textschwa/ /ay/ /\textschwa/ /b/ /t\textipa{S}/ & & /\textschwa/ /\textschwa/ /b/ /t\textipa{S}/ /d/ \\ 
& /v06/ & /\ae/ & /v08/ & /\textscripta\textupsilon/ & & /t\textipa{S}/ /d/ /\textipa{D}/ /eh/ /\textrevepsilon/ & & /d/ /\textipa{D}/ /\textipa{E}/ /\textrevepsilon/ /f/ \\ 
& /v07/ & /\textschwa/ /ay/ /b/ /t\textipa{S}/ & /v09/ & /\textscripta/ & & /\textrevepsilon/ /g/ /\textipa{H}/ /\textsci\textschwa/ /\textipa{N}/ & & /f/ /\textipa{H}/ /d\textipa{Z}/ /\textipa{N}/ /\textopeno\textsci/ \\ 
& /v08/ & /\textipa{N}/ & /v10/ & /g/ /k/ /l/ /m/ & & /\textipa{N}/ /\textturnscripta/ /\textschwa\textupsilon/ /\textopeno\textsci/ /p/ & & /\textopeno\textsci/ /\textipa{S}/ /\textopeno\textschwa/ /\textupsilon/ /uw/ \\ 
& /v09/ & /\textipa{H}/ & & /p/ /r/ /s/ /t/ /\textipa{T}/ & & /p/ /\textipa{T}/ /\textopeno\textschwa/ /\textupsilon/ /v/ & & /uw/ /v/ /w/ /y/ /z/ \\ 
& /v10/ & /\textscripta/ /eh/ /\textrevepsilon/ & & /\textipa{T}/ & & /v/ /w/ /y/ /\textipa{Z}/ & & /z/ /\textipa{Z}/ \\ 
& /v11/ & /\textopeno\textschwa/ /\textupsilon/ & /v11/ & /t\textipa{S}/ /d/ /\textipa{D}/ /f/ & & & & \\ 
& /v12/ & /\textturnv/ /\textsci\textschwa/ & /v12/ & /d\textipa{Z}/ /v/ /w/ /z/ & & & & \\ 
& /v13/ & /\textipa{Z}/ & /v13/ & /b/ & & & & \\ 
& /v14/ & /\textschwa/ & /v14/ & /\textipa{S}/ /\textipa{Z}/ & & & & \\ 
& /v15/ & /\textscripta\textupsilon/ & /v15/ & /\textipa{H}/ /\textipa{N}/ & & & & \\ 
& /gar/ & /gar/ /\textopeno\textsci/ /sp/ & /sil/ & /sil/ /sil/ /sp/ & & & & \\ 
& & & /gar/ & /gar/ /\textopeno\textsci/ & & & & \\ 
 
\hline 
\end{tabular} %
}
\label{tab:lilirvmapssp03} 
\end{table} 
 
\begin{table}[!ht] 
\centering 
\caption{A speaker-dependent phoneme-to-viseme mapping derived from phoneme recognition confusions for RMAV speaker sp05} 
\resizebox{\columnwidth}{!}{%
\begin{tabular}{|l||l|l|l|l|l|l|l|l|} 
\hline 
Speaker & \multicolumn{2}{| c |}{Bear1} & \multicolumn{2}{| c |}{Bear2} & \multicolumn{2}{| c |}{Bear3} & \multicolumn{2}{| c |}{Bear4} \\ 
& Viseme & Phonemes & Viseme & Phonemes & Viseme & Phonemes & Viseme & Phonemes \\ 
\hline\hline 
\multirow{20}{*}{sp05} & /v01/ & /\ae/ /\textopeno/ /\textschwa/ /\textipa{D}/ & /v01/ & /\ae/ /\textopeno/ /\textschwa/ /eh/ & /v01/ & /ay/ /b/ /d/ /w/ & /v01/ & /ay/ /uw/ \\ 
& & /\textrevepsilon/ /ey/ /\textsci/ /iy/ /k/ & & /ey/ /\textsci/ /iy/ /\textturnscripta/ /\textschwa\textupsilon/ & /v02/ & /\ae/ /\textopeno/ /\textschwa/ /\textipa{D}/ & /v02/ & /d/ /\textipa{D}/ /f/ /d\textipa{Z}/ \\ 
& & /k/ /l/ /n/ & & /\textschwa\textupsilon/ & & /\textrevepsilon/ /ey/ /\textsci/ /iy/ /k/ & & /l/ /m/ /n/ /r/ /s/ \\ 
& /v02/ & /p/ /r/ /s/ /t/ & /v02/ & /\textipa{E}/ /\textupsilon/ & & /k/ /l/ /n/ & & /s/ /\textipa{S}/ \\ 
& & /z/ & /v03/ & /ay/ /uw/ & /sil/ & /sil/ /sil/ /sp/ & /sil/ & /sil/ /sil/ /sp/ \\ 
& /v03/ & /\textsci\textschwa/ /\textipa{N}/ /uw/ /v/ & /v04/ & /\textopeno\textschwa/ & /gar/ & /gar/ /\textscripta/ /\textturnv/ /\textscripta\textupsilon/ /\textschwa/ & /gar/ & /gar/ /\textscripta/ /\ae/ /\textturnv/ /\textopeno/ \\ 
& /v04/ & /t\textipa{S}/ /\textturnscripta/ & /v05/ & /\textturnv/ /\textscripta\textupsilon/ & & /\textipa{E}/ /f/ /g/ /\textipa{H}/ /\textsci\textschwa/ & & /\textschwa/ /\textschwa/ /b/ /t\textipa{S}/ /\textipa{E}/ \\ 
& /v05/ & /ay/ /b/ /d/ /w/ & /v06/ & /\textscripta/ /\textsci\textschwa/ & & /\textsci\textschwa/ /d\textipa{Z}/ /m/ /\textipa{N}/ /\textturnscripta/ & & /\textipa{E}/ /eh/ /\textrevepsilon/ /ey/ /g/ \\ 
& /v06/ & /f/ /m/ & /v07/ & /g/ /\textipa{H}/ /t/ /v/ & & /\textturnscripta/ /\textschwa\textupsilon/ /\textopeno\textsci/ /p/ /r/ & & /g/ /\textipa{H}/ /\textsci\textschwa/ /\textsci/ /iy/ \\ 
& /v07/ & /\textscripta/ /g/ /\textipa{H}/ & /v08/ & /p/ /w/ /y/ & & /r/ /s/ /\textipa{S}/ /t/ /\textipa{T}/ & & /iy/ /\textipa{N}/ /\textturnscripta/ /\textschwa\textupsilon/ /\textopeno\textsci/ \\ 
& /v08/ & /\textschwa\textupsilon/ /\textipa{S}/ & /v09/ & /d/ /\textipa{D}/ /f/ /d\textipa{Z}/ & & /\textipa{T}/ /\textopeno\textschwa/ /\textupsilon/ /uw/ /v/ & & /\textopeno\textsci/ /p/ /t/ /\textipa{T}/ /\textopeno\textschwa/ \\ 
& /v09/ & /d\textipa{Z}/ & & /l/ /m/ /n/ /r/ /s/ & & /v/ /y/ /z/ /\textipa{Z}/ & & /\textopeno\textschwa/ /\textupsilon/ /v/ /w/ /y/ \\ 
& /v10/ & /d\textipa{Z}/ & & /s/ /\textipa{S}/ & & & & /y/ /z/ /\textipa{Z}/ \\ 
& /v11/ & /\textipa{E}/ /y/ & /v10/ & /\textipa{N}/ /\textipa{T}/ & & & & \\ 
& /v12/ & /\textipa{T}/ & /v11/ & /b/ /t\textipa{S}/ & & & & \\ 
& /v13/ & /\textturnv/ /\textscripta\textupsilon/ & /v12/ & /\textipa{Z}/ & & & & \\ 
& /v14/ & /\textipa{Z}/ & /sil/ & /sil/ /sil/ /sp/ & & & & \\ 
& /v15/ & /\textopeno\textschwa/ & /gar/ & /gar/ /\textschwa/ /\textopeno\textsci/ & & & & \\ 
& /v16/ & /j/ /h/ & & & & & & \\ 
& /gar/ & /gar/ /\textschwa/ /\textopeno\textsci/ /sil/ /sp/ & & & & & & \\ 
 
\hline 
\end{tabular} %
}
\label{tab:lilirvmapssp05} 
\end{table} 
 
\begin{table}[!ht] 
\centering 
\caption{A speaker-dependent phoneme-to-viseme mapping derived from phoneme recognition confusions for RMAV speaker sp06} 
\resizebox{\columnwidth}{!}{%
\begin{tabular}{|l||l|l|l|l|l|l|l|l|} 
\hline 
Speaker & \multicolumn{2}{| c |}{Bear1} & \multicolumn{2}{| c |}{Bear2} & \multicolumn{2}{| c |}{Bear3} & \multicolumn{2}{| c |}{Bear4} \\ 
& Viseme & Phonemes & Viseme & Phonemes & Viseme & Phonemes & Viseme & Phonemes \\ 
\hline\hline 
\multirow{20}{*}{sp06} & /v01/ & /\textschwa/ /ay/ /d/ /\textipa{D}/ & /v01/ & /\textscripta/ /\ae/ /\textturnv/ /\textschwa/ & /v01/ & /\textipa{H}/ /\textipa{N}/ /\textturnscripta/ /\textschwa\textupsilon/ & /v01/ & /\textscripta/ /\ae/ /\textturnv/ /\textschwa/ \\ 
& & /eh/ /\textsci/ /k/ /l/ /n/ & & /\textrevepsilon/ /\textsci\textschwa/ /\textsci/ /\textturnscripta/ /\textschwa\textupsilon/ & /v02/ & /\textschwa/ /ay/ /d/ /\textipa{D}/ & & /\textrevepsilon/ /\textsci\textschwa/ /\textsci/ /\textturnscripta/ /\textschwa\textupsilon/ \\ 
& & /n/ /p/ /s/ /t/ & & /\textschwa\textupsilon/ & & /eh/ /\textsci/ /k/ /l/ /n/ & & /\textschwa\textupsilon/ \\ 
& /v02/ & /v/ /w/ /y/ /z/ & /v02/ & /sil/ /uw/ & & /n/ /p/ /s/ /t/ & /v02/ & /k/ /l/ /m/ /n/ \\ 
& /v03/ & /m/ & /v03/ & /ay/ /ey/ /iy/ /\textupsilon/ & /sil/ & /sil/ /sil/ /sp/ & & /r/ /s/ /\textipa{S}/ /t/ /v/ \\ 
& /v04/ & /\textipa{H}/ /\textipa{N}/ /\textturnscripta/ /\textschwa\textupsilon/ & /v04/ & /\textscripta\textupsilon/ /\textopeno\textschwa/ & /gar/ & /gar/ /\textscripta/ /\ae/ /\textturnv/ /\textopeno/ & & /v/ /w/ /y/ /z/ \\ 
& /v05/ & /ey/ /iy/ /r/ /\textipa{S}/ & /v05/ & /\textipa{E}/ & & /\textschwa/ /b/ /t\textipa{S}/ /\textrevepsilon/ /ey/ & /sil/ & /sil/ /sil/ /sp/ \\ 
& /v06/ & /\textsci\textschwa/ & /v06/ & /\textschwa/ & & /ey/ /f/ /g/ /\textsci\textschwa/ /iy/ & /gar/ & /gar/ /\textopeno/ /\textscripta\textupsilon/ /ay/ /\textschwa/ \\ 
& /v07/ & /\textscripta/ /\ae/ /\textturnv/ /\textrevepsilon/ & /v07/ & /\textopeno/ & & /iy/ /d\textipa{Z}/ /m/ /\textopeno\textsci/ /r/ & & /t\textipa{S}/ /d/ /\textipa{D}/ /\textipa{E}/ /ey/ \\ 
& /v08/ & /f/ /\textipa{T}/ /\textopeno\textschwa/ & /v08/ & /k/ /l/ /m/ /n/ & & /r/ /\textipa{S}/ /\textipa{T}/ /\textopeno\textschwa/ /\textupsilon/ & & /ey/ /f/ /g/ /\textipa{H}/ /iy/ \\ 
& /v09/ & /uw/ & & /r/ /s/ /\textipa{S}/ /t/ /v/ & & /\textupsilon/ /uw/ /v/ /w/ /y/ & & /iy/ /d\textipa{Z}/ /\textipa{N}/ /\textopeno\textsci/ /\textipa{T}/ \\ 
& /v10/ & /uw/ & & /v/ /w/ /y/ /z/ & & /y/ /z/ /\textipa{Z}/ & & /\textipa{T}/ /\textopeno\textschwa/ /\textupsilon/ /uw/ /\textipa{Z}/ \\ 
& /v11/ & /b/ /t\textipa{S}/ /g/ & /v09/ & /b/ /t\textipa{S}/ /d/ /\textipa{D}/ & & & & /\textipa{Z}/ \\ 
& /v12/ & /\textopeno/ /d\textipa{Z}/ & & /g/ /d\textipa{Z}/ & & & & \\ 
& /v13/ & /\textipa{Z}/ & /v10/ & /\textipa{Z}/ & & & & \\ 
& /v14/ & /sil/ & /v11/ & /\textipa{H}/ /\textipa{T}/ & & & & \\ 
& /v15/ & /\textschwa/ & /v12/ & /\textipa{N}/ & & & & \\ 
& /v16/ & /\textscripta\textupsilon/ & /sil/ & /sil/ /sil/ /sp/ & & & & \\ 
& /v17/ & /u/ /w/ & /gar/ & /gar/ /\textopeno\textsci/ & & & & \\ 
& /gar/ & /gar/ /\textopeno\textsci/ /sp/ & & & & & & \\ 
\hline 
\end{tabular} %
}
\label{tab:lilirvmapssp06} 
\end{table} 
 
\begin{table}[!ht] 
\centering 
\caption{A speaker-dependent phoneme-to-viseme mapping derived from phoneme recognition confusions for RMAV speaker sp08} 
\resizebox{\columnwidth}{!}{%
\begin{tabular}{|l||l|l|l|l|l|l|l|l|} 
\hline 
Speaker & \multicolumn{2}{| c |}{Bear1} & \multicolumn{2}{| c |}{Bear2} & \multicolumn{2}{| c |}{Bear3} & \multicolumn{2}{| c |}{Bear4} \\ 
& Viseme & Phonemes & Viseme & Phonemes & Viseme & Phonemes & Viseme & Phonemes \\ 
\hline\hline 
\multirow{20}{*}{sp08} & /v01/ & /eh/ /f/ /\textipa{H}/ /\textsci/ & /v01/ & /\textscripta/ /\ae/ /\textopeno/ /\textschwa/ & /v01/ & /eh/ /f/ /\textipa{H}/ /\textsci/ & /v01/ & /\textscripta/ /\ae/ /\textopeno/ /\textschwa/ \\ 
& & /l/ /m/ /\textipa{N}/ /p/ /r/ & & /eh/ /ey/ /\textsci/ /iy/ /uw/ & & /l/ /m/ /\textipa{N}/ /p/ /r/ & & /eh/ /ey/ /\textsci/ /iy/ /uw/ \\ 
& & /r/ /s/ /t/ & & /uw/ & & /r/ /s/ /t/ & & /uw/ \\ 
& /v02/ & /\textscripta/ /\ae/ /\textopeno/ /\textschwa/ & /v02/ & /\textupsilon/ & /sil/ & /sil/ /sil/ /sp/ & /v02/ & /k/ /l/ /n/ /p/ \\ 
& & /ey/ /n/ /\textupsilon/ & /v03/ & /\textturnscripta/ /\textschwa\textupsilon/ & /gar/ & /gar/ /\textscripta/ /\ae/ /\textturnv/ /\textopeno/ & & /s/ /t/ /\textipa{T}/ /v/ /w/ \\ 
& /v03/ & /ay/ /b/ /uw/ & /v04/ & /\textsci\textschwa/ & & /\textschwa/ /ay/ /\textschwa/ /b/ /t\textipa{S}/ & & /w/ /z/ \\ 
& /v04/ & /g/ & /v05/ & /\textscripta\textupsilon/ /\textipa{E}/ & & /t\textipa{S}/ /d/ /\textipa{D}/ /\textipa{E}/ /\textrevepsilon/ & /sil/ & /sil/ /sil/ /sp/ \\ 
& /v05/ & /t\textipa{S}/ & /v06/ & /\textturnv/ /\textrevepsilon/ & & /\textrevepsilon/ /ey/ /g/ /\textsci\textschwa/ /d\textipa{Z}/ & /gar/ & /gar/ /\textturnv/ /\textscripta\textupsilon/ /\textschwa/ /b/ \\ 
& /v06/ & /\textipa{S}/ /y/ & /v07/ & /\textschwa/ & & /d\textipa{Z}/ /k/ /n/ /\textturnscripta/ /\textschwa\textupsilon/ & & /d/ /\textipa{D}/ /\textipa{E}/ /\textrevepsilon/ /f/ \\ 
& /v07/ & /\textturnscripta/ & /v08/ & /k/ /l/ /n/ /p/ & & /\textschwa\textupsilon/ /\textopeno\textsci/ /\textipa{S}/ /\textipa{T}/ /\textopeno\textschwa/ & & /f/ /g/ /\textipa{H}/ /\textsci\textschwa/ /d\textipa{Z}/ \\ 
& /v08/ & /k/ & & /s/ /t/ /\textipa{T}/ /v/ /w/ & & /\textopeno\textschwa/ /\textupsilon/ /uw/ /v/ /w/ & & /d\textipa{Z}/ /m/ /\textipa{N}/ /\textturnscripta/ /\textschwa\textupsilon/ \\ 
& /v09/ & /d\textipa{Z}/ & & /w/ /z/ & & /w/ /y/ /z/ /\textipa{Z}/ & & /\textschwa\textupsilon/ /\textopeno\textsci/ /\textipa{S}/ /\textopeno\textschwa/ /\textupsilon/ \\ 
& /v10/ & /\textipa{D}/ /v/ /w/ /z/ & /v09/ & /d/ /\textipa{D}/ /f/ /\textipa{H}/ & & & & /\textupsilon/ /y/ /\textipa{Z}/ \\ 
& /v11/ & /\textipa{T}/ /\textipa{Z}/ & & /\textipa{N}/ & & & & \\ 
& /v12/ & /\textrevepsilon/ /\textsci\textschwa/ & /v10/ & /g/ /d\textipa{Z}/ & & & & \\ 
& /v13/ & /\textscripta\textupsilon/ /\textschwa\textupsilon/ & /v11/ & /b/ /t\textipa{S}/ /\textipa{S}/ & & & & \\ 
& /v14/ & /\textturnv/ /\textipa{E}/ & /v12/ & /\textipa{Z}/ & & & & \\ 
& /v15/ & /\textopeno\textschwa/ & /v13/ & /y/ & & & & \\ 
& /v16/ & /\textschwa/ & /sil/ & /sil/ /sil/ /sp/ & & & & \\ 
& /gar/ & /gar/ /\textopeno\textsci/ /sil/ /sp/ & /gar/ & /gar/ /\textopeno\textsci/ /\textopeno\textschwa/ & & & & \\ 
 
\hline 
\end{tabular} %
}
\label{tab:lilirvmapssp08} 
\end{table} 
 
\begin{table}[!ht] 
\centering 
\caption{A speaker-dependent phoneme-to-viseme mapping derived from phoneme recognition confusions for RMAV speaker sp09} 
\resizebox{\columnwidth}{!}{%
\begin{tabular}{|l||l|l|l|l|l|l|l|l|} 
\hline 
Speaker & \multicolumn{2}{| c |}{Bear1} & \multicolumn{2}{| c |}{Bear2} & \multicolumn{2}{| c |}{Bear3} & \multicolumn{2}{| c |}{Bear4} \\ 
& Viseme & Phonemes & Viseme & Phonemes & Viseme & Phonemes & Viseme & Phonemes \\ 
\hline\hline 
\multirow{22}{*}{sp09} & /v01/ & /\textipa{D}/ /\textipa{E}/ /\textrevepsilon/ /g/ & /v01/ & /\textipa{E}/ /\textrevepsilon/ & /v01/ & /\textopeno/ /\textipa{N}/ /\textturnscripta/ /\textschwa\textupsilon/ & /v01/ & /\textturnv/ /\textschwa\textupsilon/ \\ 
& & /k/ /l/ /m/ /n/ /p/ & /v02/ & /\textscripta/ /\ae/ /\textopeno/ /\textschwa/ & /v02/ & /\textipa{D}/ /\textipa{E}/ /\textrevepsilon/ /g/ & /v02/ & /\textscripta/ /\ae/ /\textopeno/ /\textschwa/ \\ 
& & /p/ & & /eh/ /ey/ /\textsci/ /iy/ /\textturnscripta/ & & /k/ /l/ /m/ /n/ /p/ & & /eh/ /ey/ /\textsci/ /iy/ /\textturnscripta/ \\ 
& /v02/ & /\textsci\textschwa/ /y/ & & /\textturnscripta/ & & /p/ & & /\textturnscripta/ \\ 
& /v03/ & /ay/ /r/ /s/ /\textipa{S}/ & /v03/ & /\textupsilon/ /uw/ & /v03/ & /ay/ /r/ /s/ /\textipa{S}/ & /v03/ & /k/ /l/ /m/ /n/ \\ 
& & /v/ /w/ /z/ & /v04/ & /\textopeno\textschwa/ & & /v/ /w/ /z/ & & /p/ /r/ /s/ /\textipa{S}/ /t/ \\ 
& /v04/ & /\ae/ /\textturnv/ /\textschwa/ /b/ & /v05/ & /\textsci\textschwa/ & /sil/ & /sil/ /sil/ /sp/ & & /t/ /\textipa{T}/ /z/ \\ 
& & /\textipa{T}/ & /v06/ & /\textscripta\textupsilon/ & /gar/ & /gar/ /\textscripta/ /\ae/ /\textturnv/ /\textscripta\textupsilon/ & /sil/ & /sil/ /sil/ /sp/ \\ 
& /v05/ & /eh/ /ey/ /f/ /\textsci/ & /v07/ & /\textturnv/ /\textschwa\textupsilon/ & & /\textschwa/ /b/ /t\textipa{S}/ /d/ /eh/ & /gar/ & /gar/ /\textscripta\textupsilon/ /\textschwa/ /b/ /t\textipa{S}/ \\ 
& /v06/ & /\textopeno/ /\textipa{N}/ /\textturnscripta/ /\textschwa\textupsilon/ & /v08/ & /sil/ & & /eh/ /ey/ /f/ /\textipa{H}/ /\textsci\textschwa/ & & /\textipa{D}/ /\textipa{E}/ /\textrevepsilon/ /f/ /g/ \\ 
& /v07/ & /\textscripta/ & /v09/ & /k/ /l/ /m/ /n/ & & /\textsci\textschwa/ /\textsci/ /d\textipa{Z}/ /\textopeno\textsci/ /\textipa{T}/ & & /g/ /\textipa{H}/ /\textsci\textschwa/ /d\textipa{Z}/ /\textopeno\textsci/ \\ 
& /v08/ & /\textscripta/ & & /p/ /r/ /s/ /\textipa{S}/ /t/ & & /\textipa{T}/ /\textopeno\textschwa/ /\textupsilon/ /uw/ /y/ & & /\textopeno\textsci/ /\textopeno\textschwa/ /\textupsilon/ /uw/ /v/ \\ 
& /v09/ & /\textupsilon/ /uw/ & & /t/ /\textipa{T}/ /z/ & & /y/ /\textipa{Z}/ & & /v/ /w/ /y/ /\textipa{Z}/ \\ 
& /v10/ & /d\textipa{Z}/ & /v10/ & /f/ & & & & \\ 
& /v11/ & /t\textipa{S}/ & /v11/ & /d/ /\textipa{D}/ /d\textipa{Z}/ & & & & \\ 
& /v12/ & /\textipa{Z}/ & /v12/ & /g/ /v/ /w/ /y/ & & & & \\ 
& /v13/ & /\textopeno\textschwa/ & /v13/ & /b/ & & & & \\ 
& /v14/ & /sil/ & /v14/ & /t\textipa{S}/ /\textipa{H}/ & & & & \\ 
& /v15/ & /\textipa{H}/ & /v15/ & /\textipa{Z}/ & & & & \\ 
& /v16/ & /\textscripta\textupsilon/ & /sil/ & /sil/ /sil/ /sp/ & & & & \\ 
& /v17/ & /a/ /a/ & /gar/ & /gar/ /\textschwa/ /\textopeno\textsci/ & & & & \\ 
& /gar/ & /gar/ /\textschwa/ /\textopeno\textsci/ /sp/ & & & & & & \\ 
 
\hline 
\end{tabular} %
}
\label{tab:lilirvmapssp09} 
\end{table} 
 
\begin{table}[!ht] 
\centering 
\caption{A speaker-dependent phoneme-to-viseme mapping derived from phoneme recognition confusions for RMAV speaker sp10} 
\resizebox{\columnwidth}{!}{%
\begin{tabular}{|l||l|l|l|l|l|l|l|l|} 
\hline 
Speaker & \multicolumn{2}{| c |}{Bear1} & \multicolumn{2}{| c |}{Bear2} & \multicolumn{2}{| c |}{Bear3} & \multicolumn{2}{| c |}{Bear4} \\ 
& Viseme & Phonemes & Viseme & Phonemes & Viseme & Phonemes & Viseme & Phonemes \\ 
\hline\hline 
\multirow{20}{*}{sp10} & /v01/ & /\textsci/ /iy/ /d\textipa{Z}/ /l/ & /v01/ & /\textschwa/ /ay/ /eh/ /\textrevepsilon/ & /v01/ & /\textschwa/ /uw/ /v/ /w/ & /v01/ & /\ae/ /\textturnv/ /\textopeno/ /\textupsilon/ \\ 
& & /\textipa{N}/ & & /\textsci/ /iy/ /\textturnscripta/ /\textschwa\textupsilon/ & /v02/ & /\textipa{H}/ /n/ /\textturnscripta/ /\textschwa\textupsilon/ & /v02/ & /\textschwa/ /ay/ /eh/ /\textrevepsilon/ \\ 
& /v02/ & /\textipa{H}/ /n/ /\textturnscripta/ /\textschwa\textupsilon/ & /v02/ & /\ae/ /\textturnv/ /\textopeno/ /\textupsilon/ & & /r/ /s/ /t/ /\textipa{T}/ & & /\textsci/ /iy/ /\textturnscripta/ /\textschwa\textupsilon/ \\ 
& & /r/ /s/ /t/ /\textipa{T}/ & /v03/ & /\textopeno\textschwa/ & /sil/ & /sil/ /sil/ /sp/ & /v03/ & /d/ /\textipa{D}/ /f/ /\textipa{H}/ \\ 
& /v03/ & /b/ & /v04/ & /\textipa{E}/ /uw/ & /gar/ & /gar/ /\textscripta/ /\ae/ /\textturnv/ /\textopeno/ & & /l/ /m/ /n/ /p/ /r/ \\ 
& /v04/ & /\ae/ /d/ /\textipa{D}/ /\textipa{E}/ & /v05/ & /\textscripta/ /\textsci\textschwa/ & & /ay/ /\textschwa/ /b/ /t\textipa{S}/ /d/ & & /r/ /s/ /t/ /v/ /w/ \\ 
& & /ey/ /f/ & /v06/ & /\textscripta\textupsilon/ & & /d/ /\textipa{D}/ /\textipa{E}/ /eh/ /\textrevepsilon/ & & /w/ /z/ \\ 
& /v05/ & /k/ & /v07/ & /sil/ & & /\textrevepsilon/ /ey/ /f/ /g/ /\textsci\textschwa/ & /v04/ & /b/ /t\textipa{S}/ /y/ \\ 
& /v06/ & /\textschwa/ /uw/ /v/ /w/ & /v08/ & /\textopeno\textsci/ & & /\textsci\textschwa/ /\textsci/ /iy/ /d\textipa{Z}/ /k/ & /sil/ & /sil/ /sil/ /sp/ \\ 
& /v07/ & /ay/ /\textipa{S}/ /sil/ & /v09/ & /\textschwa/ & & /k/ /l/ /m/ /\textipa{N}/ /\textopeno\textsci/ & /gar/ & /gar/ /\textscripta/ /\textscripta\textupsilon/ /\textschwa/ /\textipa{E}/ \\ 
& /v08/ & /\textupsilon/ & /v10/ & /d/ /\textipa{D}/ /f/ /\textipa{H}/ & & /\textopeno\textsci/ /\textipa{S}/ /\textopeno\textschwa/ /\textupsilon/ /z/ & & /\textsci\textschwa/ /d\textipa{Z}/ /\textipa{N}/ /\textopeno\textsci/ /\textipa{S}/ \\ 
& /v09/ & /\textturnv/ /\textopeno/ /z/ & & /l/ /m/ /n/ /p/ /r/ & & /z/ /\textipa{Z}/ & & /\textipa{S}/ /\textipa{T}/ /\textopeno\textschwa/ /uw/ /\textipa{Z}/ \\ 
& /v10/ & /\textsci\textschwa/ & & /r/ /s/ /t/ /v/ /w/ & & & & /\textipa{Z}/ \\ 
& /v11/ & /t\textipa{S}/ /g/ & & /w/ /z/ & & & & \\ 
& /v12/ & /\textschwa/ /\textrevepsilon/ & /v11/ & /\textipa{S}/ & & & & \\ 
& /v13/ & /\textscripta/ /\textscripta\textupsilon/ & /v12/ & /g/ /d\textipa{Z}/ /\textipa{N}/ & & & & \\ 
& /v14/ & /\textipa{Z}/ & /v13/ & /b/ /t\textipa{S}/ /y/ & & & & \\ 
& /v15/ & /\textopeno\textschwa/ & /v14/ & /\textipa{Z}/ & & & & \\ 
& /v16/ & /\textopeno\textsci/ & /v15/ & /\textipa{T}/ & & & & \\ 
& /gar/ & /gar/ /sp/ & /sil/ & /sil/ /sil/ /sp/ & & & & \\ 
 
\hline 
\end{tabular} %
}
\label{tab:lilirvmapssp10} 
\end{table} 
 
\begin{table}[!ht] 
\centering 
\caption{A speaker-dependent phoneme-to-viseme mapping derived from phoneme recognition confusions for RMAV speaker sp11} 
\resizebox{\columnwidth}{!}{%
\begin{tabular}{|l||l|l|l|l|l|l|l|l|} 
\hline 
Speaker & \multicolumn{2}{| c |}{Bear1} & \multicolumn{2}{| c |}{Bear2} & \multicolumn{2}{| c |}{Bear3} & \multicolumn{2}{| c |}{Bear4} \\ 
& Viseme & Phonemes & Viseme & Phonemes & Viseme & Phonemes & Viseme & Phonemes \\ 
\hline\hline 
\multirow{24}{*}{sp11} & /v01/ & /iy/ /k/ /m/ /n/ & /v01/ & /uw/ & /v01/ & /\textopeno/ /\textschwa/ /ay/ /t\textipa{S}/ & /v01/ & /\ae/ /\textschwa/ /ay/ /\textipa{E}/ \\ 
& & /\textturnscripta/ /p/ /r/ /s/ /t/ & /v02/ & /\ae/ /\textschwa/ /ay/ /\textipa{E}/ & & /ey/ & & /\textrevepsilon/ /ey/ /\textsci/ /iy/ \\ 
& & /t/ & & /\textrevepsilon/ /ey/ /\textsci/ /iy/ & /v02/ & /d/ /\textipa{D}/ /f/ & /v02/ & /d\textipa{Z}/ /k/ /l/ /m/ \\ 
& /v02/ & /v/ & /v03/ & /\textscripta/ & /v03/ & /iy/ /k/ /m/ /n/ & & /\textipa{N}/ /p/ /r/ /s/ /t/ \\ 
& /v03/ & /\textopeno/ /\textschwa/ /ay/ /t\textipa{S}/ & /v04/ & /\textturnv/ /\textopeno/ /\textschwa\textupsilon/ & & /\textturnscripta/ /p/ /r/ /s/ /t/ & & /t/ /w/ \\ 
& & /ey/ & /v05/ & /\textturnscripta/ & & /t/ & /sil/ & /sil/ /sil/ /sp/ \\ 
& /v04/ & /d/ /\textipa{D}/ /f/ & /v06/ & /\textupsilon/ & /sil/ & /sil/ /sil/ /sp/ & /gar/ & /gar/ /\textscripta/ /\textturnv/ /\textopeno/ /\textscripta\textupsilon/ \\ 
& /v05/ & /w/ & /v07/ & /\textopeno\textschwa/ & /gar/ & /gar/ /\textscripta/ /\ae/ /\textturnv/ /\textscripta\textupsilon/ & & /b/ /t\textipa{S}/ /d/ /\textipa{D}/ /f/ \\ 
& /v06/ & /\textipa{S}/ & /v08/ & /sil/ & & /b/ /\textipa{E}/ /\textrevepsilon/ /g/ /\textipa{H}/ & & /f/ /g/ /\textipa{H}/ /\textsci\textschwa/ /\textturnscripta/ \\ 
& /v07/ & /\textscripta/ /\ae/ /\textturnv/ /b/ & /v09/ & /\textopeno\textsci/ & & /\textipa{H}/ /\textsci\textschwa/ /\textsci/ /d\textipa{Z}/ /l/ & & /\textturnscripta/ /\textschwa\textupsilon/ /\textopeno\textsci/ /\textipa{S}/ /\textipa{T}/ \\ 
& /v08/ & /\textscripta\textupsilon/ /\textipa{E}/ /\textrevepsilon/ /\textsci/ & /v10/ & /\textsci\textschwa/ & & /l/ /\textschwa\textupsilon/ /\textopeno\textsci/ /\textipa{S}/ /\textipa{T}/ & & /\textipa{T}/ /\textopeno\textschwa/ /\textupsilon/ /uw/ /v/ \\ 
& /v09/ & /\textipa{T}/ /\textopeno\textschwa/ & /v11/ & /\textscripta\textupsilon/ & & /\textipa{T}/ /\textopeno\textschwa/ /\textupsilon/ /uw/ /v/ & & /v/ /y/ /z/ /\textipa{Z}/ \\ 
& /v10/ & /\textschwa\textupsilon/ & /v12/ & /d\textipa{Z}/ /k/ /l/ /m/ & & /v/ /w/ /y/ /z/ /\textipa{Z}/ & & \\ 
& /v11/ & /g/ /y/ /z/ & & /\textipa{N}/ /p/ /r/ /s/ /t/ & & /\textipa{Z}/ & & \\ 
& /v12/ & /\textipa{H}/ /l/ & & /t/ /w/ & & & & \\ 
& /v13/ & /\textsci\textschwa/ /uw/ & /v13/ & /d/ /f/ /g/ /\textipa{H}/ & & & & \\ 
& /v14/ & /\textipa{Z}/ & /v14/ & /\textipa{S}/ & & & & \\ 
& /v15/ & /\textupsilon/ & /v15/ & /y/ /z/ & & & & \\ 
& /v16/ & /sil/ & /v16/ & /\textipa{D}/ /\textipa{T}/ /v/ & & & & \\ 
& /v17/ & /d\textipa{Z}/ & /v17/ & /t\textipa{S}/ & & & & \\ 
& /v18/ & /\textschwa/ & /v18/ & /\textipa{Z}/ & & & & \\ 
& /gar/ & /gar/ /\textopeno\textsci/ /sp/ & /v19/ & /b/ & & & & \\ 
& & & /sil/ & /sil/ /sil/ /sp/ & & & & \\ 
& & & /gar/ & /gar/ /\textschwa/ & & & & \\ 
 
\hline 
\end{tabular} %
}
\label{tab:lilirvmapssp11} 
\end{table} 
 
\begin{table}[!ht] 
\centering 
\caption{A speaker-dependent phoneme-to-viseme mapping derived from phoneme recognition confusions for RMAV speaker sp13} 
\resizebox{\columnwidth}{!}{%
\begin{tabular}{|l||l|l|l|l|l|l|l|l|} 
\hline 
Speaker & \multicolumn{2}{| c |}{Bear1} & \multicolumn{2}{| c |}{Bear2} & \multicolumn{2}{| c |}{Bear3} & \multicolumn{2}{| c |}{Bear4} \\ 
& Viseme & Phonemes & Viseme & Phonemes & Viseme & Phonemes & Viseme & Phonemes \\ 
\hline\hline 
\multirow{20}{*}{sp13} & /v01/ & /\textopeno/ /d/ /\textsci/ /k/ & /v01/ & /\ae/ /\textopeno/ /\textschwa/ /ay/ & /v01/ & /\textopeno/ /d/ /\textsci/ /k/ & /v01/ & /\ae/ /\textopeno/ /\textschwa/ /ay/ \\ 
& & /n/ /p/ /s/ /uw/ /v/ & & /\textrevepsilon/ /ey/ /\textsci\textschwa/ /\textsci/ /iy/ & & /n/ /p/ /s/ /uw/ /v/ & & /\textrevepsilon/ /ey/ /\textsci\textschwa/ /\textsci/ /iy/ \\ 
& & /v/ /z/ /\textipa{Z}/ & & /iy/ & & /v/ /z/ /\textipa{Z}/ & & /iy/ \\ 
& /v02/ & /\textsci\textschwa/ & /v02/ & /\textipa{E}/ /\textturnscripta/ /\textschwa\textupsilon/ /uw/ & /sil/ & /sil/ /sil/ /sp/ & /v02/ & /d/ /f/ /g/ /k/ \\ 
& /v03/ & /\textrevepsilon/ /f/ /g/ /r/ & /v03/ & /\textscripta\textupsilon/ & /gar/ & /gar/ /\textscripta/ /\ae/ /\textturnv/ /\textscripta\textupsilon/ & & /m/ /n/ /\textipa{N}/ /p/ /s/ \\ 
& /v04/ & /b/ /\textipa{D}/ /\textipa{E}/ /eh/ & /v04/ & /\textturnv/ /\textupsilon/ & & /ay/ /\textschwa/ /b/ /t\textipa{S}/ /\textipa{D}/ & & /s/ /t/ /v/ /w/ /z/ \\ 
& /v05/ & /t\textipa{S}/ & /v05/ & /\textscripta/ /\textopeno\textschwa/ & & /\textipa{D}/ /\textipa{E}/ /eh/ /\textrevepsilon/ /ey/ & & /z/ \\ 
& /v06/ & /\textscripta\textupsilon/ /iy/ /\textturnscripta/ /\textschwa\textupsilon/ & /v06/ & /sil/ & & /ey/ /f/ /g/ /\textipa{H}/ /\textsci\textschwa/ & /sil/ & /sil/ /sil/ /sp/ \\ 
& /v07/ & /\textschwa/ /\textupsilon/ & /v07/ & /\textschwa/ & & /\textsci\textschwa/ /iy/ /d\textipa{Z}/ /m/ /\textipa{N}/ & /gar/ & /gar/ /\textscripta/ /\textturnv/ /\textscripta\textupsilon/ /\textschwa/ \\ 
& /v08/ & /\ae/ /\textturnv/ /\textschwa/ /ay/ & /v08/ & /d\textipa{Z}/ /r/ /\textipa{S}/ /y/ & & /\textipa{N}/ /\textturnscripta/ /\textschwa\textupsilon/ /\textopeno\textsci/ /r/ & & /t\textipa{S}/ /\textipa{D}/ /\textipa{E}/ /\textipa{H}/ /d\textipa{Z}/ \\ 
& /v09/ & /\textscripta/ /y/ & /v09/ & /d/ /f/ /g/ /k/ & & /r/ /\textipa{S}/ /t/ /\textipa{T}/ /\textopeno\textschwa/ & & /d\textipa{Z}/ /\textturnscripta/ /\textschwa\textupsilon/ /\textopeno\textsci/ /r/ \\ 
& /v10/ & /m/ /sil/ /t/ /\textipa{T}/ & & /m/ /n/ /\textipa{N}/ /p/ /s/ & & /\textopeno\textschwa/ /\textupsilon/ /w/ /y/ & & /r/ /\textipa{S}/ /\textipa{T}/ /\textopeno\textschwa/ /\textupsilon/ \\ 
& /v11/ & /\textipa{S}/ & & /s/ /t/ /v/ /w/ /z/ & & & & /\textupsilon/ /uw/ /y/ /\textipa{Z}/ \\ 
& /v12/ & /ey/ & & /z/ & & & & \\ 
& /v13/ & /\textopeno\textschwa/ /w/ & /v10/ & /\textipa{H}/ & & & & \\ 
& /v14/ & /\textipa{N}/ & /v11/ & /b/ /t\textipa{S}/ /\textipa{D}/ & & & & \\ 
& /gar/ & /gar/ /\textopeno\textsci/ /sp/ & /v12/ & /\textipa{Z}/ & & & & \\ 
& & & /v13/ & /\textipa{T}/ & & & & \\ 
& & & /sil/ & /sil/ /sil/ /sp/ & & & & \\ 
& & & /gar/ & /gar/ /\textopeno\textsci/ & & & & \\ 
 
\hline 
\end{tabular} %
}
\label{tab:lilirvmapssp13} 
\end{table} 
 
\begin{table}[!ht] 
\centering 
\caption{A speaker-dependent phoneme-to-viseme mapping derived from phoneme recognition confusions for RMAV speaker sp14} 
\resizebox{\columnwidth}{!}{%
\begin{tabular}{|l||l|l|l|l|l|l|l|l|} 
\hline 
Speaker & \multicolumn{2}{| c |}{Bear1} & \multicolumn{2}{| c |}{Bear2} & \multicolumn{2}{| c |}{Bear3} & \multicolumn{2}{| c |}{Bear4} \\ 
& Viseme & Phonemes & Viseme & Phonemes & Viseme & Phonemes & Viseme & Phonemes \\ 
\hline\hline 
\multirow{24}{*}{sp14} & /v01/ & /t\textipa{S}/ /iy/ /d\textipa{Z}/ /m/ & /v01/ & /\ae/ /\textopeno/ /\textschwa/ /ay/ & /v01/ & /\ae/ /\textrevepsilon/ /ey/ /f/ & /v01/ & /\ae/ /\textopeno/ /\textschwa/ /ay/ \\ 
& & /\textschwa\textupsilon/ /p/ /r/ /s/ /t/ & & /eh/ /\textrevepsilon/ /ey/ /\textsci/ /iy/ & /v02/ & /\textipa{S}/ /v/ /w/ /y/ & & /eh/ /\textrevepsilon/ /ey/ /\textsci/ /iy/ \\ 
& & /t/ /\textipa{T}/ & & /iy/ & /v03/ & /t\textipa{S}/ /iy/ /d\textipa{Z}/ /m/ & & /iy/ \\ 
& /v02/ & /\textschwa/ /ay/ /\textipa{N}/ & /v02/ & /uw/ & & /\textschwa\textupsilon/ /p/ /r/ /s/ /t/ & /v02/ & /\textipa{D}/ /f/ /\textipa{H}/ /k/ \\ 
& /v03/ & /\textopeno/ /b/ /d/ /\textipa{D}/ & /v03/ & /\textupsilon/ & & /t/ /\textipa{T}/ & & /m/ /n/ /r/ /s/ /\textipa{S}/ \\ 
& & /l/ & /v04/ & /\textsci\textschwa/ /\textturnscripta/ /\textschwa\textupsilon/ & /v04/ & /\textopeno/ /b/ /d/ /\textipa{D}/ & & /\textipa{S}/ /t/ /v/ /w/ \\ 
& /v04/ & /\textipa{S}/ /v/ /w/ /y/ & /v05/ & /\textturnv/ /sil/ & & /l/ & /sil/ & /sil/ /sil/ /sp/ \\ 
& /v05/ & /g/ /\textipa{H}/ /k/ & /v06/ & /\textscripta\textupsilon/ & /sil/ & /sil/ /sil/ /sp/ & /gar/ & /gar/ /\textscripta/ /\textturnv/ /\textscripta\textupsilon/ /\textschwa/ \\ 
& /v06/ & /\textipa{E}/ /\textupsilon/ & /v07/ & /\textscripta/ & /gar/ & /gar/ /\textscripta/ /\textturnv/ /\textscripta\textupsilon/ /\textschwa/ & & /t\textipa{S}/ /d/ /g/ /\textsci\textschwa/ /d\textipa{Z}/ \\ 
& /v07/ & /\ae/ /\textrevepsilon/ /ey/ /f/ & /v08/ & /\textscripta/ & & /\textschwa/ /\textipa{E}/ /g/ /\textipa{H}/ /\textsci\textschwa/ & & /d\textipa{Z}/ /\textipa{N}/ /\textturnscripta/ /\textschwa\textupsilon/ /\textopeno\textsci/ \\ 
& /v08/ & /\textscripta/ /uw/ & /v09/ & /\textopeno\textschwa/ & & /\textsci\textschwa/ /k/ /\textipa{N}/ /\textturnscripta/ /\textopeno\textsci/ & & /\textopeno\textsci/ /p/ /\textipa{T}/ /\textopeno\textschwa/ /\textupsilon/ \\ 
& /v09/ & /\textsci\textschwa/ & /v10/ & /a/ /a/ & & /\textopeno\textsci/ /\textopeno\textschwa/ /\textupsilon/ /uw/ /\textipa{Z}/ & & /\textupsilon/ /uw/ /y/ /z/ /\textipa{Z}/ \\ 
& /v10/ & /\textturnv/ /\textturnscripta/ & /v11/ & /\textipa{D}/ /f/ /\textipa{H}/ /k/ & & /\textipa{Z}/ & & /\textipa{Z}/ \\ 
& /v11/ & /\textsci\textschwa/ & & /m/ /n/ /r/ /s/ /\textipa{S}/ & & & & \\ 
& /v12/ & /\textipa{Z}/ & & /\textipa{S}/ /t/ /v/ /w/ & & & & \\ 
& /v13/ & /\textopeno\textschwa/ & /v12/ & /z/ & & & & \\ 
& /v14/ & /sil/ & /v13/ & /y/ & & & & \\ 
& /v15/ & /\textscripta\textupsilon/ & /v14/ & /b/ /t\textipa{S}/ /d/ /\textipa{T}/ & & & & \\ 
& /v16/ & /i/ /a/ & /v15/ & /p/ & & & & \\ 
& /gar/ & /gar/ /\textschwa/ /\textopeno\textsci/ /sp/ & /v16/ & /g/ & & & & \\ 
& & & /v17/ & /d\textipa{Z}/ /\textipa{N}/ & & & & \\ 
& & & /v18/ & /\textipa{Z}/ & & & & \\ 
& & & /sil/ & /sil/ /sil/ /sp/ & & & & \\ 
& & & /gar/ & /gar/ /\textschwa/ /\textopeno\textsci/ & & & & \\ 
 
\hline 
\end{tabular} %
}
\label{tab:lilirvmapssp14} 
\end{table} 
 
\begin{table}[!ht] 
\centering 
\caption{A speaker-dependent phoneme-to-viseme mapping derived from phoneme recognition confusions for RMAV speaker sp15} \resizebox{\columnwidth}{!}{%
\begin{tabular}{|l||l|l|l|l|l|l|l|l|} 
\hline 
Speaker & \multicolumn{2}{| c |}{Bear1} & \multicolumn{2}{| c |}{Bear2} & \multicolumn{2}{| c |}{Bear3} & \multicolumn{2}{| c |}{Bear4} \\ 
& Viseme & Phonemes & Viseme & Phonemes & Viseme & Phonemes & Viseme & Phonemes \\ 
\hline\hline 
\multirow{20}{*}{sp15} & /v01/ & /\textschwa/ /d/ /\textipa{D}/ /ey/ & /v01/ & /\textschwa/ /ay/ /eh/ /ey/ & /v01/ & /\textschwa/ /d/ /\textipa{D}/ /ey/ & /v01/ & /\textschwa/ /ay/ /eh/ /ey/ \\ 
& & /\textsci/ /iy/ /k/ /l/ /m/ & & /iy/ /\textschwa\textupsilon/ /uw/ & & /\textsci/ /iy/ /k/ /l/ /m/ & & /iy/ /\textschwa\textupsilon/ /uw/ \\ 
& & /m/ /n/ /y/ & /v02/ & /\textturnv/ /\textopeno/ /\textscripta\textupsilon/ /\textipa{E}/ & & /m/ /n/ /y/ & /v02/ & /b/ /d/ /\textipa{D}/ /f/ \\ 
& /v02/ & /\textsci\textschwa/ /p/ /r/ /s/ & & /\textopeno\textsci/ & /sil/ & /sil/ /sil/ /sp/ & & /k/ /l/ /m/ /n/ /\textipa{N}/ \\ 
& & /t/ /\textipa{T}/ /z/ & /v03/ & /\textturnscripta/ & /gar/ & /gar/ /\textscripta/ /\ae/ /\textturnv/ /\textopeno/ & & /\textipa{N}/ /p/ /v/ \\ 
& /v03/ & /eh/ /\textschwa\textupsilon/ & /v04/ & /\textscripta/ /\ae/ /\textrevepsilon/ & & /ay/ /\textschwa/ /b/ /t\textipa{S}/ /\textipa{E}/ & /sil/ & /sil/ /sil/ /sp/ \\ 
& /v04/ & /\textscripta/ /\ae/ /\textturnv/ /\textopeno/ & /v05/ & /sil/ /\textopeno\textschwa/ & & /\textipa{E}/ /eh/ /\textrevepsilon/ /g/ /\textipa{H}/ & /gar/ & /gar/ /\textscripta/ /\ae/ /\textturnv/ /\textopeno/ \\ 
& /v05/ & /\textturnscripta/ & /v06/ & /\textupsilon/ & & /\textipa{H}/ /\textsci\textschwa/ /d\textipa{Z}/ /\textipa{N}/ /\textturnscripta/ & & /\textschwa/ /t\textipa{S}/ /\textipa{E}/ /\textrevepsilon/ /\textipa{H}/ \\ 
& /v06/ & /\textipa{N}/ /uw/ /v/ & /v07/ & /\textschwa/ & & /\textturnscripta/ /\textschwa\textupsilon/ /\textopeno\textsci/ /p/ /r/ & & /\textipa{H}/ /\textsci\textschwa/ /d\textipa{Z}/ /\textturnscripta/ /\textopeno\textsci/ \\ 
& /v07/ & /\textupsilon/ & /v08/ & /b/ /d/ /\textipa{D}/ /f/ & & /r/ /s/ /\textipa{S}/ /t/ /\textipa{T}/ & & /\textopeno\textsci/ /r/ /s/ /\textipa{S}/ /t/ \\ 
& /v08/ & /g/ /\textipa{H}/ /d\textipa{Z}/ & & /k/ /l/ /m/ /n/ /\textipa{N}/ & & /\textipa{T}/ /\textopeno\textschwa/ /\textupsilon/ /uw/ /v/ & & /t/ /\textipa{T}/ /\textopeno\textschwa/ /\textupsilon/ /w/ \\ 
& /v09/ & /\textrevepsilon/ & & /\textipa{N}/ /p/ /v/ & & /v/ /w/ /z/ /\textipa{Z}/ & & /w/ /y/ /z/ /\textipa{Z}/ \\ 
& /v10/ & /b/ /t\textipa{S}/ & /v09/ & /r/ /s/ /\textipa{S}/ /t/ & & & & \\ 
& /v11/ & /\textrevepsilon/ & & /z/ & & & & \\ 
& /v12/ & /ay/ /\textipa{E}/ & /v10/ & /d\textipa{Z}/ & & & & \\ 
& /v13/ & /sil/ /\textopeno\textschwa/ & /v11/ & /\textipa{Z}/ & & & & \\ 
& /v14/ & /\textscripta\textupsilon/ /\textopeno\textsci/ & /v12/ & /w/ /y/ & & & & \\ 
& /v15/ & /\textipa{Z}/ & /v13/ & /\textipa{H}/ & & & & \\ 
& /v16/ & /e/ /r/ & /v14/ & /t\textipa{S}/ & & & & \\ 
& /gar/ & /gar/ /\textschwa/ /sp/ & /sil/ & /sil/ /sil/ /sp/ & & & & \\ 
 
\hline 
\end{tabular} %
}
\label{tab:lilirvmapssp15} 
\end{table} 

\end{document}